\documentclass[10pt,twocolumn,letterpaper]{article}

\usepackage[pagenumbers]{wacv} % To force page numbers, e.g. for an arXiv version

\definecolor{wacvblue}{rgb}{0.21,0.49,0.74}
\usepackage[pagebackref,breaklinks,colorlinks,allcolors=wacvblue]{hyperref}
\usepackage[breakable, skins]{tcolorbox}
\usepackage{arydshln}
\def\wacvPaperID{15} % *** Enter the WACV Paper ID here
\def\confName{WACV}
\def\confYear{2027}

\title{Gaze Heads: How VLMs Look at What They Describe}

\author{Rohit Gandikota$^{*}$ \hspace{5em} David Bau \hspace{.7em} \vspace{3pt} \\ 
Northeastern University}

\begin{document}
\maketitle
\begin{abstract}
How a vision-language model internally solves the task of describing an image is far from obvious. We find that the model develops a specific mechanism for this: a small set of attention heads in its language-model backbone, which we call \emph{gaze heads}, whose attention tracks the image region the model is currently describing. We find them with a simple correlation score from a few forward passes, using comic strips as a controlled testbed where narrative order is laid out spatially. These \emph{gaze heads} do not just track the image tokens being described: redirecting their attention to a chosen region forces the VLM to describe that region instead. A single attention-mask intervention on the top-100 \emph{gaze heads}, fewer than $9\%$ of all heads, steers the model's answer to any chosen comic panel at $83.1\%$ accuracy, while the same intervention on random heads fails to redirect the answer, and intervening on all heads destroys generation. The same lever also extends to continuous control: switching the gaze target mid-generation makes the model wrap up its current panel description and move to the new one within a few tokens. Beyond comics, the same intervention redirects answers to chosen regions in natural COCO images. The mechanism further recurs across model sizes from 2B to 32B parameters and across other VLM architectures, although some frozen-encoder families show no comparable head set. More broadly, this shows that targeted edits identified through mechanistic analysis can serve as practical inference-time levers for steering multimodal model behavior, without any retraining. Our code, interactive demo, and datasets are available at \href{https://gaze.baulab.info/}{\textcolor[rgb]{0.21,0.49,0.74}{gaze.baulab.info}}.
\end{abstract}
    
\section{Introduction}
\label{sec:intro}

Modern vision-language models begin as language-model checkpoints and are fine-tuned to ingest image tokens alongside text. How the text-pretrained backbone adapts to that task internally, which of its thousand-plus attention heads take on visual roles, and what those roles are, remains largely open. A natural starting point is to ask whether anything inside the model behaves like human gaze: when we describe what we see, our gaze follows our words, fixating on each object as we mention it, and the attention mechanism was built on exactly this intuition. We find that the answer is yes, and only in a remarkably narrow channel of the network. In Qwen3-VL-8B, only $100$ of $1{,}152$ heads (8.7\%), all sitting in a band of mid-late layers, attend to the image region the model is currently describing, and switch to the next as the model finishes one and moves on. We call these heads \emph{gaze heads}.

Prior interpretability work has identified heads that attend to images as a whole~\cite{golovanevsky2025vlms, gandelsman2023interpreting}, including works that pick out small LVLM head sets and use them as a \emph{signal source}: \emph{Image Heads}~\cite{deng2025maskcd} are masked to build a contrastive-decoding signal against hallucination, and \emph{Localization Heads}~\cite{kang2025your} have their attention maps read out to predict bounding boxes for visual grounding. The question we ask is narrower and \emph{temporal}: which heads shift their attention, token by token, to whichever region the model is currently describing, and is that subset of heads \emph{causally sufficient} to steer the model's output?
\let\thefootnote\relax \footnote{$^{*}$Correspondence to \texttt{gandikota.ro@northeastern.edu}}

What makes \emph{gaze heads} interesting is not just that they exist, but that they appear to control what the model describes. Redirecting their attention to a different part of the image is enough to steer the VQA answer to that part at $83.1\%$ accuracy (chance $16.7\%$), while the same intervention on random non-gaze heads fails to redirect the answer. Fewer than $9\%$ of the model's attention heads carry the mechanism for which visual region gets grounded into language, giving us a lever we can move at inference time without any retraining. The effect is sharply tuned rather than monotonic: redirecting fewer heads gives only partial control, and redirecting more overrides heads the model needs for fluent output, breaking generation entirely. We can even move the lever mid-generation, and the model wraps up its current panel and starts describing whichever panel we steer toward next.

We study this using comic strips, where narrative order is encoded spatially: panels are laid out left to right, and the model must attend to each panel in sequence to describe the story. This structure lets us precisely track which heads look where and when, and verify whether changing their attention actually changes the output. We do not aim to solve reading order in comics~\cite{sachdeva2024manga, sachdeva2025panels, vivoli2024comix}; instead, we use comics as a controlled testbed with the goal of understanding how a general-purpose VLM routes visual information internally.

Finding \emph{gaze heads} is cheap: it needs no training and no labeled supervision, just simple forward passes. The same procedure recovers a comparable head set across model sizes from 2B to 32B parameters and across multiple other VLM architectures, and the lever transfers beyond comic panels to natural images, hinting that \emph{gaze heads} are a recurrent organizational feature of vision-language models.

\section{Related Work}
\label{sec:related}

\paragraph{Attention heads as units of computation.}
Mechanistic interpretability has found that individual attention heads implement identifiable functions~\cite{elhage2021mathematical, conmy2023automated}: induction heads that copy in context~\cite{olsson2022context}, heads for indirect object identification~\cite{wang2022interpretability}, and the broader finding that most heads can be pruned while a small subset does the heavy lifting~\cite{michel2019sixteen, voita2019analyzing}. In vision, CLIP's representation decomposes across heads with spatial specializations~\cite{gandelsman2023interpreting}, and causal mediation has linked heads to object detection in VLMs~\cite{golovanevsky2025vlms}. \emph{Gaze heads} continue this line of research, localizing visual grounding to a small, interpretable set of heads, but with a function none of these works isolate: tracking the region the model is currently describing.

\paragraph{Image-attending heads in LVLMs.}
\textit{Image Heads}~\cite{deng2025maskcd} are attention heads whose image-token attention is outlying within their layer; the MaskCD framework masks their image attention to build a degraded contrastive sample, then subtracts the masked-pass logits from the original to suppress hallucinations. \textit{Localization Heads}~\cite{kang2025your} are the few heads whose text-to-image attention is spatially concentrated (high attention sum, low spatial entropy); their attention maps are assembled directly into a bounding-box or segmentation-mask prediction for training-free visual grounding, with only three heads sufficient. In both works, LVLM heads serve as a \emph{signal source}, scored by a static property of a single image-text input and either masked to subtract logits (MaskCD) or read off as the answer (Localization Heads). We instead treat LVLM heads as a \emph{causal control surface}, and we select heads by a \emph{temporal} criterion: which heads re-route their attention to match the queried region across multiple forward passes. Intervening on that head set is causally sufficient to redirect what the model describes to a chosen visual region; neither prior work attempts this kind of output-level steering. We adopt Image Heads and Localization Heads as our baselines and run them through an identical intervention.

\paragraph{Gaze, steering, and VLM internals.}
A separate line uses human gaze as a training signal: Voila-A~\cite{yan2024voila} and Gaze-VLM~\cite{pani2025gaze} supervise VLM attention toward human fixations. We use no gaze supervision; the gaze-like mechanism is already present, and we simply locate and control it. On the methods side, our intervention builds on representation steering, where a difference-of-means direction added to the residual stream shifts model behavior~\cite{turner2023activation, zou2023representation}; we use this to localize the relevant layers, then move to direct attention-head edits for per-region control. More broadly, studies of VLM internals show that cross-modal transfer happens from the middle layers onward~\cite{cohen2025performance, basu2024understanding, neo2024towards} and that position is bound to visual features by emergent indexing~\cite{assouel2025visual}, but which heads actively direct visual focus during generation has remained open.

\paragraph{Spatial bias and comics as a testbed.}
VLMs systematically favor left-positioned content and misallocate attention on spatial tasks~\cite{chaudhary2025spatial, chen2025spatial, chen2024spatialvlm, lee2025perspective}, and reordering inputs can swing accuracy substantially~\cite{chen2024premise}; these works document the behavior, while \emph{gaze heads} offer a mechanism behind it. We study comics because they encode narrative order spatially, giving an unambiguous ground truth for which region the model should attend to at each step. Computational comics research has largely treated reading order as a task to solve~\cite{aizawa2020manga109, iyyer2017amazing, vivoli2024comix, vivoli2024survey, sachdeva2024manga, sachdeva2025panels}; we instead use the spatial layout of comics as a controlled testbed for studying how VLMs route visual attention internally.

\section{Experimental Setup}
\label{sec:setup}

All experiments are primarily conducted on Qwen3-VL-8B-Instruct, with both discovery and evaluation on six-panel comic strips. The discrete panels give us unambiguous ground truth for which image region the model should attend to at each step. We later extend the analysis to natural images (\cref{sec:natural_images}), to other VLM sizes and architectures (\cref{sec:generalization}), and to varying panel counts and prompt formulations (\cref{app:generalization}).

\paragraph{Model.}
Qwen3-VL-8B-Instruct~\cite{qwen3vl} pairs a ViT-based vision encoder with a 36-layer language-model backbone of 32 attention heads per layer ($1{,}152$ heads in total; hidden dimension 4096, head dimension 128). All experiments run with \texttt{eager} attention so attention weights and hidden states are directly accessible.

\paragraph{Dataset.}
Discovery runs on COMICS~\cite{iyyer2017amazing}, a corpus of 3{,}948 comics. For each sample we take $N{=}6$ consecutive panels from one comic, resize them to a common height, and concatenate them horizontally into a strip; the entire strip is then fed to the model as a single image input. Panel widths vary across comics, so each strip has a different total width and a different number of image tokens per panel. Evaluation uses a held-out set of 500 six-panel strips generated with GPT Image-1~\cite{gptimage_2025}, where every panel is a visually distinct scene; this lets us verify unambiguously which panel the model is grounding its answer in. All redirection and narration results in the paper come from this validation set, disjoint from the discovery data. Hardware, sample sizes, and hyperparameters are in \cref{app:details}.

\begin{figure}[t]
  \centering
  \includegraphics[width=\linewidth]{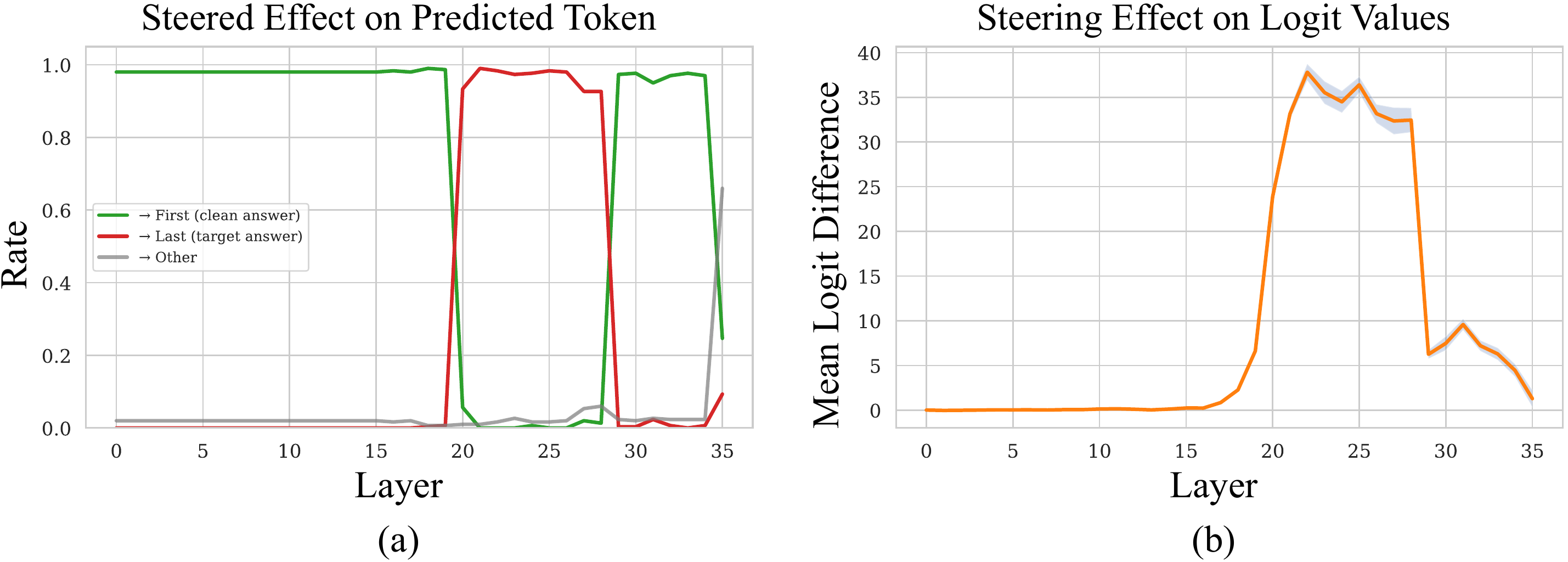}
  \caption{Layer-wise steering analysis on Qwen3-VL-8B.
  (a)~Adding a per-layer ``read-in-reverse'' direction to the residual stream: layers 20--28 sharply switch the predicted panel label from the first panel (green) to the target last panel (red), while other layers leave the answer unchanged.
  (b)~The corresponding change in logit difference (last minus first panel label) peaks over the same band.
  Visual attention routing is concentrated in a narrow middle-layer band rather than spread through the network.}
  \label{fig:layer_flip}
\end{figure}
% =============================================================================
\section{Localizing Gaze in the Network}
\label{sec:layer_analysis}

Comic strips give us a natural way to ask where in the model the notion of ``reading order'' lives. They have a clear left-to-right layout, and the model can be asked to identify a specific panel and answer correctly. We test whether the representation behind this localizes to a particular band of layers.

To probe this, we overlay each panel with a random A--Z label so the model's answer is a single letter, and run two prompts on the same strip: a normal prompt asking for the label on the $k$-th panel, and a ``reverse'' prompt that prepends \textit{``Read the comic in reverse,''} to the same question. Under the reverse prompt the model returns the label on the $k$-th panel counted from the right rather than the left. From the activations preceding the answer, averaged over 500 (normal, reverse) pairs, we take the difference-of-means to get a per-layer \emph{read-in-reverse} direction.

We then add this direction back into the residual stream at one layer at a time during a fresh forward pass with the normal prompt, and measure the rate at which the model's predicted label flips from the original (left-to-right) answer to the reverse-reading answer. \cref{fig:layer_flip} shows the result. Only a narrow band of layers $20$--$28$ produces the flip; outside the band the same direction has no effect. The direction also transfers to free-form narration and reverses the order in which the model describes the strip (\cref{app:layer_freeform}).

This isolates \emph{reverse} as a coherent residual direction in the mid-layer band, but only reverse. We repeated the construction for all $6!{=}720$ panel orderings; only reverse produces strong steering ($91.3\%$), and the other 719 produce much weaker steering (\cref{app:arbitrary_orderings}). And yet the model has no trouble returning the right panel when asked for any $k$. So whatever mechanism handles arbitrary panel queries cannot be a global residual direction; it must live elsewhere.

\begin{figure*}[t]
  \centering
  \includegraphics[width=\linewidth]{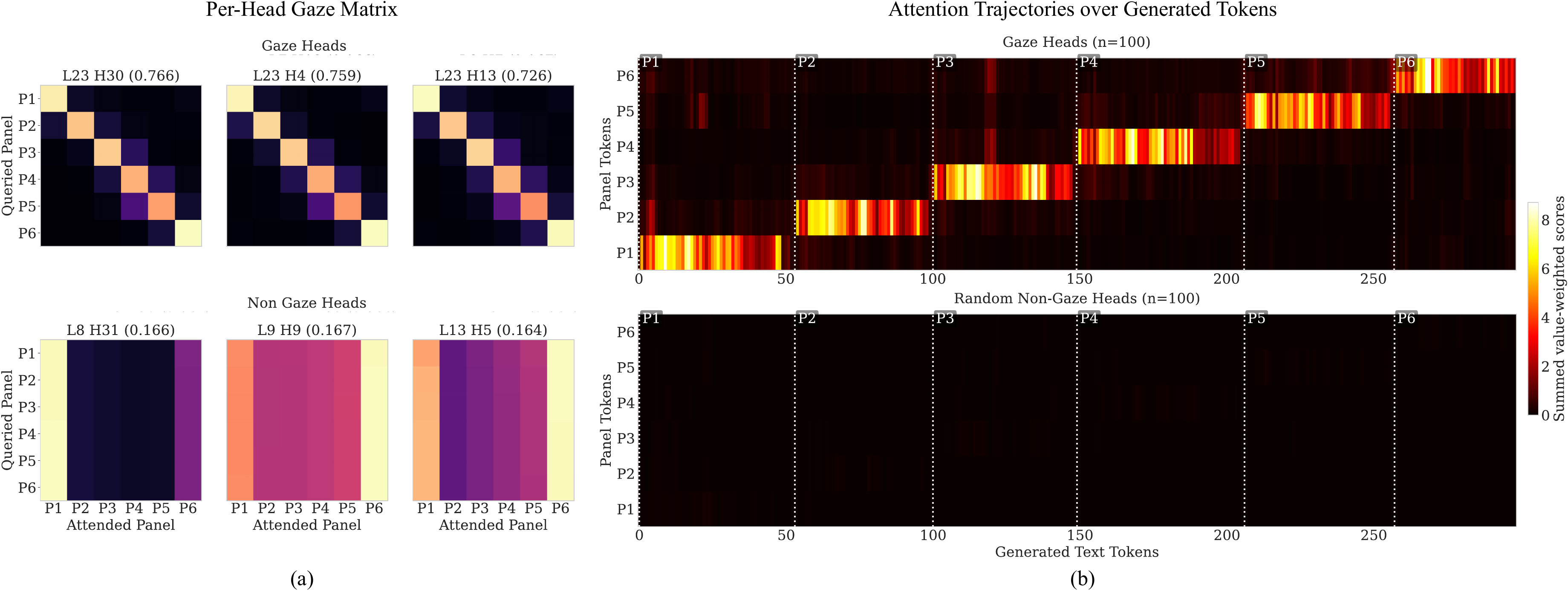}
  \caption{Gaze heads track the queried panel under both controlled prompting and unconstrained narration.
  (a)~Per-head $6{\times}6$ gaze matrices, with rows the queried panel and columns the attended panel. The three top-scoring gaze heads (top) place attention on the diagonal, tracking whichever panel is queried; three non-gaze heads (bottom) attend diffusely and prompt-independently.
  (b)~During free-form narration, the top-100 gaze heads (top) shift attention panel-by-panel in a staircase aligned with the generated text, whereas 100 random non-gaze heads (bottom) show no panel-tracking structure. Dashed lines mark where the model finishes one panel description and begins the next.
  }
  \label{fig:head_gaze_heatmap}
\end{figure*}

% =============================================================================
\section{Discovering Gaze Heads}
\label{sec:gaze_heads}

Attention heads are a natural place to look for this mechanism, since they are how the model routes between text and image tokens. But the model has over a thousand of them, and we don't know in advance which are doing the work. So we score every head, across all layers, on how its attention re-routes as the queried panel changes.

\subsection{Gaze Score}
\label{sec:gaze_score}

For each panel index $k \in \{1, \ldots, 6\}$ in a test strip, we run a forward pass with the same natural-language query, \textit{``Look carefully at this six-panel comic strip. What is happening in the $k$-th panel from the left? Answer briefly.''} Unlike the labeled probe in \cref{sec:layer_analysis}, this prompt has no letter overlays, so the model must rely on spatial position alone to identify the queried panel. From each forward pass we pull the post-softmax attention weights from the final prompt token to all image tokens, grouped by which panel they belong to.

Across the six queries, every head produces a $6{\times}6$ attention matrix: rows are queried panels and columns are attended panels. A head that perfectly tracks the queried panel would put its mass on the diagonal. The \emph{gaze score} measures exactly this:
\begin{equation}
  \text{GazeScore}(l, h) \;=\; \frac{1}{6}\sum_{k=1}^{6} A^{(l,h)}_{k,k}
  \label{eq:gaze_score}
\end{equation}
where $A^{(l,h)}_{k,j}$ is the raw post-softmax attention mass that the generation token places on panel~$j$'s image tokens when the prompt asks about panel~$k$, summed over those tokens and averaged over 500 strips.

We use raw attention scores rather than normalizing them since we want heads that \emph{both} look at images \emph{and} concentrate that look on the queried panel. A normalized score would catch only the second property; a head that ignored image tokens entirely could still produce a perfectly diagonal \emph{shape} once normalized. The raw variant scores such a head near zero, and only boosts the heads that put real mass on the right image tokens.

\cref{fig:head_gaze_heatmap}a contrasts the $6{\times}6$ matrices of the top-scoring gaze heads with low-scoring control heads. Gaze heads produce a clean near-diagonal pattern, putting attention on panel~$k$ when asked about panel~$k$; non-gaze heads attend diffusely and prompt-independently. The top-scoring heads concentrate in layers $20$--$28$, the same band the residual analysis localized in \cref{sec:layer_analysis}, even though our gaze-score search ranged over all 1{,}152 heads without restriction (full distribution in \cref{app:gaze_distribution}). We pick the top-100 heads by gaze score as our default set; \cref{sec:redirect} shows that redirection accuracy saturates around this threshold. The discovery procedure is intentionally cheap: ask the model about each panel, record which heads shift, sort. What it leaves open is whether heads picked this way, under controlled prompting, also govern unconstrained generation.

\subsection{Gaze Heads Track Narration in Real Time}
\label{sec:trajectory}

During free generation the model gets no panel query; it has to decide where to look on its own. To check whether \emph{gaze heads} still track the relevant region in this setting, we prompt the model to describe each panel in order and record value-weighted attention~\cite{kobayashi2020attention} at every decode token, aggregated per panel, comparing the 100 \emph{gaze heads} against 100 random non-gaze heads.

\cref{fig:head_gaze_heatmap}b shows that the gaze-head attention forms a clean staircase: it sits on panel~1 while the model narrates the first panel, jumps to panel~2 within a few tokens once the narration moves on, and continues panel by panel through all six. The non-gaze control shows no such structure. Prompted to narrate in reverse, the same heads produce a mirror-image reverse staircase (\cref{fig:reverse_trajectory}). Gaze heads faithfully track the panel being narrated. The tracking is a property of the heads themselves, not an artifact of the controlled-prompting setup that found them.

\begin{figure*}[t]
    \centering
    \includegraphics[width=\linewidth]{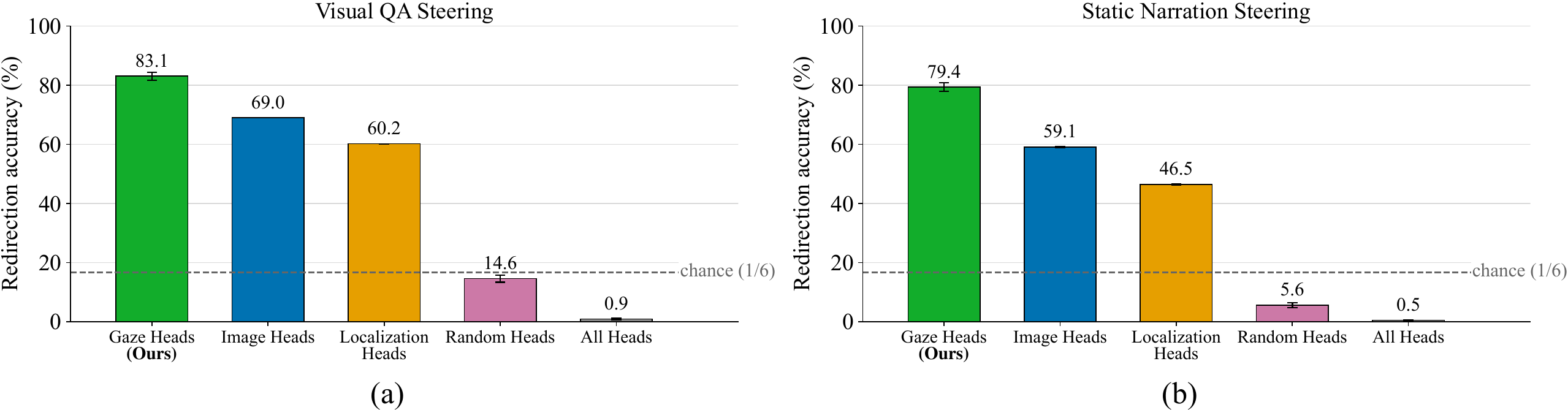}
    \caption{Redirecting attention with a single attention-mask intervention, over 500 strips ($n{=}3{,}000$ strip-target pairs), forced 1-of-6 LLM judge, chance $16.7\%$, bootstrap 95\% CIs.
    (a)~Visual question answering and (b)~static narration.
    Redirecting the top-100 gaze heads reaches $83.1\%$ and $79.4\%$ accuracy, above the Image Heads~\cite{deng2025maskcd} and Localization Heads~\cite{kang2025your} baselines run through the same intervention. Random non-gaze heads fail to redirect the answer, and intervening on all heads destroys generation.}
    \label{fig:causal_swap}
\end{figure*}

\begin{figure*}[t]
  \centering
  \includegraphics[width=\linewidth]{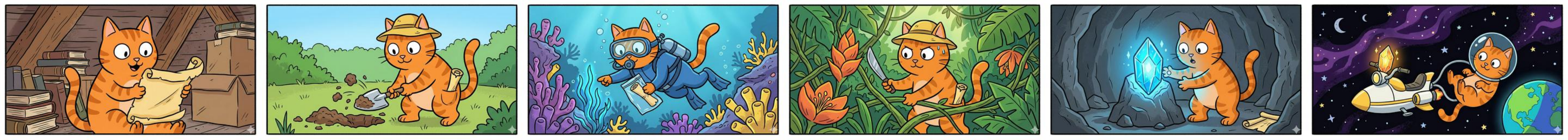}
  \vspace{4pt}
  \begin{tcolorbox}[colback=white, colframe=black!70, fontupper=\small,
        title={\small Visual QA task with gaze steering},
        coltitle=black, colbacktitle=white,
        left=4pt, right=4pt, top=3pt, bottom=3pt]
    \scriptsize
    \textit{``Read each of the panels and tell me. What is the main action happening in this particular comic panel? Just output the answer in few words, do not include any other text.''}

    \medskip
    \textcolor{red!70!black}{Baseline (no steering):} Cat discovers treasure map

    \medskip
    \textcolor{blue!70!black}{Steered responses when gaze redirected to:}\\[3pt]
    \begin{tabular}{@{}ll}
    \textcolor{blue!70!black}{Panel 1:} & Cat discovers treasure map. \\
    \textcolor{blue!70!black}{Panel 2:} & Cat digs, digs, and digs. \\
    \textcolor{blue!70!black}{Panel 3:} & Cat explores underwater. \\
    \textcolor{blue!70!black}{Panel 4:} & Cat explores jungle. \\
    \textcolor{blue!70!black}{Panel 5:} & Cat discovers gem. \\
    \textcolor{blue!70!black}{Panel 6:} & Cat explores space.
    \end{tabular}
    \end{tcolorbox}
  \caption{Gaze-head steering on visual question answering. The same question is asked in every condition. Without steering, the answer summarizes across all six panels; redirecting the gaze heads to a chosen panel makes the answer describe that panel's content only.}
  \label{fig:main_qual_vqa}
\end{figure*}
% =============================================================================
\section{Gaze Heads Steer What the Model Describes}
\label{sec:steer}

The staircase shows that \emph{gaze heads} track which panel is being described, but tracking is correlational. We now ask the causal question: if we force these heads to attend elsewhere, does the model describe that panel instead?

We test redirection on two complementary tasks. In \emph{visual question answering (VQA)}, the model sees a single question about the strip (\textit{``What is the main action or event happening in this comic strip? Answer briefly.''}) and we score whether the steered answer describes the chosen target panel rather than the full strip. In \emph{static narration}, the model is asked \textit{``What is happening in this panel of the comic strip?''} without specifying which panel, with the gaze heads held on a single target panel; we score whether the answer resolves the ambiguity to the target panel rather than the model's default reading (the first panel or a whole-strip summary). VQA tests whether redirection overrides a strip-level answer; static narration tests whether redirection alone decides which panel the model talks about.

\subsection{Redirecting Gaze}
\label{sec:redirect}

For each of the 100 gaze heads, we inject an additive bias into the pre-softmax attention mask during both prefill and decoding: $+\delta$ on the target panel's image tokens and $-\delta$ on every other panel, with $\delta = +\infty$. Text-token attention is left untouched, and nothing else about the model is modified. The redirection effect is not sensitive to this choice; a sweep over $\delta$ (\cref{app:delta_ablation}) shows it saturates well before the hard limit. The result is also robust to the wording of the VQA prompt (\cref{app:prompt_sensitivity}).

We evaluate on the 500 held-out strips, targeting each panel in turn ($3{,}000$ strip-target pairs). A forced 1-of-6 LLM judge (Claude Sonnet~\cite{anthropic2025claude}; \cref{app:judge_match}) sees the strip and the steered text and picks the single panel the answer best matches; junk and unmatchable outputs count as misses, with chance at $1/6$. \cref{fig:causal_swap} reports redirection accuracy for visual question answering and for static narration. Redirecting the top-100 gaze heads steers the answer to the chosen panel with $83.1\%$ accuracy on VQA and $79.4\%$ on narration, far above chance. The same intervention on random non-gaze heads fails to redirect the answer, and applying it to all 1{,}152 heads collapses generation to junk: the effect is specific to the gaze head set, and fewer than $9\%$ of the model's heads are enough to control which region gets grounded into language. \cref{fig:main_qual_vqa} illustrates this on a single strip, where one question yields six different answers depending on where the gaze heads are pointed.

\cref{fig:causal_swap} compares against the two prior head sets. Running the Image Heads~\cite{deng2025maskcd} and Localization Heads~\cite{kang2025your} selectors through the \emph{identical} intervention redirects the model well above chance but below gaze heads, on both VQA and narration. The gap traces back to what each criterion measures: Image Heads and Localization Heads rank heads by how much, or how concentrated, their image attention is in a single forward pass, whereas the gaze score rewards heads that \emph{re-route} as the queried region changes. It is that temporal-tracking signal, which neither single-pass criterion captures, that picks out the heads most worth steering. The three head sets are far from interchangeable: at $K{=}10$ they are nearly disjoint, and only $13$ heads sit in all three top-$100$ sets (\cref{tab:head_set_agreement}).

\begin{figure*}[t]
  \centering
  \includegraphics[width=\linewidth]{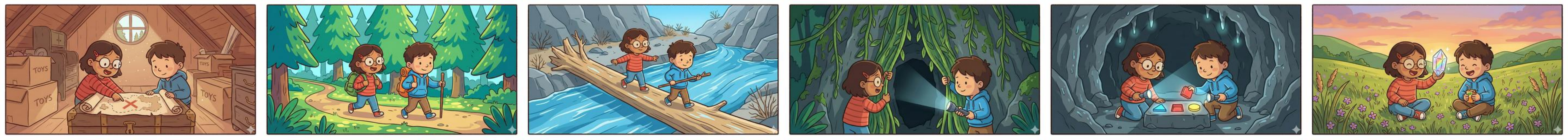}
  \includegraphics[width=0.9\linewidth]{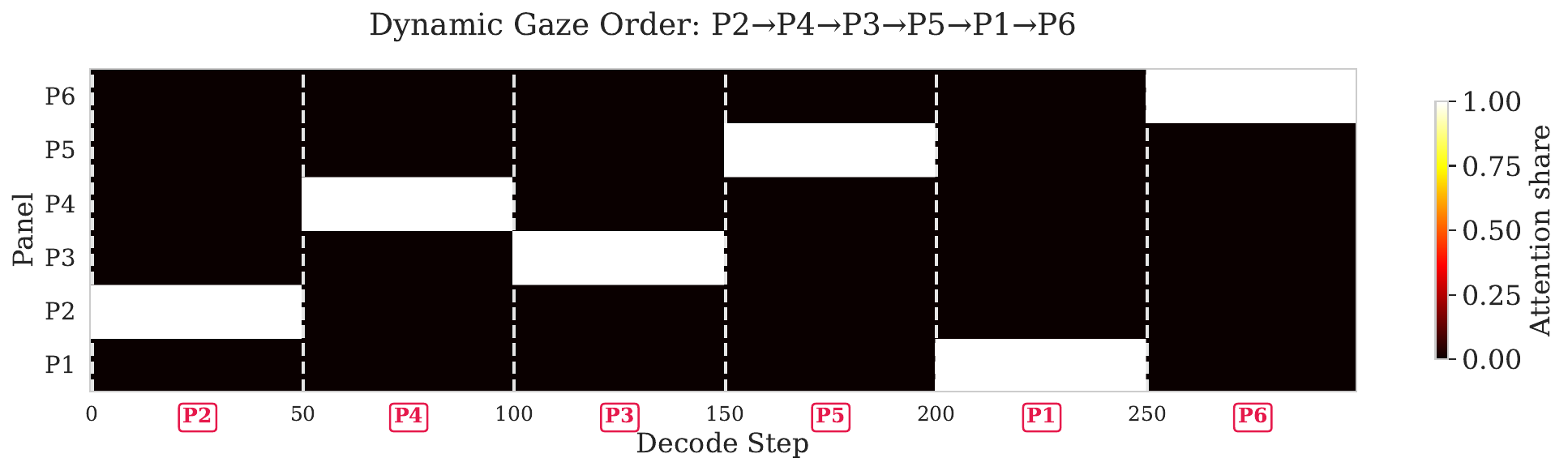}
  \begin{tcolorbox}[colback=white, colframe=black!70, fontupper=\small,
    title={\small Generated narration under dynamic gaze switching \\ {\scriptsize(schedule: \textcolor{red!70!black}{P2}\,$\to$\,\textcolor{blue!70!black}{P4}\,$\to$\,\textcolor{orange!80!black}{P3}\,$\to$\,\textcolor{green!50!black}{P5}\,$\to$\,\textcolor{purple!70!black}{P1}\,$\to$\,\textcolor{brown!70!black}{P6}, switching every 50 tokens)}},
    coltitle=black, colbacktitle=white,
    left=4pt, right=4pt, top=3pt, bottom=3pt]
  \scriptsize
  \textcolor{red!70!black}{1. Two friends hike through forest, backpacks in tow, exploring nature's wonders. Trees tower as they trek, adventure awaiting beyond the woods' embrace. 2. Forest path winds through lush greenery, friends hike with backpacks and trees.}
  \textcolor{blue!70!black}{Nature's mystery awaits as they explore, hidden treasures hidden in shadows. 3. Children explore jungle, flashlight illuminates path. Hidden cave looms ahead, mystery beckons as vines dangle, adventure awaits in darkness. 4. Kids discover hidden cave}
  \textcolor{orange!80!black}{, river flows nearby. Nature's beauty surrounds them, adventure awaits as they cross streams, exploring wild landscapes. 5. Kids cross river, bridge spans rocky path. Nature's beauty surrounds them, adventure awaits as they explore, crossing streams, finding hidden}
  \textcolor{green!50!black}{treasures. 6. Kids explore cave, glowing gems shine. Adventure ends as they discover hidden treasures, cave's secrets revealed in glowing light.}
  \textcolor{purple!70!black}{7. Kids explore attic, treasure chest awaits. Hidden secrets await, mystery unfolds as they uncover hidden treasures, childhood dreams come alive. 8. Kids explore attic, treasure chest awaits. Hidden secrets await, mystery unfolds as they uncover}
  \textcolor{brown!70!black}{hidden treasures, childhood dreams come alive. 9. Kids explore field, flowers bloom. Joyful moments unfold, nature's beauty surrounds them, friendship blooms with every step.}
  \end{tcolorbox}
\caption{A single dynamic-switching run. Top: the six-panel strip. Middle: gaze-head attention during generation, with the target switched to a new panel every 50 tokens. Bottom: the model keeps its default ``1, 2, 3\ldots'' numbering but describes the content of whichever panel the gaze heads are steered toward, transitioning cleanly at each switch. The numbering follows the model's default textual structure while the visual content follows the gaze heads, suggesting the two mechanisms are functionally separate.}
  \label{fig:dynamic_switch}
\end{figure*}

\begin{figure}[t]
    \centering
    \includegraphics[width=\linewidth]{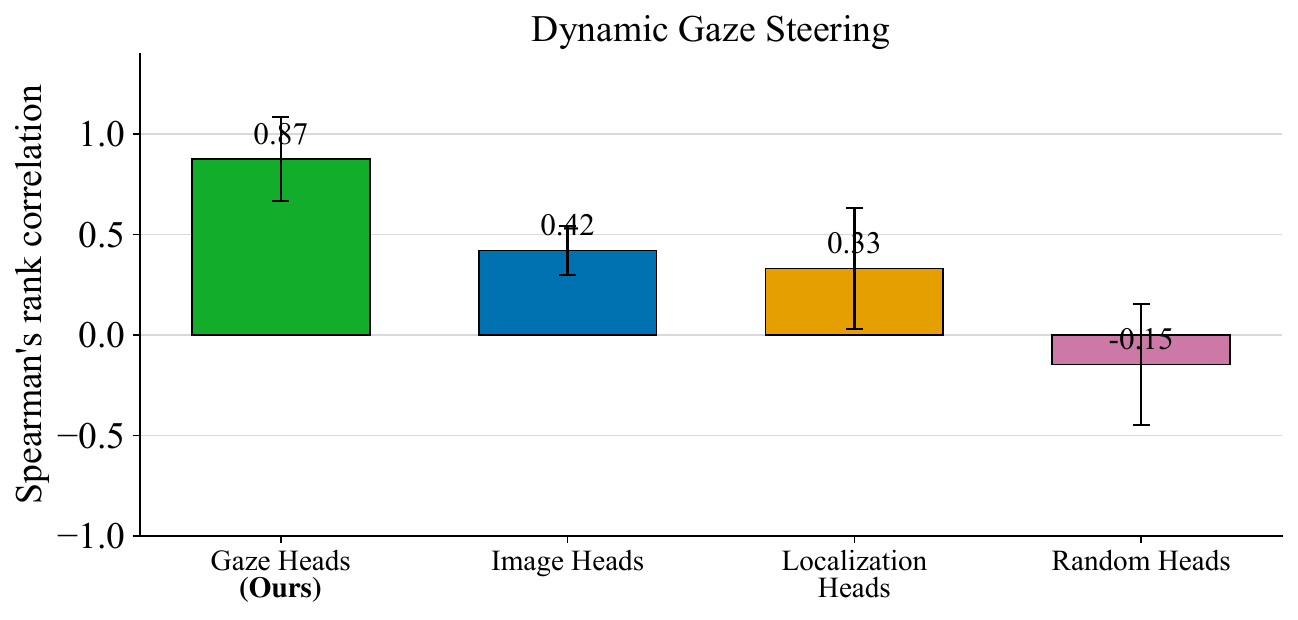}
    \caption{Dynamic gaze steering. The target panel is switched every 50 generated tokens through a random derangement schedule. Spearman correlation between the schedule and the order the model actually describes is shown. The top-100 gaze heads follow the schedule ($\rho{=}0.87$); the baselines follow it only weakly, and random non-gaze heads are slightly anti-correlated.}
    \label{fig:dynamic_bar}
\end{figure}

\subsection{Dynamic Gaze Switching During Generation}
\label{sec:dynamic_switch}

Redirection so far points the \emph{gaze heads} at one fixed panel. Can we switch the target mid-generation, and does the model fold each switch into its narration? We generate a 300-token narration while changing the gaze-head target every 50 decode steps. Each strip uses an independently sampled derangement of the six panels, so no schedule starts at panel~1 or follows the model's default left-to-right order.
\begin{figure*}
  \centering \includegraphics[width=\linewidth]{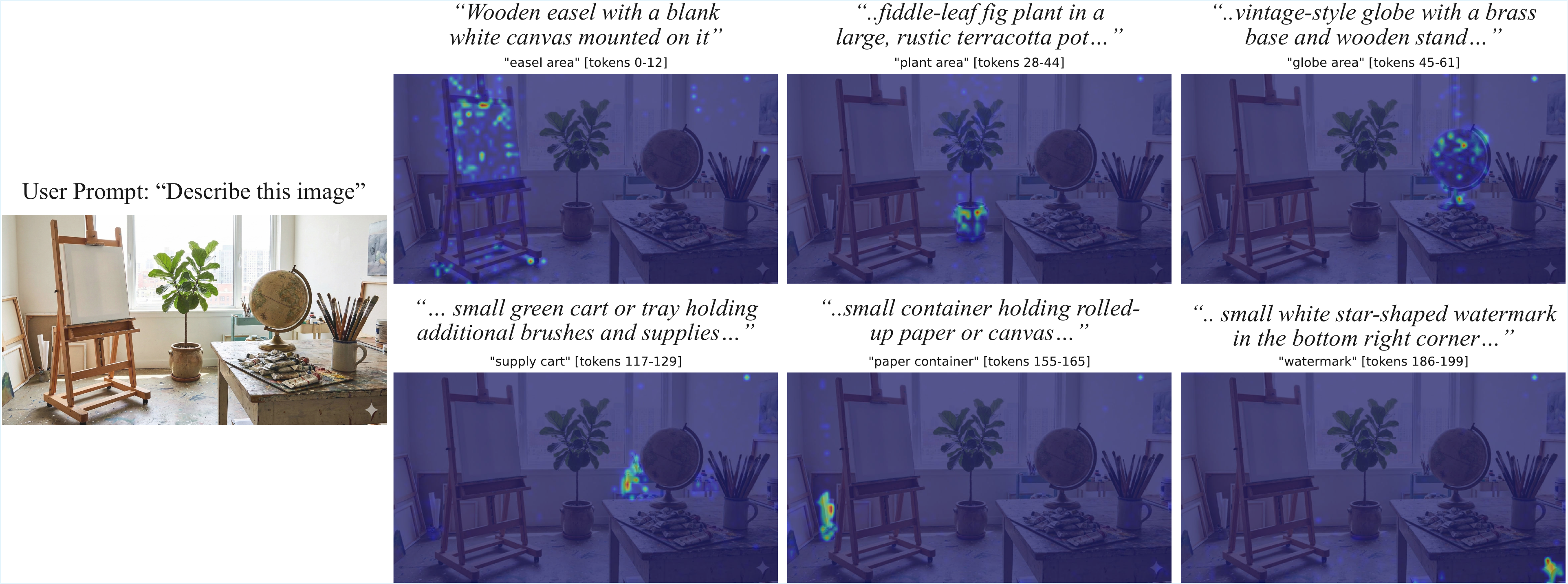}
  \caption{Gaze-head attention on a natural image. Left: the original image. Each heatmap averages gaze-head attention over the output tokens where the model describes one object. Attention concentrates on the spatial region of the described object, showing that gaze heads ground attention spatially beyond comic panels.}
  \label{fig:natural_image_gaze}
\end{figure*}
\begin{figure}[t]
  \centering \includegraphics[width=\linewidth]{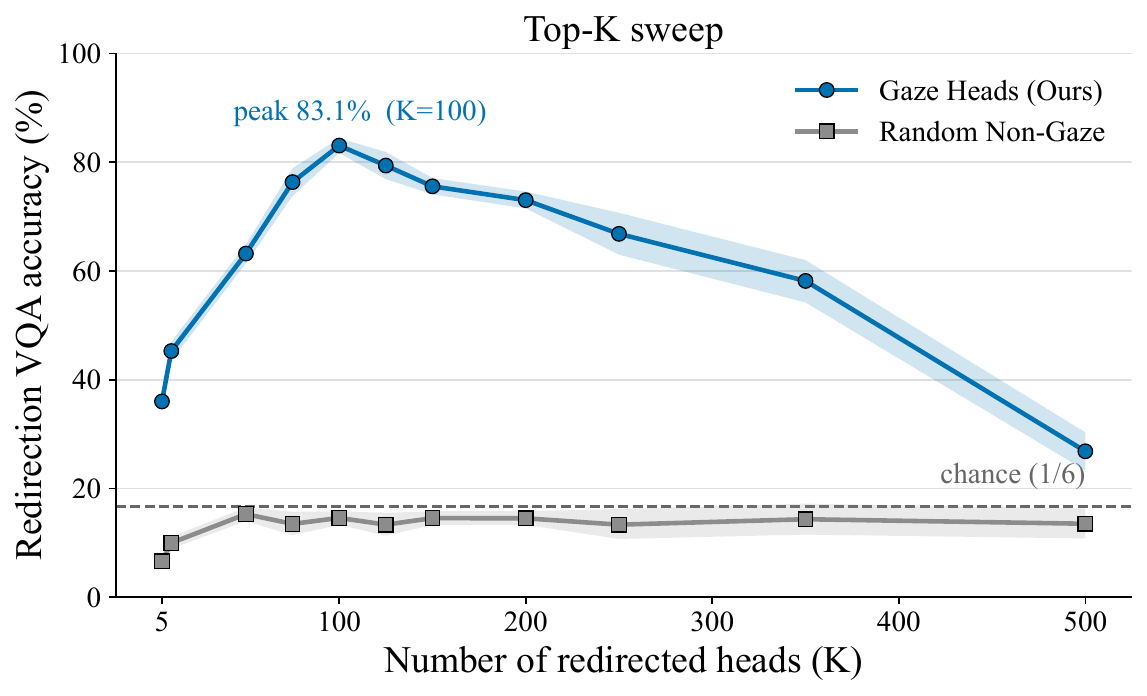}
  \caption{Top-$K$ saturation for VQA redirection. Accuracy as a function of how many top-ranked gaze heads are redirected. It rises steeply, peaks at $83.1\%$ with $K{=}100$ heads (under $9\%$ of the model), and declines past the peak as the intervention starts overriding heads needed for coherent generation. The random non-gaze control fails to redirect, staying near the $1/6$ chance line throughout.}
  \label{fig:topk_saturation}
\end{figure}

\cref{fig:dynamic_bar} measures how well the panel the model actually describes tracks the steering schedule, using Spearman correlation. The top-100 gaze heads track the schedule strongly ($\rho{=}0.87$). The Image Heads and Localization Heads selectors track it only weakly, and random non-gaze heads are slightly anti-correlated; without a working lever, the model falls back to its default left-to-right scan, which a derangement schedule is built to oppose. \cref{fig:dynamic_switch} shows a single run. The model keeps its usual ``Panel 1, 2, 3\ldots'' numbering but describes the content of whichever panel the gaze heads are pointed at, wrapping up one description and opening the next at each switch. The numbering follows the model's default textual structure while the visual content follows the gaze heads, suggesting the two mechanisms are functionally separate. A schedule-blind trajectory judge confirms that gaze steering does not merely disrupt the default order but replaces it (\cref{app:dynamic_traj_judge}). Try this yourself at \href{https://gaze.baulab.info/#demo}{\textcolor[rgb]{0.21,0.49,0.74}{gaze.baulab.info/\#demo}} (needs latest Chrome or Firefox).

\subsection{How Many Heads Are Enough?}
\label{sec:saturation}

Every result so far redirects a fixed set of 100 heads. Is that number special, or would fewer do? \cref{fig:topk_saturation} sweeps the number of redirected heads. VQA redirection accuracy climbs from $36\%$ at $K{=}5$ heads to a peak of $83.1\%$ at $K{=}100$, then declines gracefully as the intervention starts to override heads the model needs for fluent generation. The gaze function is thus concentrated in roughly the top-100 heads: enough to seize control of visual grounding, few enough that the rest of the model keeps working. The random non-gaze control fails to redirect across the whole sweep.

\subsection{Gaze Heads on Natural Images}
\label{sec:natural_images}

\begin{figure*}
  \centering
  \includegraphics[width=\linewidth]{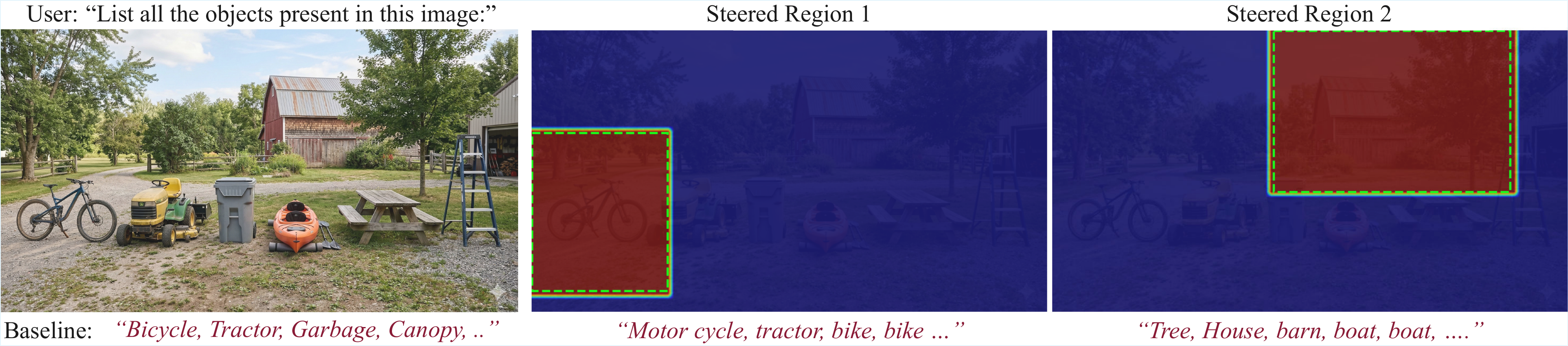}
  \caption{Gaze-head steering on a natural image. Left: the original image, with the baseline response listing objects across the whole scene. Middle and right: steering the gaze heads to a chosen region restricts the response to objects within that region.}
  \label{fig:steer_natural_main}
\end{figure*}

Comics give clean panel boundaries; natural images do not. Do gaze heads still ground attention spatially when no explicit regions exist? We prompt the model to describe a natural image and record gaze-head attention per output span. Attention shifts to the spatial region of each object as the model describes it (\cref{fig:natural_image_gaze}): upper-left for an easel, center for a plant, upper-right for a globe. Steering also works with natural images: concentrating gaze-head attention on a chosen region makes the model describe objects in that region only (\cref{fig:steer_natural_main}).

To quantify this, we redirect gaze heads on COCO val2017~\cite{lin2014microsoft} images, steering them to a target object's bounding box and asking what object is in that region; an LLM judge checks whether the answer names the target COCO category (\cref{app:coco_natural_vqa}). \cref{tab:coco_main} reports the result: gaze redirection more than doubles the non-gaze control in every object-size class, confirming that the heads found through comic probing also steer the model toward an arbitrary region of a natural image. The intervention works best on larger objects, whose bounding boxes span enough image tokens for the bias to bite, and weakens on small ones.

\begin{table}[t]
\centering
\small
\caption{Gaze-head steering on COCO val2017~\cite{lin2014microsoft}. Redirection accuracy when the top-ranked gaze heads are steered to a target object's bounding box, by COCO object-size class, with bootstrap 95\% CIs. An LLM judge checks whether the steered answer names the target category.}
\label{tab:coco_main}
\begin{tabular}{lcc}
\toprule
Size class (COCO) & Gaze heads & Non-gaze \\
\midrule
Large  ($>96^2$\,px) & 80.3 $\pm$ 1.2\,\% & 36.6 $\pm$ 1.0\,\% \\
Medium ($<96^2$\,px) & 76.2 $\pm$ 1.1\,\% & 19.4 $\pm$ 0.7\,\% \\
Small ($27^2$--$32^2$\,px) & 61.7 $\pm$ 4.2\,\% & 18.6 $\pm$ 4.0\,\% \\
\addlinespace[3pt]
\hdashline
\addlinespace[3pt]
\textbf{Overall}              & \textbf{76.5 }$\pm$ \textbf{0.8}\,\% & \textbf{25.9} $\pm$ \textbf{0.6}\,\% \\
\bottomrule
\end{tabular}
\end{table}

\subsection{Generalization Across Models}
\label{sec:generalization}

\emph{Gaze heads} are not a Qwen3-VL idiosyncrasy. We apply the same pipeline to four Qwen3-VL sizes from 2B to 32B parameters and to six other VLMs spanning different vision encoders, tokenizers, and alignment recipes: Qwen2-VL~\cite{wang2024qwen2}, Ovis1.5~\cite{lu2024ovis}, InternVL3.5~\cite{wang2025internvl3}, LLaVA-1.5~\cite{liu2024improved}, LLaVA-NeXT~\cite{liu2024llavanext}, and Bunny-3B~\cite{he2024bunny}. We provide implementation details in \cref{app:cross_arch_preproc}.

\cref{tab:cross_arch} reports peak redirection across the four Qwen3-VL sizes and the six other architectures. The mechanism transfers cleanly to Ovis1.5, Qwen2-VL, and InternVL3.5, where the gaze head set redirects the answer well above the non-gaze and all-heads controls. Qwen3-VL-8B is the strongest at $83.1\%$, and the other working models land in the $60$--$70\%$ range. The LLaVA model family and Bunny-3B show no comparable gaze mechanism. One pattern consistent with the split, which we present as a hypothesis rather than a confirmed cause, is whether the vision encoder is trained with the LM: all three families above $60\%$ fine-tune their encoder on the VLM task, while the three that plateau or fail (both LLaVAs and Bunny) keep a frozen CLIP or SigLIP encoder behind a thin MLP. Bunny offers a particularly suggestive same-backbone comparison: it freezes the same SigLIP-so400m backbone that Ovis fine-tunes, yet yields $8.3\%$ peak gaze accuracy versus Ovis's $68.7\%$. We treat this as evidence consistent with the hypothesis above, not as proof; a fully controlled comparison is left to future work (\cref{app:generalization_all}). Full $K$-sweeps and within-family scale comparisons are in \cref{tab:gaze_model_sizes,tab:cross_arch_scale}.

\begin{table}[t]
\centering
\small
\caption{Peak gaze-redirection accuracy on Qwen3-VL sizes (top block) and on other VLMs (bottom block). \emph{Peak $K$} is the number of redirected heads at the per-model accuracy peak, and \emph{All-heads} runs the same intervention on every head. The mechanism transfers cleanly to every trained-encoder model we tested ($60$--$83\%$ peak); LLaVA and Bunny (frozen-encoder) are the exceptions.}
\label{tab:cross_arch}
\resizebox{\columnwidth}{!}{%
\begin{tabular}{lcccc}
\toprule
Model & Peak $K$ & Gaze & Non-Gaze & All-Heads \\
\midrule
Qwen3-VL-2B  & 10  & 68.6 & 1.3  & 0.0  \\
Qwen3-VL-4B  & 75  & 72.9 & 1.5  & 0.0  \\
Qwen3-VL-8B  & 100 & 83.1 & 14.6 & 0.9 \\
Qwen3-VL-32B & 500 & 70.2 & 12.0 & 0.2  \\
\addlinespace[3pt]
\hdashline
\addlinespace[3pt]
Ovis1.5-8B      & 100 & 68.7 & 13.0 & 2.7 \\
Qwen2-VL-7B     & 90  & 66.2 & 0.0  & 0.8 \\
InternVL3.5-8B  & 140 & 62.7 & 31.0 & 0.8 \\
LLaVA-1.5-13B   & 160 & 39.0 & 13.8 & 0.0 \\
LLaVA-NeXT-7B   & 100 & 35.3 & 26.7 & 0.0 \\
Bunny-3B        & 10  & 8.3  & 0.0  & 0.0 \\
\bottomrule
\end{tabular}
}
\end{table}

Across every trained-encoder VLM we tested, the same correlation score recovers a \emph{gaze head} set that can be steered with a single attention-mask edit. Where this picture breaks, our analysis points toward the vision encoder, though confirming this would require controlled experiments we leave to future work.

\section{Conclusion}
\label{sec:conclusion}
\emph{Gaze heads} are a causal control surface: a small head set through which a vision-language model couples what it looks at to what it says. We identify them with a simple correlation score, and a single attention-mask intervention on just the top 100 redirects what the model describes to any chosen image region, with no retraining. Where prior work treats image-attending heads only as a signal source for contrastive decoding or localization readout, \emph{gaze heads} are a causal control surface, and changing where they look changes what the model says.

The mechanism recurs across model sizes, multiple architectures, and natural images, but it is not universal: frozen-encoder families show no comparable gaze head set, suggesting the mechanism depends on training the vision encoder together with the language model. What makes some architectures amenable to \emph{gaze head} formation, and whether the same heads mediate other visually grounded behaviors such as spatial reasoning and hallucination, are open questions we hope this work helps frame.
\section*{Acknowledgments}
RG and DB are supported by Open Philanthropy and NSF grant \#2403304.

\section*{Code}
Source code, demo, and datasets for reproducing our results can be found at \href{https://gaze.baulab.info/}{\textcolor[rgb]{0.21,0.49,0.74}{gaze.baulab.info}} and at our GitHub repo \href{https://github.com/rohitgandikota/gaze-heads/}{\textcolor[rgb]{0.21,0.49,0.74}{github.com/rohitgandikota/gaze-heads}}.

{
    \small
    \bibliographystyle{ieeenat_fullname}
    \bibliography{main}

@inproceedings{iyyer2017amazing,
  title={The amazing mysteries of the gutter: Drawing inferences between panels in comic book narratives},
  author={Iyyer, Mohit and Manjunatha, Varun and Guha, Anupam and Vyas, Yogarshi and Boyd-Graber, Jordan and Daume, Hal and Davis, Larry S},
  booktitle={Proceedings of the IEEE Conference on Computer Vision and Pattern recognition},
  pages={7186--7195},
  year={2017}
}

@article{chaudhary2025spatial,
  title={Investigating Spatial Attention Bias in Vision-Language Models},
  author={Chaudhary, Aryan and Goyal, Sanchit and Narang, Pratik and Kumar, Dhruv},
  journal={arXiv preprint arXiv:2512.18231},
  year={2025}
}

@article{qwen3vl,
  title={Qwen3-vl technical report},
  author={Bai, Shuai and Cai, Yuxuan and Chen, Ruizhe and Chen, Keqin and Chen, Xionghui and Cheng, Zesen and Deng, Lianghao and Ding, Wei and Gao, Chang and Ge, Chunjiang and others},
  journal={arXiv preprint arXiv:2511.21631},
  year={2025}
}

@article{chen2024premise,
  title={Premise order matters in reasoning with large language models},
  author={Chen, Xinyun and Chi, Ryan A and Wang, Xuezhi and Zhou, Denny},
  journal={arXiv preprint arXiv:2402.08939},
  year={2024}
}

@article{aizawa2020manga109,
  title={Building a manga dataset “manga109” with annotations for multimedia applications},
  author={Aizawa, Kiyoharu and Fujimoto, Azuma and Otsubo, Atsushi and Ogawa, Toru and Matsui, Yusuke and Tsubota, Koki and Ikuta, Hikaru},
  journal={IEEE multimedia},
  volume={27},
  number={2},
  pages={8--18},
  year={2020},
  publisher={IEEE}
}

@article{vivoli2024comix,
  title={Comix: A comprehensive benchmark for multi-task comic understanding},
  author={Vivoli, Emanuele and Bertini, Marco and Karatzas, Dimosthenis},
  journal={Advances in Neural Information Processing Systems},
  volume={37},
  pages={140828--140846},
  year={2024}
}

@article{vivoli2024survey,
  title={One missing piece in vision and language: A survey on comics understanding},
  author={Vivoli, Emanuele and Souibgui, Mohamed Ali and Barsky, Andrey and LLabres, Artemis and Bertini, Marco and Karatzas, Dimosthenis},
  journal={arXiv preprint arXiv:2409.09502},
  year={2024}
}

@inproceedings{sachdeva2024manga,
  title={The manga whisperer: Automatically generating transcriptions for comics},
  author={Sachdeva, Ragav and Zisserman, Andrew},
  booktitle={Proceedings of the IEEE/CVF Conference on Computer Vision and Pattern Recognition},
  pages={12967--12976},
  year={2024}
}

@inproceedings{sachdeva2025panels,
  title={From panels to prose: Generating literary narratives from comics},
  author={Sachdeva, Ragav and Zisserman, Andrew},
  booktitle={Proceedings of the IEEE/CVF International Conference on Computer Vision},
  pages={21864--21873},
  year={2025}
}

@inproceedings{chen2024spatialvlm,
  title={Spatialvlm: Endowing vision-language models with spatial reasoning capabilities},
  author={Chen, Boyuan and Xu, Zhuo and Kirmani, Sean and Ichter, Brain and Sadigh, Dorsa and Guibas, Leonidas and Xia, Fei},
  booktitle={Proceedings of the IEEE/CVF Conference on Computer Vision and Pattern Recognition},
  pages={14455--14465},
  year={2024}
}

@inproceedings{lee2025perspective,
  title={Perspective-aware reasoning in vision-language models via mental imagery simulation},
  author={Lee, Phillip Y and Je, Jihyeon and Park, Chanho and Uy, Mikaela Angelina and Guibas, Leonidas and Sung, Minhyuk},
  booktitle={Proceedings of the IEEE/CVF International Conference on Computer Vision},
  pages={9241--9251},
  year={2025}
}

@article{elhage2021mathematical,
  title={A mathematical framework for transformer circuits},
  author={Elhage, Nelson and Nanda, Neel and Olsson, Catherine and Henighan, Tom and Joseph, Nicholas and Mann, Ben and Askell, Amanda and Bai, Yuntao and Chen, Anna and Conerly, Tom and others},
  journal={Transformer Circuits Thread},
  volume={1},
  number={1},
  pages={12},
  year={2021}
}

@article{conmy2023automated,
  title={Towards automated circuit discovery for mechanistic interpretability},
  author={Conmy, Arthur and Mavor-Parker, Augustine and Lynch, Aengus and Heimersheim, Stefan and Garriga-Alonso, Adri{\`a}},
  journal={Advances in Neural Information Processing Systems},
  volume={36},
  pages={16318--16352},
  year={2023}
}

@article{golovanevsky2025vlms,
  title={What do vlms notice? a mechanistic interpretability pipeline for gaussian-noise-free text-image corruption and evaluation},
  author={Golovanevsky, Michal and Rudman, William and Palit, Vedant and Singh, Ritambhara and Eickhoff, Carsten},
  journal={arXiv preprint arXiv:2406.16320},
  year={2024}
}

@inproceedings{cohen2025performance,
  title={Performance gap in entity knowledge extraction across modalities in vision language models},
  author={Cohen, Ido and Gottesman, Daniela and Geva, Mor and Giryes, Raja},
  booktitle={Proceedings of the 63rd Annual Meeting of the Association for Computational Linguistics (Volume 1: Long Papers)},
  pages={29095--29108},
  year={2025}
}

@article{chen2025spatial,
  title={Why is spatial reasoning hard for vlms? an attention mechanism perspective on focus areas},
  author={Chen, Shiqi and Zhu, Tongyao and Zhou, Ruochen and Zhang, Jinghan and Gao, Siyang and Niebles, Juan Carlos and Geva, Mor and He, Junxian and Wu, Jiajun and Li, Manling},
  journal={arXiv preprint arXiv:2503.01773},
  year={2025}
}

@article{yan2024voila,
  title={Voila-a: Aligning vision-language models with user's gaze attention},
  author={Yan, Kun and Wang, Zeyu and Ji, Lei and Wang, Yuntao and Duan, Nan and Ma, Shuai},
  journal={Advances in neural information processing systems},
  volume={37},
  pages={1890--1918},
  year={2024}
}

@article{pani2025gaze,
  title={Gaze-vlm: Bridging gaze and vlms through attention regularization for egocentric understanding},
  author={Pani, Anupam and Yang, Yanchao},
  journal={Advances in Neural Information Processing Systems},
  volume={38},
  pages={163544--163577},
  year={2026}
}

@inproceedings{neo2024towards,
  title={Towards interpreting visual information processing in vision-language models},
  author={Neo, Clement and Ong, Luke and Torr, Philip and Geva, Mor and Krueger, David and Barez, Fazl},
  booktitle={International Conference on Learning Representations},
  volume={2025},
  pages={57172--57189},
  year={2025}
}

@article{olsson2022context,
  title={In-context learning and induction heads},
  author={Olsson, Catherine and Elhage, Nelson and Nanda, Neel and Joseph, Nicholas and DasSarma, Nova and Henighan, Tom and Mann, Ben and Askell, Amanda and Bai, Yuntao and Chen, Anna and others},
  journal={arXiv preprint arXiv:2209.11895},
  year={2022}
}

@article{wang2022interpretability,
  title={Interpretability in the wild: a circuit for indirect object identification in gpt-2 small},
  author={Wang, Kevin and Variengien, Alexandre and Conmy, Arthur and Shlegeris, Buck and Steinhardt, Jacob},
  journal={arXiv preprint arXiv:2211.00593},
  year={2022}
}

@inproceedings{gandelsman2023interpreting,
  title={Interpreting clip's image representation via text-based decomposition},
  author={Gandelsman, Yossi and Efros, Alexei and Steinhardt, Jacob},
  booktitle={International Conference on Learning Representations},
  volume={2024},
  pages={18395--18416},
  year={2024}
}

@article{basu2024understanding,
  title={Understanding information storage and transfer in multi-modal large language models},
  author={Basu, Samyadeep and Grayson, Martin and Morrison, Cecily and Nushi, Besmira and Feizi, Soheil and Massiceti, Daniela},
  journal={Advances in Neural Information Processing Systems},
  volume={37},
  pages={7400--7426},
  year={2024}
}

@article{assouel2025visual,
  title={Visual symbolic mechanisms: Emergent symbol processing in vision language models},
  author={Assouel, Rim and Campbell, Declan and Bengio, Yoshua and Webb, Taylor},
  journal={arXiv preprint arXiv:2506.15871},
  year={2025}
}

@misc{nanobananapro_2025,
  author = {{Google DeepMind}},
  title = {Nano Banana Pro (Gemini 3 Pro Image)},
  year = {2025},
  month = {November},
  url = {https://blog.google/innovation-and-ai/products/nano-banana-pro/},
}

@misc{anthropic2025claude,
  author = {{Anthropic}},
  title = {Claude-4.6 Sonnet},
  year = {2026},
  url = {https://claude.ai},
}

@article{turner2023activation,
  title={Steering language models with activation engineering},
  author={Turner, Alexander Matt and Thiergart, Lisa and Leech, Gavin and Udell, David and Vazquez, Juan J and Mini, Ulisse and MacDiarmid, Monte},
  journal={arXiv preprint arXiv:2308.10248},
  year={2023}
}

@article{zou2023representation,
  title={Representation engineering: A top-down approach to ai transparency},
  author={Zou, Andy and Phan, Long and Chen, Sarah and Campbell, James and Guo, Phillip and Ren, Richard and Pan, Alexander and Yin, Xuwang and Mazeika, Mantas and Dombrowski, Ann-Kathrin and others},
  journal={arXiv preprint arXiv:2310.01405},
  year={2023}
}

@inproceedings{kobayashi2020attention,
  title={Attention is not only a weight: Analyzing transformers with vector norms},
  author={Kobayashi, Goro and Kuribayashi, Tatsuki and Yokoi, Sho and Inui, Kentaro},
  booktitle={Proceedings of the 2020 Conference on Empirical Methods in Natural Language Processing (EMNLP)},
  pages={7057--7075},
  year={2020}
}

@article{wang2025internvl3,
  title={Internvl3. 5: Advancing open-source multimodal models in versatility, reasoning, and efficiency},
  author={Wang, Weiyun and Gao, Zhangwei and Gu, Lixin and Pu, Hengjun and Cui, Long and Wei, Xingguang and Liu, Zhaoyang and Jing, Linglin and Ye, Shenglong and Shao, Jie and others},
  journal={arXiv preprint arXiv:2508.18265},
  year={2025}
}

@article{deng2025maskcd,
  title={Maskcd: Mitigating lvlm hallucinations by image head masked contrastive decoding},
  author={Deng, Jingyuan and Yang, Yujiu},
  journal={arXiv preprint arXiv:2510.02790},
  year={2025}
}

@inproceedings{kang2025your,
  title={Your large vision-language model only needs a few attention heads for visual grounding},
  author={Kang, Seil and Kim, Jinyeong and Kim, Junhyeok and Hwang, Seong Jae},
  booktitle={Proceedings of the Computer Vision and Pattern Recognition Conference},
  pages={9339--9350},
  year={2025}
}

@inproceedings{liu2024improved,
  title={Improved baselines with visual instruction tuning},
  author={Liu, Haotian and Li, Chunyuan and Li, Yuheng and Lee, Yong Jae},
  booktitle={Proceedings of the IEEE/CVF conference on computer vision and pattern recognition},
  pages={26296--26306},
  year={2024}
}

@article{wang2024qwen2,
  title={Qwen2-vl: Enhancing vision-language model's perception of the world at any resolution},
  author={Wang, Peng and Bai, Shuai and Tan, Sinan and Wang, Shijie and Fan, Zhihao and Bai, Jinze and Chen, Keqin and Liu, Xuejing and Wang, Jialin and Ge, Wenbin and others},
  journal={arXiv preprint arXiv:2409.12191},
  year={2024}
}

@article{lu2024ovis,
  title={Ovis: Structural embedding alignment for multimodal large language model},
  author={Lu, Shiyin and Li, Yang and Chen, Qing-Guo and Xu, Zhao and Luo, Weihua and Zhang, Kaifu and Ye, Han-Jia},
  journal={arXiv preprint arXiv:2405.20797},
  year={2024}
}

@misc{liu2024llavanext,
    title={LLaVA-NeXT: Improved reasoning, OCR, and world knowledge},
    url={https://llava-vl.github.io/blog/2024-01-30-llava-next/},
    author={Liu, Haotian and Li, Chunyuan and Li, Yuheng and Li, Bo and Zhang, Yuanhan and Shen, Sheng and Lee, Yong Jae},
    month={January},
    year={2024}
}

@article{michel2019sixteen,
  title={Are sixteen heads really better than one?},
  author={Michel, Paul and Levy, Omer and Neubig, Graham},
  journal={Advances in neural information processing systems},
  volume={32},
  year={2019}
}

@inproceedings{voita2019analyzing,
  title={Analyzing multi-head self-attention: Specialized heads do the heavy lifting, the rest can be pruned},
  author={Voita, Elena and Talbot, David and Moiseev, Fedor and Sennrich, Rico and Titov, Ivan},
  booktitle={Proceedings of the 57th annual meeting of the association for computational linguistics},
  pages={5797--5808},
  year={2019}
}

@inproceedings{lin2014microsoft,
  title={Microsoft coco: Common objects in context},
  author={Lin, Tsung-Yi and Maire, Michael and Belongie, Serge and Hays, James and Perona, Pietro and Ramanan, Deva and Doll{\'a}r, Piotr and Zitnick, C Lawrence},
  booktitle={European conference on computer vision},
  pages={740--755},
  year={2014},
  organization={Springer}
}

@article{he2024bunny,
  title={Efficient multimodal learning from data-centric perspective},
  author={He, Muyang and Liu, Yexin and Wu, Boya and Yuan, Jianhao and Wang, Yueze and Huang, Tiejun and Zhao, Bo},
  journal={arXiv preprint arXiv:2402.11530},
  year={2024}
}

@misc{gptimage_2025,
  author = {OpenAI},
  title = {GPT Image 1 Model Card},
  year = {2025},
  month = {March},
  url = {https://developers.openai.com/api/docs/models/gpt-image-1},
}
}
\clearpage
\appendix
\setcounter{page}{1}
\maketitlesupplementary

% =============================================================================
\section{Experimental Details}
\label{app:details}

\paragraph{Hardware.}
All experiments run on a single NVIDIA RTX A6000 48\,GB GPU in \texttt{bfloat16} precision.

\paragraph{Sample sizes.}
Gaze-head discovery collects per-head statistics over 500 strips. All redirection and narration results are validated on a held-out set of 500 six-panel comic strips generated with GPT Image (\cref{app:openai_dataset}), with each of the 6 panels targeted in turn for $3{,}000$ (strip, target) pairs. We report 95\% bootstrap confidence intervals (10{,}000 resamples) for all primary metrics.

\paragraph{Hyperparameters.}
We fix one configuration throughout. The residual-stream steering scale is $\alpha = 1.0$. Gaze-head discovery retains the top 100 heads by gaze score as candidates. Gaze-head redirection sets the attention-mask bias to $\delta = +\infty$, applied at both prefill and decode tokens for VQA and static narration, and at decode tokens only for dynamic narration (where the target switches mid-generation); this hard-reassigns each head's image-attention onto the target panel and zeros it everywhere else. Smaller $\delta$ values produce softer reassignment; we use the saturation limit throughout for a clean and reproducible intervention. Dynamic gaze-steered narration redirects the top 100 heads and switches target panels every $T = 50$ generated tokens. All generation uses greedy decoding (\texttt{do\_sample=False}), with \texttt{max\_new\_tokens=15} for the brief VQA prompt (``Answer briefly''), \texttt{max\_new\_tokens=100} for static narration, and \texttt{max\_new\_tokens=300} for dynamic narration (six $50$-token segments).

\paragraph{Comic strip details.}
Strips consist of 6 panels by default, resized to a common height of 256 pixels and concatenated horizontally.
Each panel is overlaid with a random letter label drawn uniformly without replacement from A--Z.
For unlabeled experiments, panels are concatenated without any overlay.

%\clearpage
\paragraph{Custom comic panel generation.}
For qualitative examples on visually diverse content, we also generate comic strips of 6 panels each using Google's Nano Banana Pro~\cite{nanobananapro_2025}. Each strip is produced with the prompt: \textit{``Please generate a 6 panel comic strip with a smooth narrative story. Each panel shows a unique action, background, or object. Please provide each individual panel image separately.''}
This ensures that each panel within a strip contains distinct visual content, making it possible to unambiguously verify which panel the model is grounding its response in.

\paragraph{Replication dataset.}
\label{app:openai_dataset}
The 500-strip validation set is generated end-to-end with OpenAI: the panel descriptions are written by \texttt{gpt-4o-mini} and rendered by \texttt{gpt-image-1} at $1024{\times}1024$. The story-writing system prompt enforces six rules:
\begin{enumerate}
\item one consistent protagonist and art style across the 6 panels;
\item every panel shows a clearly different action in a distinct setting with different salient objects;
\item no text, captions, or speech bubbles in the panels;
\item each panel description is rich enough to render unambiguously;
\item no location, object, or action repeats across panels;
\item safe content (no violence, weapons, romantic content, etc., to avoid moderation rejections).
\end{enumerate}
The image-rendering prompt for each panel restates the protagonist and style and reiterates the no-text rule. The full generator script (with retries, panel-level on-disk caching for resumable runs, and a small set of hand-written safe-themed replacement stories used to backfill the few moderation rejections) is released with the code, so anyone can re-build the dataset from scratch.

\paragraph{Evaluation metrics.}
For forced-choice probes, we report accuracy (fraction of correct panel label predictions).
For free-form narration, we extract the order in which panel labels (or positional references, for unlabeled strips) appear in the generated text and compute the Spearman rank correlation $\rho$ against the target ordering.
We also report the ``starts-correct'' rate: the fraction of narrations whose first panel mention matches the target first panel.
For visual question answering and narration redirection, we use Claude Sonnet~\cite{anthropic2025claude} as an LLM judge. We report 95\% bootstrap confidence intervals (10{,}000 resamples) for all primary metrics.

\paragraph{LLM judge: forced-choice panel match.}
\label{app:judge_match}
For panel-redirection accuracy on comic strips, the judge is a forced 1-of-6 choice: given the strip image and the steered answer, Claude picks the single panel whose visual content the answer best describes. The verdict is HIT iff the matched panel equals the target. The judge prompt is:
\begin{quote}\small
\textit{This is a 6-panel comic strip (panels numbered 1 to 6 from left to right). Steered answer: ``\{steered\}''.
Ignore any panel numbering inside the steered answer (`Panel 1:', `Panel 2:'); the model often numbers sequentially regardless of which panel it is actually describing. Match by visual content. Pick exactly one panel (an integer 1..6) whose visual content the steered answer best describes. If the answer is incoherent, repetitive, degenerate, empty, or just numbers/labels, set is\_junk=true and matched\_panel=null. Return ONLY a JSON object: \{``matched\_panel'': $<$1..6 or null$>$, ``is\_junk'': $<$true/false$>$\}.}
\end{quote}
Junk and unmatchable outputs count as misses, so the denominator is always the total number of (strip, target) pairs and the chance baseline is exactly $1/6$. To prevent control conditions from inflating when the steered answer is essentially identical to the unsteered baseline, we mark such pairs as misses without a judge call: specifically, if the steered and baseline answers have token-level Jaccard similarity above $0.9$, we record HIT$=$false directly. This leaves genuinely steered outputs unaffected.

\paragraph{LLM judge: object match for natural-image VQA.}
\label{app:judge_object}
For COCO val2017 natural-image VQA, we ask the model \textit{``What is the main object in this part of the image? Answer in a few words.''} while steering gaze heads to a specific object's bounding-box region. The judge sees the steered answer and the list of COCO categories present in the image, and is prompted:
\begin{quote}\small
\textit{The image contains these objects: \{label list\}. The VLM's attention was steered toward a region containing: \{target label\}. The VLM responded: ``\{steered\}''. Does the response describe or refer to the target object? Consider synonyms (e.g., ``car''$\sim$``automobile''). Return ONLY a JSON object: \{``match'': $<$true/false$>$, ``predicted\_label'': ``$<$best matching object$>$''\}.}
\end{quote}
Accuracy is the fraction of \texttt{match=true} judgments, broken down by COCO size class.

\paragraph{Random non-gaze sampling and API retries.}
\label{app:nongaze_sampling}
The random non-gaze control samples $K$ (layer, head) pairs uniformly at random from layers $20$--$35$ of the model, excluding any head that belongs to the gaze head set. This matches the gaze heads in layer range while ensuring the sampled heads are not themselves gaze heads. \cref{tab:nongaze_ablation} reports alternative percentile-based sampling choices for comparison; all stay well below the gaze condition. All judge calls are wrapped in an exponential-backoff retry loop (up to 6 attempts, base delay 2\,s, doubling) for transient API errors so that no samples are silently dropped from the denominator.

\begin{table}[t]
\centering
\small
\caption{Non-gaze control sampling choice. VQA accuracy on 500 strips ($n{=}3{,}000$) for alternative non-gaze sampling cutoffs. Headline runs in the paper sample non-gaze heads uniformly at random from layers $20$--$35$, excluding the gaze head set. Gaze accuracy is essentially unchanged across choices; the non-gaze (control) accuracy is what shifts.}
\label{tab:nongaze_ablation}
\begin{tabular}{lccc}
\toprule
Non-gaze pool & Gaze & Non-gaze & All-heads \\
\midrule
below 50\% (median) & 83.1 & 38.6 & 0.9 \\
below 5\% & 83.1 & 14.7 & 0.9 \\
layers 20--35 (used)  & 83.1 & 14.6 & 0.9 \\
\bottomrule
\end{tabular}
\end{table}

% =============================================================================
\section{Layer-Level Steering and Position Representations}
\label{app:layer_extended}

\subsection{Free-Form Narration via Layer Steering}
\label{app:layer_freeform}
Layer steering achieves near-perfect binary control on the forced-choice probe, but probes are artificial. To test whether the same direction governs open-ended behavior, we steer layers 20--28 simultaneously for every generated token and let the model generate a free-form narration (prompted with ``Please describe what happens in each panel, in order.'').

\begin{figure}[t]
  \centering
  \includegraphics[width=\linewidth]{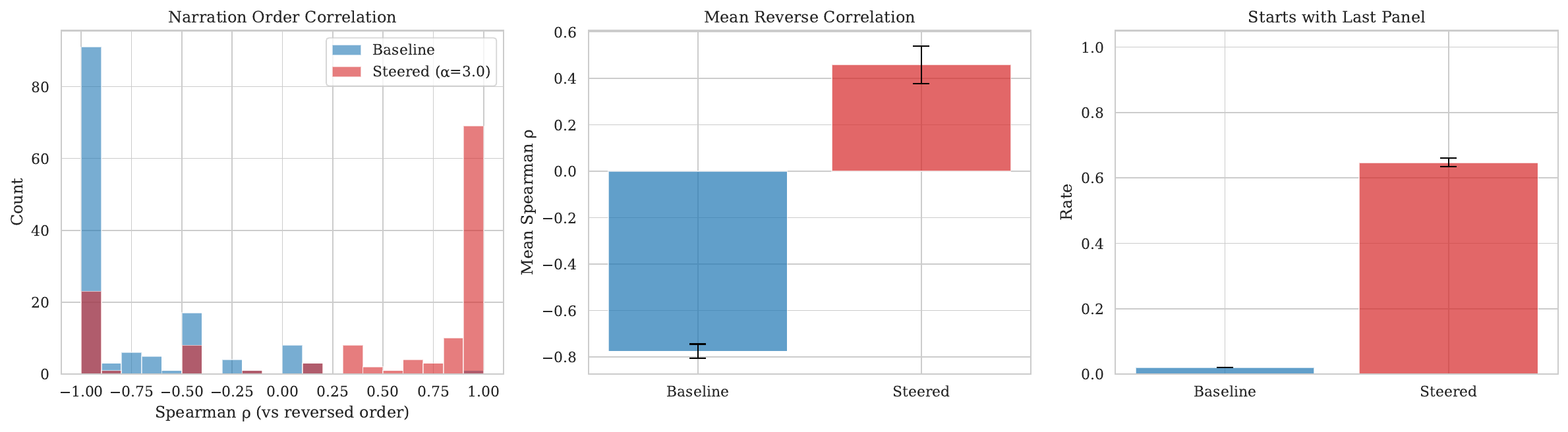}
  \caption{Binary reverse narration via layer steering.
  Steered narrations shift from strong left-to-right order ($\rho \approx -0.8$) to positive right-to-left correlation ($\rho \approx +0.5$), with 65\% starting from the last panel.}
  \label{fig:binary_narration}
\end{figure}

Across 100 test strips and three seeds, the baseline narration produces $\rho = -0.78$ (strong left-to-right ordering; \cref{fig:binary_narration}). Here $\rho$ is the Spearman rank correlation between the order in which panels are mentioned and the target (reversed) ordering.
After steering, $\rho$ rises to $+0.46$, and 65\% of narrations begin with the last panel.
Steered narrations frequently open with phrases like ``reverse order, from right to left'' and proceed to describe panels accordingly, confirming that the model interprets the direction vector as an instruction to reverse its visual attention.

\subsection{Arbitrary Orderings via Prompting vs.\ Steering}
\label{app:arbitrary_orderings}
A natural question is whether the model can follow arbitrary panel orderings, and if so, whether this ability can be extracted as a steering direction. We test both.

\begin{figure}[t]
  \centering
  \begin{subfigure}[t]{0.48\linewidth}
    \centering
    \includegraphics[width=\linewidth]{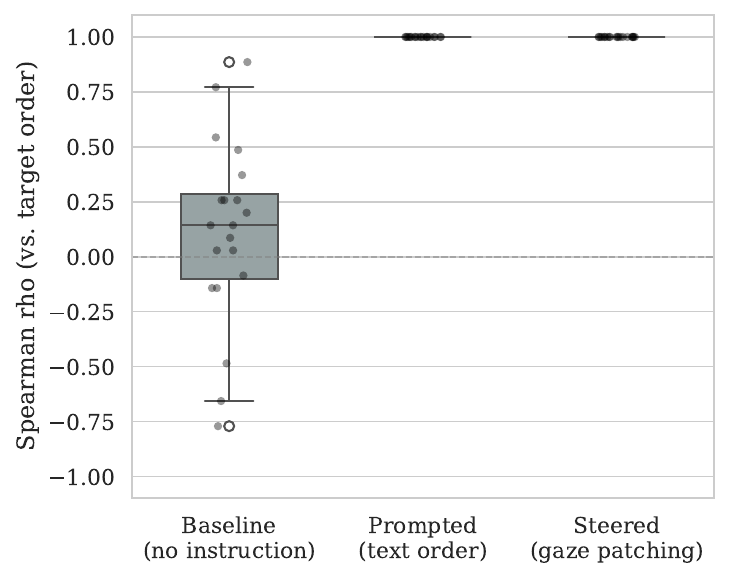}
    \caption{Prompted order following.}
    \label{fig:prompted_order}
  \end{subfigure}
  \hfill
  \begin{subfigure}[t]{0.48\linewidth}
    \centering
    \includegraphics[width=\linewidth]{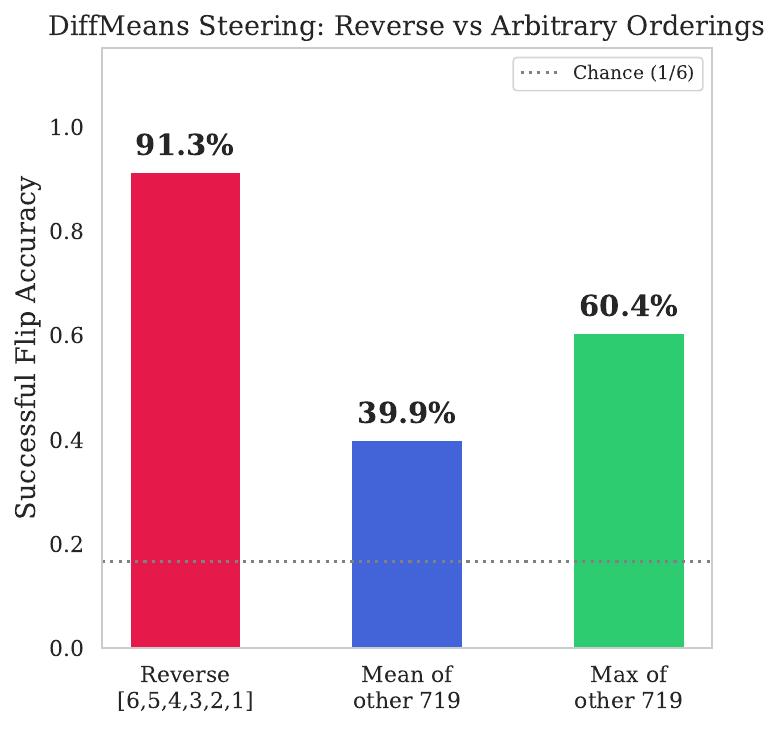}
    \caption{Steering via difference-of-means.}
    \label{fig:permutation_patching}
  \end{subfigure}
  \caption{Arbitrary orderings: prompting vs.\ steering.
  (a)~The model follows arbitrary panel orderings with perfect fidelity when instructed via text ($\rho{=}1.0$ for all 720 permutations); the baseline produces near-zero correlation.
  (b)~Difference-of-means steering only works for the reverse direction ($acc = 91.3\%$); all other 719 permutations produce much weaker steering effect.}
  \label{fig:permutation}
\end{figure}

When prompted in text to narrate in a specific order (e.g., ``Describe the panels in the following order: 4, 2, 6, 1, 3, 5''), the model follows with perfect fidelity. \cref{fig:permutation}a shows the result across all $6!{=}720$ permutations: the model achieves $\rho{=}1.0$ on every ordering, while the baseline (default prompt) produces $\rho{\approx}0.1$ against random targets. The model can clearly solve this task when given explicit text instructions.

However, this ability does not correspond to extractable steering directions. We compute difference-of-means vectors for all 720 permutations and evaluate their steering effectiveness. As \cref{fig:permutation}b shows, only the reverse direction produces meaningful steering ($acc = 91.3\%$); all other permutations yield much weaker effect ($acc \approx 40\%$). This suggests that ``reverse'' is a coherent concept encoded as a single direction in the residual stream, while arbitrary orderings are resolved dynamically during generation, likely by attending back to the prompted sequence tokens after each panel transition rather than through a single representational state.

\subsection{Generalization Across Experimental Setup}
\label{app:generalization}

We test whether the layer-steering mechanism generalizes across panel counts, prompt formulations, and labeling schemes.
All experiments use layer~22 steering with $\alpha = 1.0$ on 500 test strips.

\begin{figure}[t]
  \centering
  \includegraphics[width=0.75\linewidth]{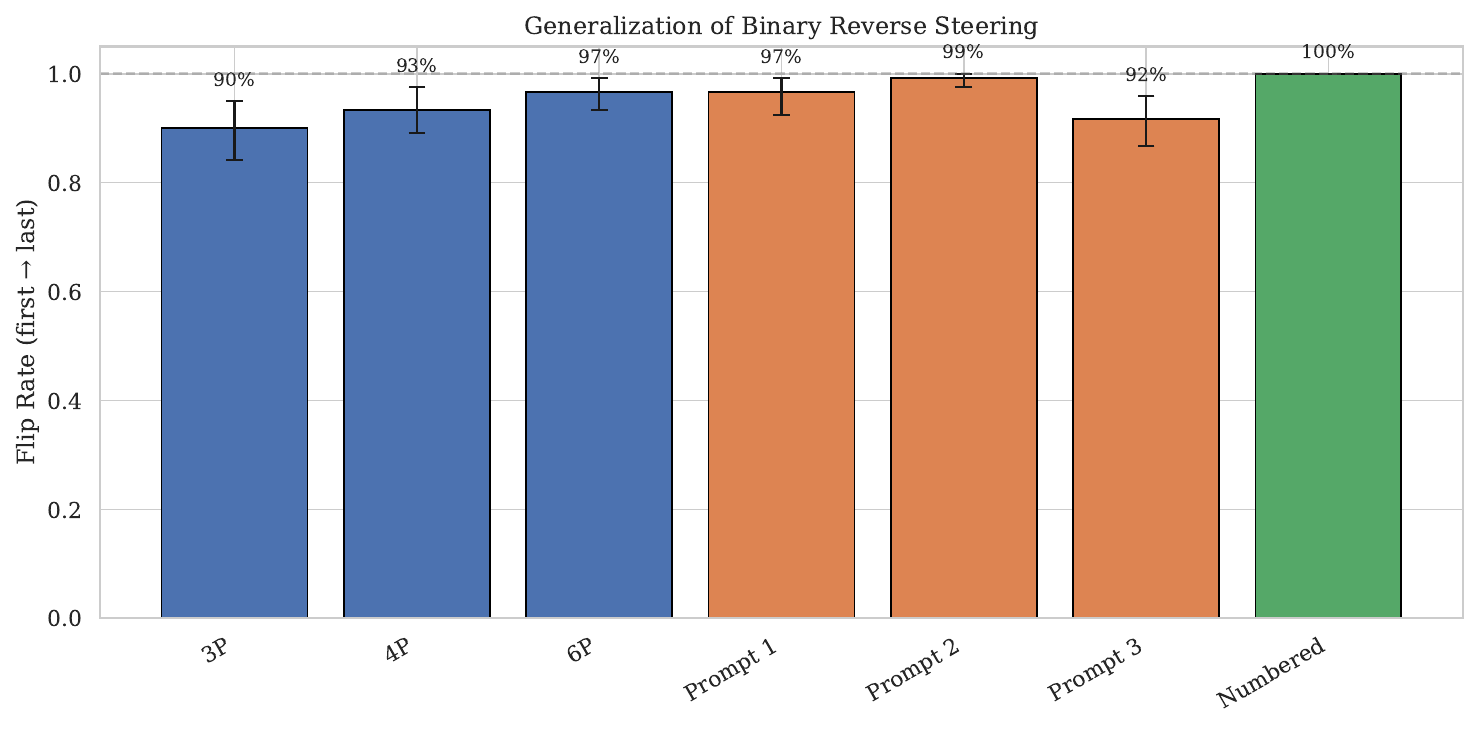}
  \caption{Generalization of layer steering.
  Layer steering transfers across panel counts, prompt formulations, and labeling schemes.
  Bars show flip rates with 95\% bootstrap CIs.}
  \label{fig:generalization}
\end{figure}

\paragraph{Panel counts.}
Flip rates remain high across strip lengths: 90\% for 3 panels, 93\% for 4 panels, and 97\% for 6 panels (\cref{fig:generalization}).
Longer strips are easier to steer, likely because they provide more spatial context for the direction vector.

\paragraph{Prompt variations.}
We test three alternative phrasings of the forced-choice probe.
Flip rates range from 92\% to 99\%, indicating that the learned direction is robust to superficial prompt differences.

\paragraph{Numbered labels.}
When panels are labeled with digits 1--6 instead of random letters, the flip rate reaches 100\%.
This is expected: numeric labels are maximally congruent with the ``$k$-th panel'' prompt structure, eliminating any label-position ambiguity.

\subsection{Position Representations via PCA}
\label{app:pca}

To understand how the model internally represents panel position, we perform PCA on the colon-token activations collected across all six position prompts.

\paragraph{Setup.}
For 500 test strips, we collect hidden-state activations at the colon token under each of the six position prompts, yielding $500 \times 6 = 3000$ activation vectors per layer.
We project these into 2D via PCA and color-code by queried panel index.

\begin{figure}[t]
  \centering
  \includegraphics[width=\linewidth]{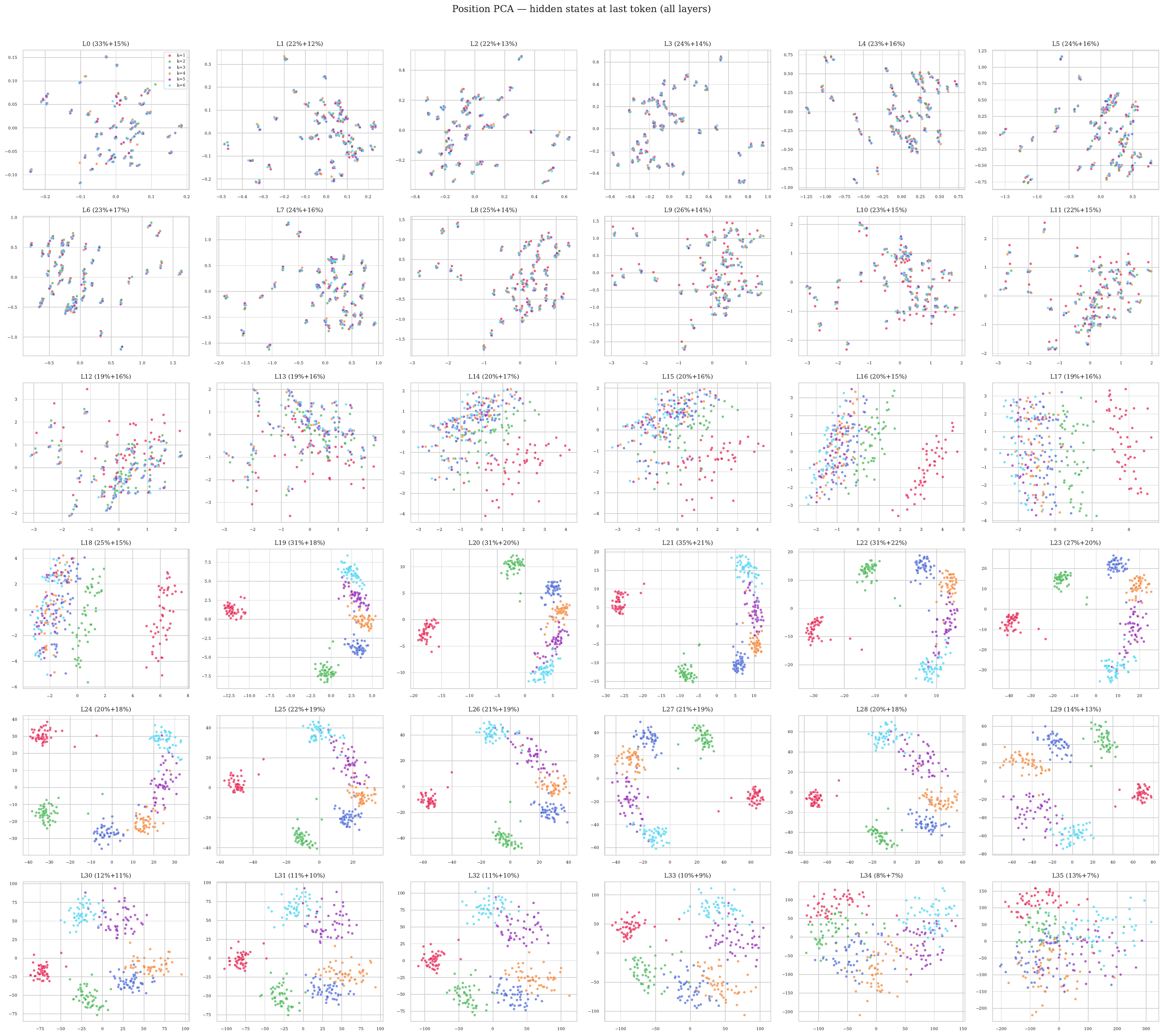}
  \caption{PCA of colon-token activations across all 36 layers.
  Panel-position clusters emerge in the middle layers and dissolve in late layers.}
  \label{fig:pca_grid}
\end{figure}

\paragraph{Results.}
\cref{fig:pca_grid} shows the PCA projections for all 36 layers.
In early layers ($\ell < 15$), all six conditions overlap: the model has not yet differentiated between panel queries.
In middle layers ($\ell \approx 20$--$28$), distinct clusters emerge, with the six panel conditions separating into distinct groups arranged in a spatial gradient from panel~1 on one side to panel~6 on the other.
In late layers ($\ell > 30$), the clusters merge back as the representation converges toward output tokens.

\begin{figure}[t]
  \centering
  \includegraphics[width=0.6\linewidth]{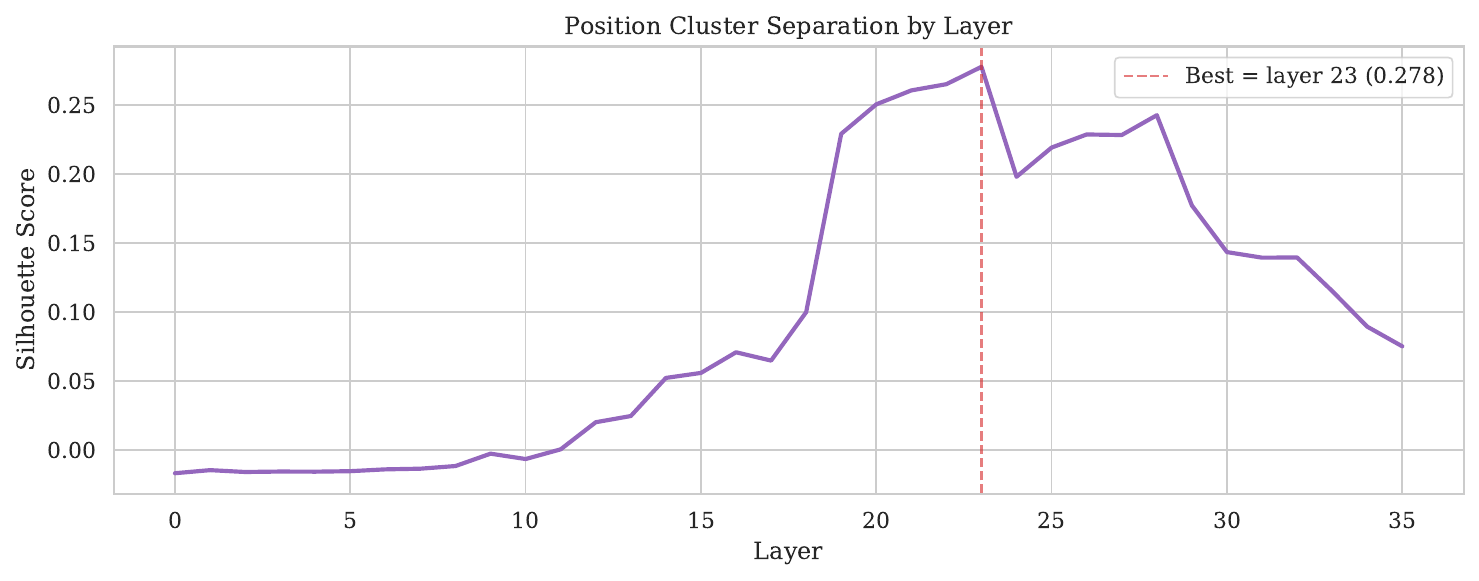}
  \caption{Silhouette score by layer.
  Cluster quality peaks at layer~23, matching the effective steering range.}
  \label{fig:silhouette}
\end{figure}

The silhouette score (\cref{fig:silhouette}) quantifies cluster quality across layers, peaking at layer~23.
This aligns precisely with the layer range identified by the steering experiments in \cref{sec:layer_analysis}, providing converging evidence that the middle layers encode a position-aware representation.

% =============================================================================
\section{Gaze-Head Discovery: Extended Analysis}
\label{app:gaze}

\subsection{Gaze Score Distribution}
\label{app:gaze_distribution}

\begin{figure}[t]
  \centering
  \includegraphics[width=0.7\linewidth]{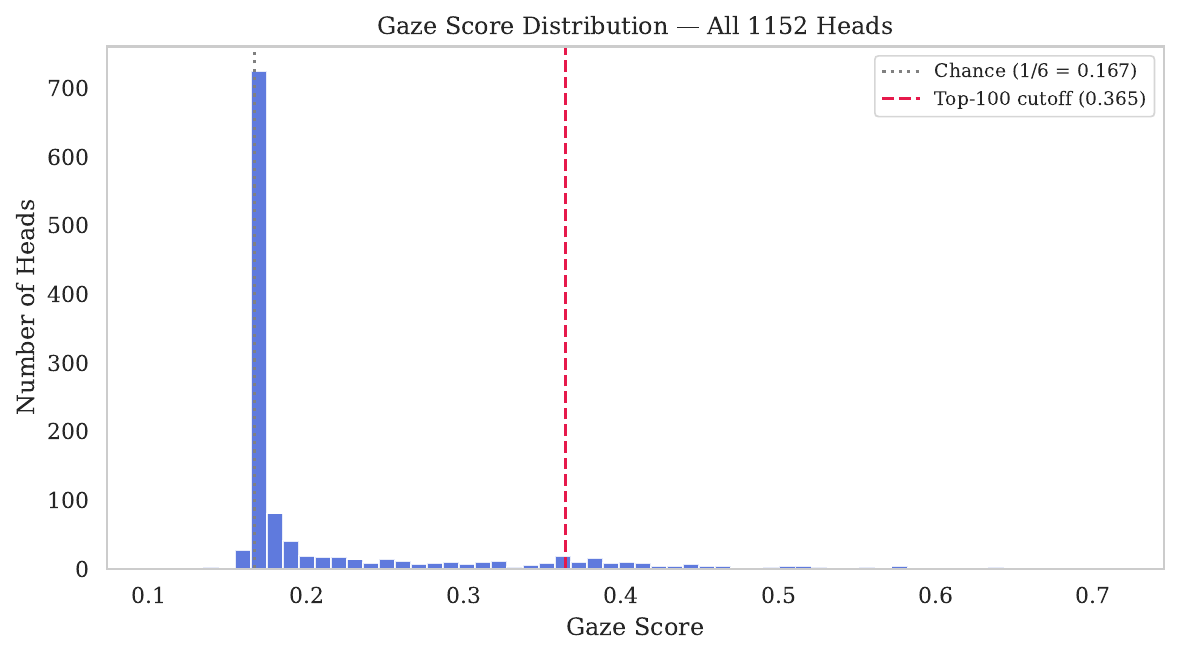}
  \caption{Gaze score histogram.
  Distribution of gaze scores for all 1{,}152 heads.
  Most heads score near zero (no image-token attention); the top-100 cutoff (dashed red) isolates the tracking heads.}
  \label{fig:gaze_histogram}
\end{figure}

\cref{fig:gaze_histogram} shows the distribution of raw gaze scores across all 1{,}152 heads.
The bulk of heads sit at very low scores ($\le 0.05$), reflecting the fact that most heads spend nearly all of their attention budget on text tokens or distribute it diffusely across image tokens.
A long right tail extends to scores above $0.5$.
The top-100 cutoff (dashed red line) cleanly separates a small population of strong gaze heads from this background, and these heads concentrate in the same middle-to-late layer band identified by the layer-steering analysis (\cref{sec:layer_analysis}).

\begin{figure}[t]
  \centering
  \includegraphics[width=\linewidth]{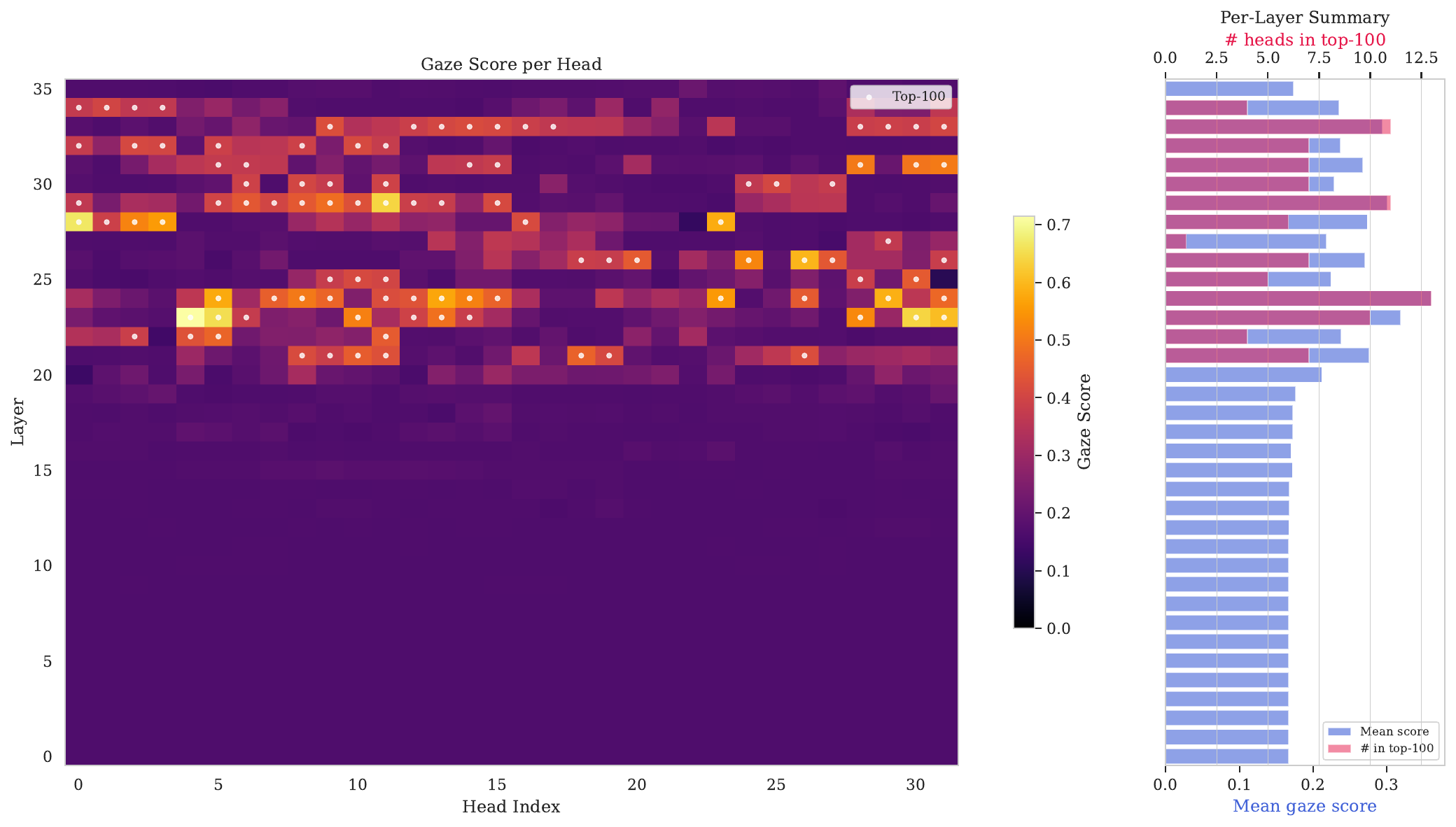}
  \caption{Gaze scores across layers and heads.
  \emph{Left:}~Each cell shows the gaze score for one (layer, head) pair; white dots mark the top-100 heads, which cluster in layers 20--28.
  \emph{Right:}~Per-layer summary showing mean gaze score (blue) and number of top-100 heads per layer (red).}
  \label{fig:gaze_layer_map}
\end{figure}

\cref{fig:gaze_layer_map} reveals where these high-scoring heads reside: gaze heads concentrate in a narrow band of middle-to-late layers (approximately layers 20--28), with the highest density around layers 21--25.
Early layers ($\ell < 15$) contain virtually no gaze heads, while late layers ($\ell > 30$) contain a few weak ones.
This spatial clustering aligns with the layer-level steering results in \cref{sec:layer_analysis}, where the same middle-layer band was identified as the locus of visual attention control.

\subsection{Reverse Narration Trajectory}
\label{app:reverse_trajectory}

To confirm that gaze heads track the narrated panel rather than following a fixed left-to-right spatial bias, we prompt the model with \textit{``Please describe what happens in each panel, in reverse order:''} and record the same per-head attention trajectories as in \cref{sec:trajectory}.

\begin{figure}[t]
  \centering
  \includegraphics[width=\linewidth]{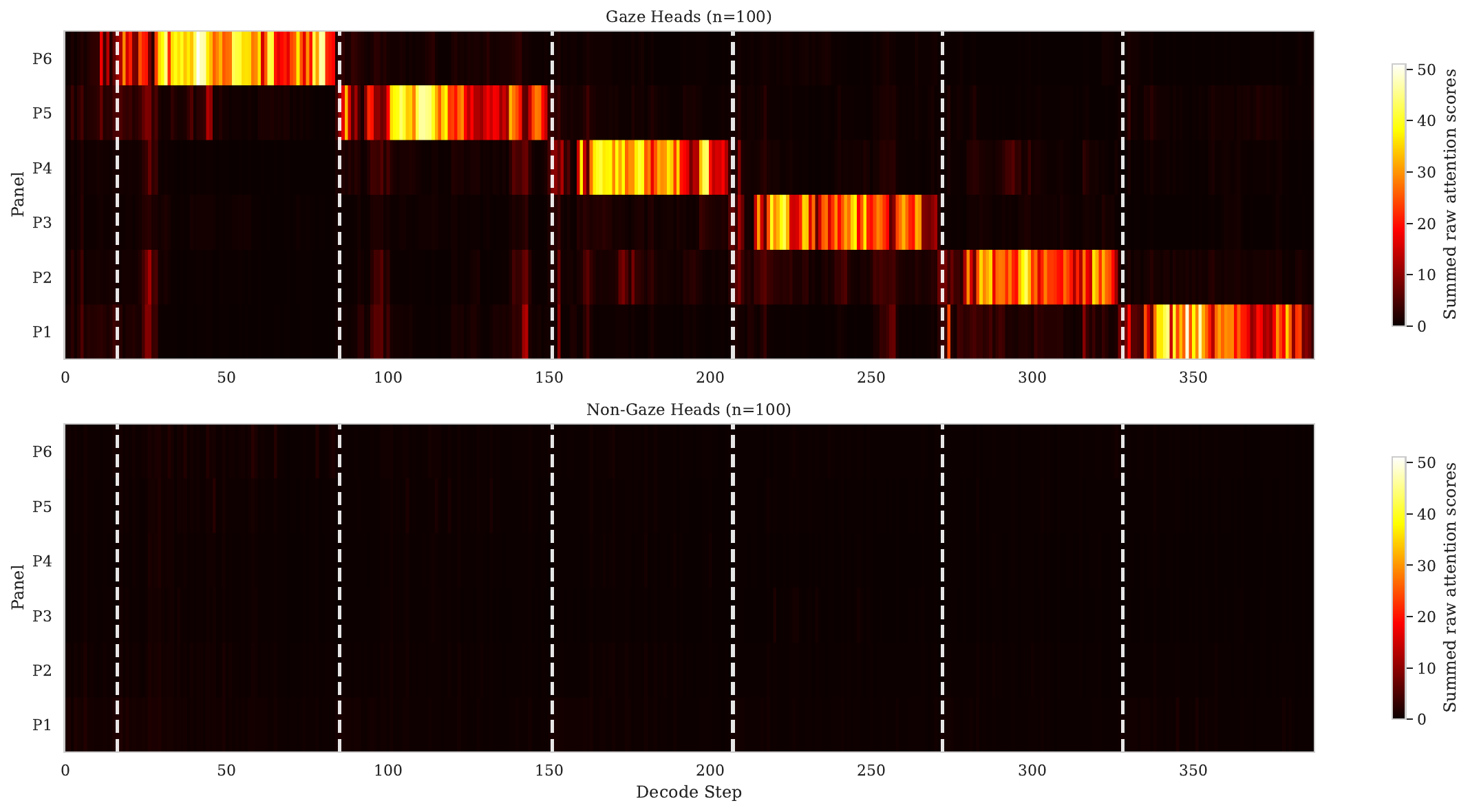}
  \caption{Gaze-head attention during reverse free narration.
  \emph{Top:}~The top-100 gaze heads show a reverse staircase aligned with the narrative: attention shifts panel-by-panel as the model describes each panel.
  \emph{Bottom:}~100 random non-gaze heads show no panel-tracking structure.
  Dashed lines mark the transition points between panel descriptions at generation.}
  \label{fig:reverse_trajectory}
\end{figure}

\cref{fig:reverse_trajectory} shows the result. The top-100 gaze heads produce a clear reverse staircase: attention begins on panel~6 and steps backward through each panel as the model describes them from right to left. The pattern is a mirror image of the forward staircase in \cref{fig:head_gaze_heatmap}, confirming that gaze heads dynamically follow the narration order rather than defaulting to a fixed spatial scan.

% =============================================================================
\section{Steering: Extended Analysis}
\label{app:steering_extended}

\subsection{Intervention-Strength Ablation}
\label{app:delta_ablation}

The main-text experiments use $\delta = +\infty$ for the attention-mask bias (implemented as $10{,}000$, which saturates the softmax in bfloat16) and so produce a \emph{hard} reassignment of each head's image-attention onto the target panel (\cref{sec:redirect}). A natural reviewer question is whether the hard limit is necessary or whether softer values of $\delta$ also redirect the model. We sweep $\delta \in \{1, 10, 100, 1{,}000, 10{,}000\}$ on Qwen3-VL-8B at $K{=}100$ (top-$100$ gaze heads, $500$ strips $\times 6$ target panels $= 3{,}000$ pairs each, non-gaze sampled from layers $20$--$35$ excluding the gaze head set), holding every other detail of the intervention fixed.

\begin{table}[ht]
\centering
\small
\caption{Intervention-strength ($\delta$) ablation on Qwen3-VL-8B at $K{=}100$. Larger $\delta$ corresponds to a harder reassignment of attention onto the target panel; $\delta=10{,}000$ saturates the bfloat16 softmax and corresponds to the $\delta = +\infty$ limit used throughout the main text. All numbers are forced 1-of-6 LLM-judge accuracy on the 500 strips $\times$ 6 target panels = 3{,}000 pairs validation set.}
\label{tab:delta_ablation}
\begin{tabular}{rccc}
\toprule
$\delta$ & Gaze (\%) & Non-gaze (\%) & Delta (pp) \\
\midrule
$1$       & 33.2\%       & 4.5\%       & $28.7$ \\
$10$      & 79.8\%       & 13.8\%      & $66.0$ \\
$100$     & 79.8\%       & 15.2\%      & $64.6$ \\
$1{,}000$ & 76.8\%       & 13.5\%      & $63.3$ \\
$10{,}000$ (used) & 83.1\% & 14.6\%   & $68.5$ \\
\bottomrule
\end{tabular}
\end{table}

The redirection result is robust to intervention strength. The sweep shows a sharp transition between $\delta{=}1$ and $\delta{=}10$: at $\delta{=}1$ the bias barely moves the softmax (gaze $33.2\%$, non-gaze $4.5\%$, a $7.4\times$ gap but only a partial steer); from $\delta{=}10$ onward the curve stays within $\sim 6$\,pp of the hard $\delta = +\infty$ limit across three orders of magnitude ($\delta{=}10$: $79.8\%$; $\delta{=}100$: $79.8\%$; $\delta{=}1000$: $76.8\%$; $\delta{=}10000$: $83.1\%$), with the maximum at the hard limit. The main-text headline is therefore \emph{not} an artifact of the hard $\pm\infty$ saturation limit, since a far softer intervention ($\delta{=}10$, well short of the bfloat16 saturation point) lands within $3$--$4$\,pp of it. We use $\delta = +\infty$ throughout the paper because it admits a clean closed-form description and gives the cleanest gaze accuracy, but any $\delta \geq 10$ produces effective redirection. The non-gaze control stays at $\sim 4$--$15\%$ across the entire sweep, confirming that the transition is in head-targeted reassignment rather than a generic ``stronger intervention $\Rightarrow$ higher accuracy'' effect.

\subsection{Head-Selection Baselines: Protocol}
\label{app:baseline_protocol}

The Image Heads~\cite{deng2025maskcd} and Localization Heads~\cite{kang2025your} baselines both publicly release LLaVA-specific implementations of their head-selection criteria (\url{https://github.com/Deng-Jingyuan/MaskCD} and \url{https://github.com/seilk/LocalizationHeads}, respectively). We port each criterion to Qwen3-VL-8B by faithfully reproducing the head-selection algorithm rather than the LLaVA-specific glue. Concretely:

\paragraph{Image Heads (MaskCD).} The MaskCD inference code sums each head's attention mass over the image-token region in a single forward pass, then z-score normalizes within each layer; heads whose per-layer z-score exceeds $2.5$ are flagged as ``image heads.'' For a top-$K$ ranking we sort all $1{,}152$ Qwen3-VL-8B heads by their (mean over the $500$ discovery panel-query prompts) per-layer z-score and take the top $K$. This reproduces MaskCD's intent on our setting: image-attending heads are those that put outlying attention on image tokens within their layer.

\paragraph{Localization Heads (Kang et al.).} Their analysis pipeline (\texttt{analyze.py}) ranks heads by two criteria. (1) ``Criterion-1'' picks heads whose image-attention sums lie above an elbow threshold (chord-distance method, also defined in \texttt{analyze.py}). (2) ``Criterion-2'' computes the spatial entropy of each head's 2D attention map (\texttt{spatial\_entropy(attn\_map\_2d, threshold)} in their code) on the $P{\times}P$ patch grid; heads with low entropy are more spatially concentrated. They additionally drop heads in layer $\leq 1$ and heads whose attention concentrates on the bottom row (a LLaVA-specific summary-token filter). Heads that pass both criteria are ranked by ascending spatial entropy. We port the algorithm exactly: collect the per-(layer, head, patch) attention from the final prompt token on each of the $6$ panel-query prompts across the $500$ discovery strips, average to a $({n\_layers}\times n\_heads\times P^2)$ tensor with $P{=}24$, then run their unchanged \texttt{analyze\_heads(\dots)} on it (porting only the necessary config keys, not the algorithm). Layer-skip and bottom-row-focus filters are kept as in the original code.

\paragraph{Non-gaze control.} The non-gaze control samples $100$ heads uniformly from layers $20$--$35$ of the model, excluding the gaze head set, and is identical across all rows of the baseline table; this gives an apples-to-apples ``what if we just intervened on heads in the same layer band that aren't gaze heads'' control regardless of which positive-head criterion we are evaluating.

\paragraph{Intervention.} Identical across rows: the same boost-suppress attention-mask intervention used for our gaze heads ($\delta = +\infty$, hard reassignment of each head's image-attention onto the target panel; \cref{sec:redirect}).

\subsection{Head-Selection Baselines: Full K Sweep}
\label{app:cross_arch_sweep_baselines}

\cref{tab:cross_arch_sweep_baselines} reports the full top-$K$ sweep for our gaze score, the Image Heads selection~\cite{deng2025maskcd}, and the Localization Heads selection~\cite{kang2025your} on Qwen3-VL-8B (500 strips $\times$ 6 target panels = 3{,}000 pairs, non-gaze sampled from layers $20$--$35$ excluding the gaze head set). For both baselines we re-implement their published head-identification algorithms faithfully (protocol in \cref{app:baseline_protocol}) and apply them to the same 500-strip discovery set. The intervention is identical across rows (the same boost-suppress attention-mask edit used for our gaze heads). The gaze score is the most head-efficient criterion at small $K$, where head-identification quality dominates the result; at large $K$ every reasonable image-attention criterion eventually saturates the intervention.

\begin{table*}[ht]
\centering
\small
\caption{Comparison of head-selection criteria for the same attention-mask intervention. Gaze-redirection accuracy on Qwen3-VL-8B; 500 strips $\times$ 6 target panels = 3{,}000 pairs each. The non-gaze control samples 100 heads uniformly from layers $20$--$35$ excluding the gaze head set, identical across all rows for direct comparability.}
\label{tab:cross_arch_sweep_baselines}
\begin{tabular}{lccccc}
\toprule
$K$ & Ours (gaze) & Image Heads~\cite{deng2025maskcd} & Localization Heads~\cite{kang2025your} & Non-gaze & All-heads \\
\midrule
10  & 45.3 & 10.0 & 24.3 & 9.8  & 0.9 \\
50  & 63.2 & 36.2 & 41.7 & 15.7 & 0.9 \\
100 & 83.1 & 69.0 & 60.2 & 14.6 & 0.9 \\
150 & 75.6 & 60.8 & 59.5 & 13.8 & 0.9 \\
200 & 73.0 & 58.2 & 58.8 & 13.5 & 0.9 \\
\bottomrule
\end{tabular}
\end{table*}

\cref{tab:head_set_agreement} measures how much the three criteria agree on \emph{which} heads they select, at both the $K{=}10$ and $K{=}100$ cuts.

\begin{table*}[ht]
\centering
\small
\caption{Agreement between head-selection criteria on Qwen3-VL-8B ($1{,}152$ heads total). For each pair we report the overlap of their top-$K$ heads (out of $K$) and the Jaccard index, at $K{=}10$ and $K{=}100$. A random pair of size-$K$ sets would overlap $\approx 0.1$ heads at $K{=}10$ and $\approx 8.7$ at $K{=}100$. The three criteria are correlated in their top-$100$ but nearly disjoint in their top-$10$, and only $13$ heads are shared by all three even at $K{=}100$.}
\label{tab:head_set_agreement}
\begin{tabular}{lcccc}
\toprule
 & \multicolumn{2}{c}{Top-$10$} & \multicolumn{2}{c}{Top-$100$} \\
\cmidrule(lr){2-3} \cmidrule(lr){4-5}
Head-set pair & Overlap & Jaccard & Overlap & Jaccard \\
\midrule
Gaze $\cap$ Image Heads~\cite{deng2025maskcd}      & 0/10 & 0.000 & 43/100 & 0.274 \\
Gaze $\cap$ Localization Heads~\cite{kang2025your} & 1/10 & 0.053 & 26/100 & 0.149 \\
Image Heads $\cap$ Localization Heads              & 0/10 & 0.000 & 33/100 & 0.198 \\
\midrule
All three (common core)                            & 0/10 & 0.000 & 13/100 & 0.062 \\
\bottomrule
\end{tabular}
\end{table*}

The picture is sharpest at small $K$. At $K{=}10$ the three criteria pick essentially disjoint sets: the gaze score shares one head with Localization Heads, none with Image Heads, and the two baselines share none with each other, against $\approx 0.1$ heads expected by chance. At small $K$, where head-selection quality dominates redirection accuracy (\cref{tab:cross_arch_sweep_baselines}), the three criteria do not even agree on \emph{which} heads matter most. At $K{=}100$ the overlap rises into the $26$--$43$ range, three to five times the $\approx 8.7$-head chance level, but each criterion's set is still mostly its own and only $13$ heads pass all three filters. The redirection gap in \cref{tab:cross_arch_sweep_baselines} is therefore not a matter of the same heads re-ranked: the temporal re-routing criterion behind the gaze score selects heads the single-pass image-attention criteria miss, and those are the heads that move the small-$K$ accuracy.

\subsection{Prompt Sensitivity}
\label{app:prompt_sensitivity}

We test whether the VQA-redirection result depends on the exact wording of the question. \cref{tab:prompt_sensitivity} reports redirection accuracy under five prompt variants on Qwen3-VL-8B at $K{=}100$ on the 500-strip validation set ($n{=}3{,}000$ pairs each). The default paper prompt is the first row; the other four cover shorter, longer, and reframed wordings of the same question. All variants use the same forced 1-of-6 LLM judge.

\begin{table}[ht]
\centering
\small
\caption{Prompt-sensitivity of gaze-head redirection on Qwen3-VL-8B. Top-$100$ gaze heads, boost-suppress intervention, 500 strips $\times$ 6 target panels = 3{,}000 pairs each.}
\label{tab:prompt_sensitivity}
\begin{tabular}{p{0.55\linewidth}cc}
\toprule
Prompt & Gaze  & Non-Gaze  \\
\midrule
\emph{What is the main action or event happening in this comic strip? Answer briefly.} (default) & 83.1\% & 14.6\% \\
\emph{Describe this comic briefly.} & 64.7\% & 13.3\% \\
\emph{Look at this comic strip carefully. What is the main action or event happening across the panels? Answer in one short sentence.} & 80.3\% & 7.3\% \\
\emph{What story does this comic strip tell? Answer briefly.} & 74.2\% & 13.3\% \\
\emph{Summarize the main event of this comic in a few words.} & 78.3\% & 16.2\% \\
\bottomrule
\end{tabular}
\end{table}

Across the five prompt phrasings the gaze-redirection accuracy ranges from 64.7\% to 83.1\% (an 18.4-pp span), with the non-gaze control staying near chance ($\sim 7$--$16\%$) on every variant. The redirection effect is robust to prompt wording: every variant produces a $\ge$ 50-pp gap between gaze and non-gaze, far larger than the prompt-induced variation.

\subsection{Dynamic Narration: Trajectory-Level Judge}
\label{app:dynamic_traj_judge}

The strict 1-of-6 judge used for the main-text headline (\cref{sec:dynamic_switch}) penalizes any segment whose dominant content is not the scheduled target panel by exactly that segment's boundary, including segments that legitimately finish the previous panel's sentence before transitioning. To check that the headline result is not an artifact of a strict judge, we rejudge the same generations with a trajectory-level LLM judge that hides the schedule entirely (\cref{tab:dynamic_traj}). The judge sees the strip image and all six 50-token segments at once, and is asked, for each segment, to identify which panel of the strip its content \emph{dominantly} describes. Repeats are allowed (the model may revisit a panel) and a segment whose content is incoherent or empty is mapped to \texttt{null}. From the six attributions per (strip, condition) we compute Spearman~$\rho$ of the predicted-panel sequence against (i) the steering schedule and (ii) the natural $[1{,}2{,}3{,}4{,}5{,}6]$ order, plus per-segment match rates for each.

\begin{table*}[ht]
\centering
\small
\caption{Trajectory-level rejudge of the dynamic-narration generations. 500 strips, strict derangement schedule. Judge does not see the schedule. Per-segment matches are computed only over non-null segments; junk counts list how many strips have at least 3 null segments (and so do not contribute to Spearman~$\rho$).}
\label{tab:dynamic_traj}
\begin{tabular}{lcccc}
\toprule
Condition & Match~vs.~schedule & Match~vs.~natural & $\rho$~vs.~schedule & $\rho$~vs.~natural \\
\midrule
Gaze (top-100)        & 71.1\% & 13.6\%   & $+0.589$  & $+0.259$   \\
Non-gaze              & 16.4\%          & 20.8\%   & $-0.140$            & $+0.983$ \\
All-heads             & 7.1\%         & 7.5\%               & $-^\dagger$          & $-^\dagger$           \\
\bottomrule
\end{tabular}
\end{table*}

The trajectory judge agrees directionally with the strict 1-of-6 judge: gaze tracks the schedule at $71.1\%$ per-segment match and $\rho{=}{+}0.589$, vs.\ non-gaze at $16.4\%$ and $\rho{=}{-}0.140$. The new information is the $\rho$ vs.\ \emph{natural} column: under gaze redirection, the model's predicted-panel sequence has only $\rho{=}{+}0.259$ correlation with the default left-to-right $[1,2,3,4,5,6]$ order, while the non-gaze control sits at $\rho{=}{+}0.983$, essentially the perfect default scan. The two-column comparison shows that gaze heads do not just disrupt the default scan; they replace it with the steered schedule.

$^\dagger$Spearman~$\rho$ is not reported for the all-heads condition because the junk fraction is too high ($384/500$ strips with $\geq 3$ null segments), leaving too few intact trajectories to attribute reliably.

\subsection{Gaze Heads on Natural Images}
\label{app:natural_images}

The experiments in the main paper use comic strips, where visual content is divided into discrete panels. Do gaze heads also perform spatial grounding on natural images, where regions are not explicitly delineated?

\paragraph{Setup.}
We prompt Qwen3-VL-8B with a natural image and the instruction \textit{``Describe what is happening in this image in detail:''} The model generates a free-form description of approximately 300 tokens. We capture value-weighted attention scores~\cite{kobayashi2020attention} from the top-100 gaze heads (discovered via the comic panel task) at every decode step during a single generation pass.

\paragraph{Concept segmentation.}
We segment the generated text into spatial region spans using Claude Sonnet~\cite{anthropic2025claude}, where each span covers the tokens corresponding to a single described region (e.g., ``notebooks and pen'' = tokens 48--97, ``headphones'' = tokens 135--156). This gives us token ranges indicating when the model is describing each part of the image.

\paragraph{Heatmap construction.}
For each concept span, we average the per-token image-attention vectors across all decode steps in that span and across all heads in the set (gaze or random). This produces a 1D vector over image tokens, which we reshape to the spatial grid matching the image's token layout. The resulting heatmap is upsampled and overlaid on the original image using a jet colormap.

\begin{figure*}[t]
  \centering
  \includegraphics[width=\linewidth]{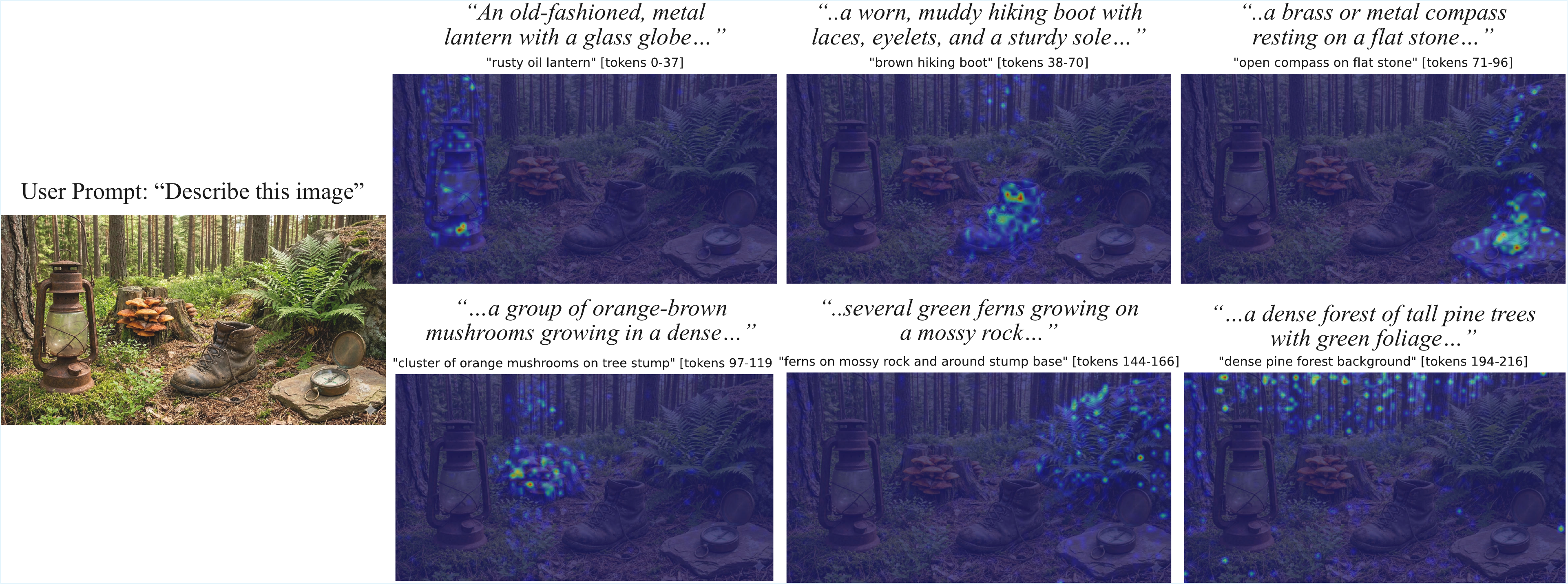}
  \caption{Gaze-head attention on a natural image. Leftmost: the original image. Remaining panels: gaze-head attention heatmaps for three concept spans during free-form description. Attention shifts to the spatial region corresponding to each described object, confirming that gaze heads perform spatial grounding beyond comic panels.}
  \label{fig:natural_image_gaze_1}
\end{figure*}

\begin{figure*}[t]
  \centering
  \includegraphics[width=\linewidth]{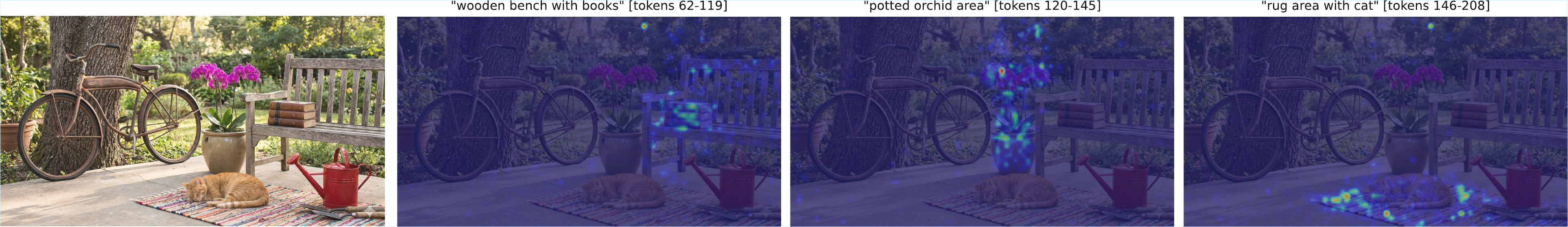}
  \caption{Gaze-head attention on a natural image. Leftmost: the original image. Remaining panels: gaze-head attention heatmaps for three concept spans during free-form description. Attention shifts to the spatial region corresponding to each described object, confirming that gaze heads perform spatial grounding beyond comic panels.}
  \label{fig:natural_image_gaze_2}
\end{figure*}

\begin{figure*}[t]
  \centering
  \includegraphics[width=\linewidth]{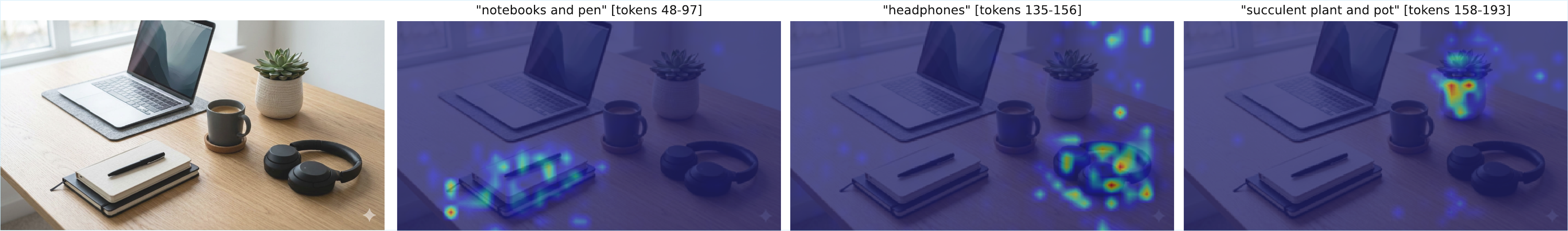}
  \caption{Gaze-head attention on a natural image. Leftmost: the original image. Remaining panels: gaze-head attention heatmaps for three concept spans during free-form description. Attention shifts to the spatial region corresponding to each described object, confirming that gaze heads perform spatial grounding beyond comic panels.}
  \label{fig:natural_image_gaze_3}
\end{figure*}

\paragraph{Findings.}
\cref{fig:natural_image_gaze_1}, \cref{fig:natural_image_gaze_2}, and \cref{fig:natural_image_gaze_3} show the results across three natural images. Gaze heads produce spatially concentrated attention that tracks the described region: when the model describes ``notebooks and pen,'' attention concentrates on the lower-left of the image where these objects are located, and when it describes ``succulent plant and pot,'' attention shifts to the upper-right. The grounding is less precise than on panel-structured images, since natural images have no explicit region boundaries, but the correspondence between described content and attended region is consistent. This suggests that the gaze heads discovered through comic strip probing are not specific to panel-structured images: they perform a general spatial grounding function, attending to the region of the image the model is currently describing.

\paragraph{Quantitative natural-image redirection (COCO val2017).}
\label{app:coco_natural_vqa}
To put a number on the natural-image steering claim, we evaluate gaze-head redirection on all $5{,}000$ images in COCO val2017~\cite{lin2014microsoft}, sweeping over the $31{,}781$ annotated objects in the set ($8{,}897$ large, $12{,}569$ medium, $10{,}315$ small across $80$ categories). For each (image, object) pair we apply a minimum bounding-box area filter of $750$ pixels: smaller boxes cover only a handful of image tokens in Qwen3-VL-8B's input grid, and the additive attention bias struggles to redirect attention onto a region that contains so few tokens. The filter is essentially a no-op for the large and medium classes but removes most of the small class, leaving an evaluation sample of $23{,}452$ pairs: $8{,}897$ large, $12{,}569$ medium, and $1{,}986$ small. For each pair we steer the top-100 gaze heads to the target object's COCO bounding box (mapping pixel coordinates to image-token positions via cell-center containment) and ask the model ``What is the main object in this part of the image? Answer in a few words.'' Claude Sonnet judges whether the steered answer names the target COCO category, with synonym matching (e.g., ``car''~$\sim$~``automobile''); see \cref{app:judge_object}. The non-gaze control samples 100 heads from layers $20$--$35$, excluding the gaze head set (\cref{app:nongaze_sampling}).

\cref{tab:coco_main} reports accuracy broken down by COCO size class. Gaze-head steering achieves $80.3\%$ on large objects ($>96^2$ px), where the object occupies enough image tokens for the bounding-box bias to bite cleanly, and $76.2\%$ on medium objects, both well above the non-gaze control ($18.6$--$36.6\%$ depending on size class). Performance drops on small objects whose bounding boxes cover only a handful of image tokens and where the natural attention bias dominates over our intervention. This converts the qualitative natural-image observation in \cref{fig:natural_image_gaze_1} into a quantitative claim: the same heads identified through comic probing also steer the model's answer toward an arbitrary spatial region of a natural image.

% =============================================================================
\section{Generalization Across Sizes and Architectures}
\label{app:generalization_all}

\subsection{Generalization Across Model Sizes}
\label{app:model_sizes}\label{app:gaze_model_sizes}

We run the full pipeline on four Qwen3-VL sizes: 2B (28 layers, 16 heads), 4B (36 layers, 32 heads), 8B (36 layers, 32 heads), and 32B (64 layers, 64 heads).

\begin{figure}[t]
  \centering
  \includegraphics[width=\linewidth]{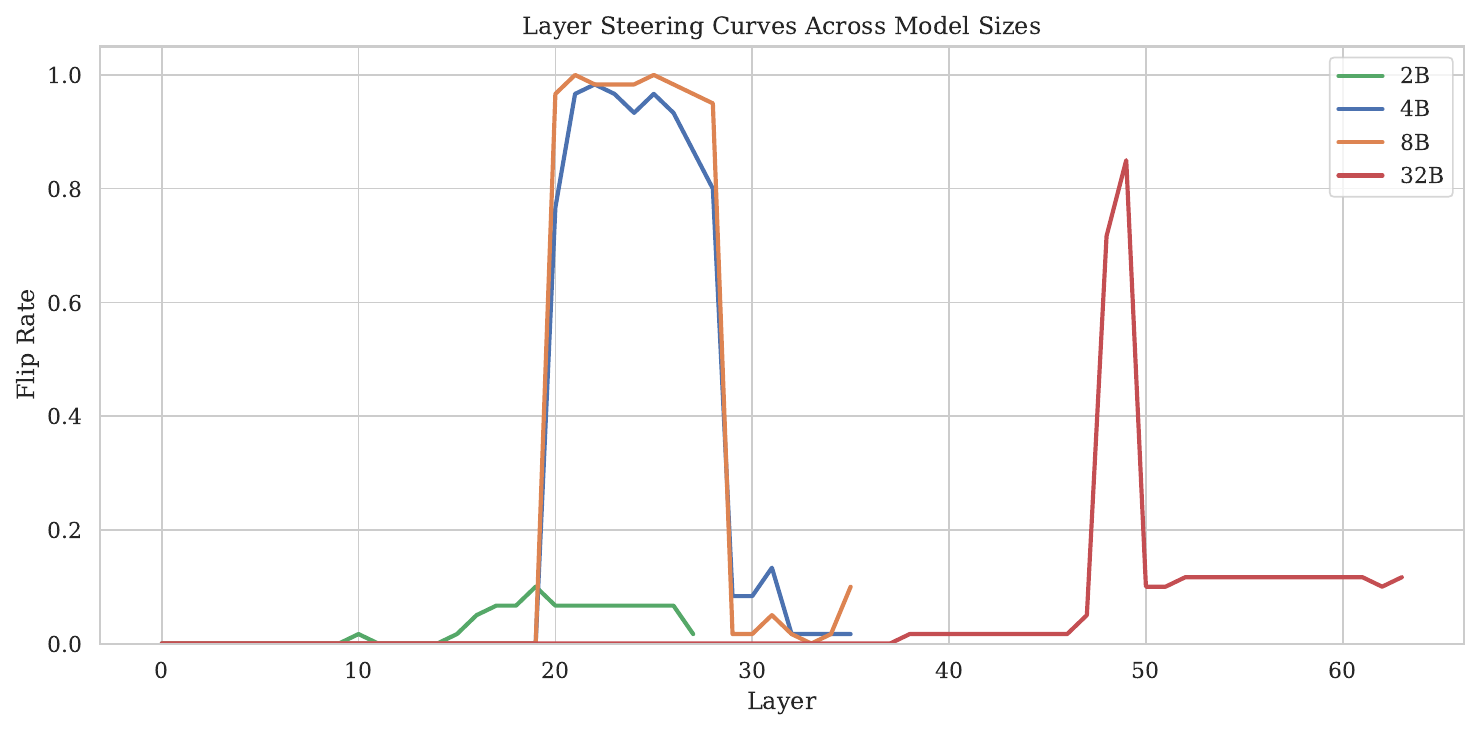}
  \caption{Layer steering across model sizes.
  The 4B and 8B models localize visual attention control in layers 20--28; the 32B model places it near layer~49.
  The 2B model shows minimal steering effect at any layer.}
  \label{fig:model_size_flip}
\end{figure}

\paragraph{Layer steering.}
\cref{fig:model_size_flip} compares per-layer flip rates across all four sizes.
The 4B and 8B models, which share the same depth (36 layers), both localize visual attention control in layers 20--28, with best flip rates of 98.3\% (L22) and 97.0\% (L21).
The 32B model (64 layers) places its effective band deeper at layer~49, achieving 85\%.
The 2B model (28 layers) is the clear outlier: its best layer (L19) achieves only 10\%, suggesting insufficient capacity for a robust gaze mechanism amenable to difference-of-means steering.

\begin{figure}[t]
  \centering
  \includegraphics[width=0.75\linewidth]{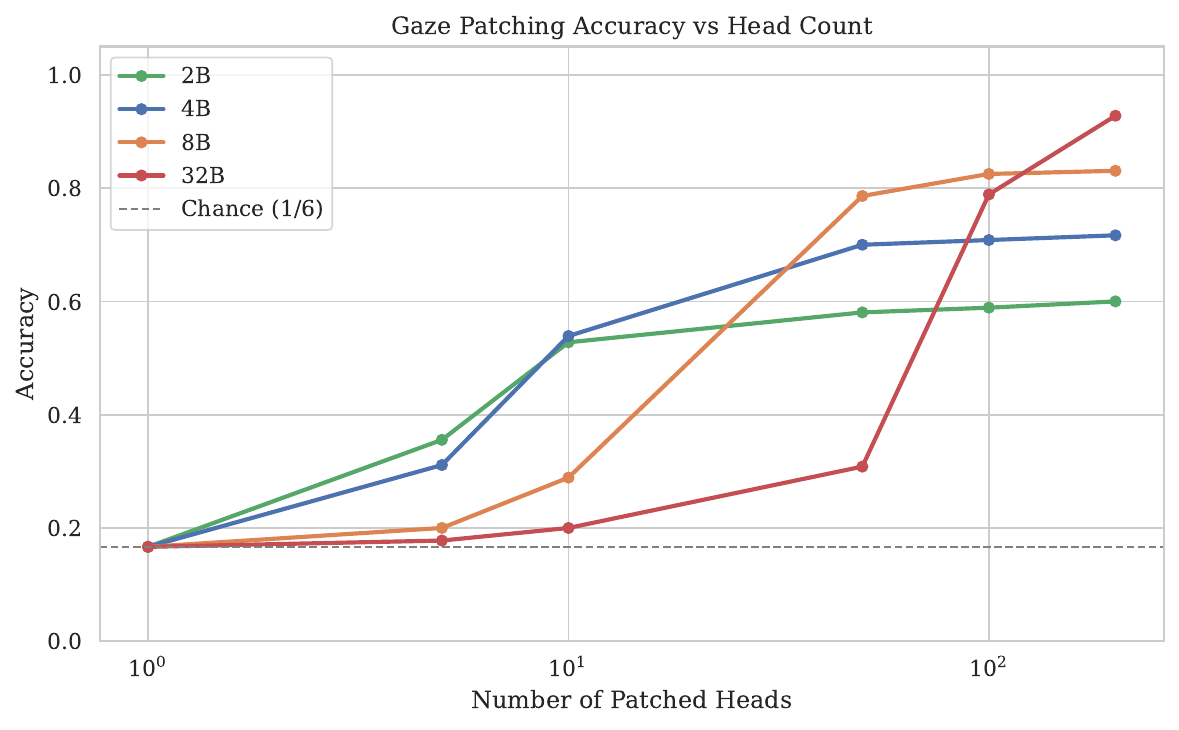}
  \caption{Gaze redirection accuracy across model sizes.
  All four Qwen3-VL sizes follow a hump-shaped saturation curve, with the peak $K$ scaling with the model's total head count (\cref{tab:gaze_model_sizes}).}
  \label{fig:model_size_gaze}
\end{figure}

\paragraph{Gaze discovery and redirection.}
We apply the same gaze-head discovery process to all four sizes and evaluate the discovered heads on VQA redirection across the full top-$K$ sweep, using the 500-strip validation set and forced 1-of-6 LLM judge of the main text. \cref{tab:gaze_model_sizes} reports gaze / non-gaze accuracy at each $K$. All four sizes follow a hump-shaped curve with a clear single peak, and the non-gaze control stays at $\le 15\%$ throughout. The 8B model achieves the highest peak (83.1\% at $K{=}100$, reproducing the main-text VQA headline); the 4B model peaks at 72.9\% at $K{=}75$, the 32B model at 70.2\% at $K{=}500$ after a longer climb, and the 2B model earliest at 68.6\% at $K{=}10$. The peak $K$ scales roughly with total head count: 2B (448 heads), 4B / 8B (1{,}152 heads), and 32B (4{,}096 heads) peak at $K$ values corresponding to $\sim 2\%$, $\sim 7$--$9\%$, and $\sim 12\%$ of all heads. \cref{fig:model_size_gaze} plots these saturation curves.

\begin{table}[t]
\centering
\footnotesize
\caption{Full top-$K$ gaze / non-gaze VQA accuracy across Qwen3-VL sizes. 500-strip validation set, forced 1-of-6 LLM judge (chance $16.7\%$); non-gaze heads are sampled from the same layer range as each model's gaze heads, excluding the gaze head set. Each cell is \textit{gaze}\,/\,\textit{non-gaze} percent. Dashes indicate $K$ values not in the per-size sweep.}
\label{tab:gaze_model_sizes}
\begin{tabular}{lcccc}
\toprule
$K$ & 2B & 4B & 8B & 32B \\
\midrule
5    & 65.0 / 13.0 & 18.9 / 8.6  & 36.0 / 6.6  & 11.7 / 2.7 \\
10   & 68.6 / 1.3  & 49.7 / 10.5 & 45.3 / 10.0 & 16.2 / 3.5 \\
50   & 48.6 / 0.5  & 68.2 / 4.3  & 63.2 / 15.3 & 30.3 / 7.3 \\
75   & ---         & 72.9 / 1.5  & 76.3 / 13.5 & --- \\
100  & 46.7 / 0.4  & 70.3 / 1.6  & 83.1 / 14.6 & 32.5 / 8.3 \\
125  & ---         & 68.3 / 1.8  & 79.4 / 13.3 & --- \\
150  & 50.5 / 0.5  & 69.7 / 1.6  & 75.6 / 14.5 & 29.5 / 8.7 \\
200  & 36.6 / 0.5  & 55.3 / 1.6  & 73.0 / 14.5 & 42.8 / 11.2 \\
300  & ---         & ---         & ---         & 63.5 / 12.3 \\
400  & ---         & ---         & ---         & 68.8 / 12.2 \\
500  & ---         & ---         & ---         & 70.2 / 12.0 \\
600  & ---         & ---         & ---         & 60.2 / 12.3 \\
700  & ---         & ---         & ---         & 55.2 / 12.5 \\
800  & ---         & ---         & ---         & 59.7 / 11.8 \\
900  & ---         & ---         & ---         & 58.8 / 12.3 \\
1000 & ---         & ---         & ---         & 47.7 / 12.0 \\
\bottomrule
\end{tabular}
\end{table}

\begin{figure*}[t]
  \centering
  \includegraphics[width=\linewidth]{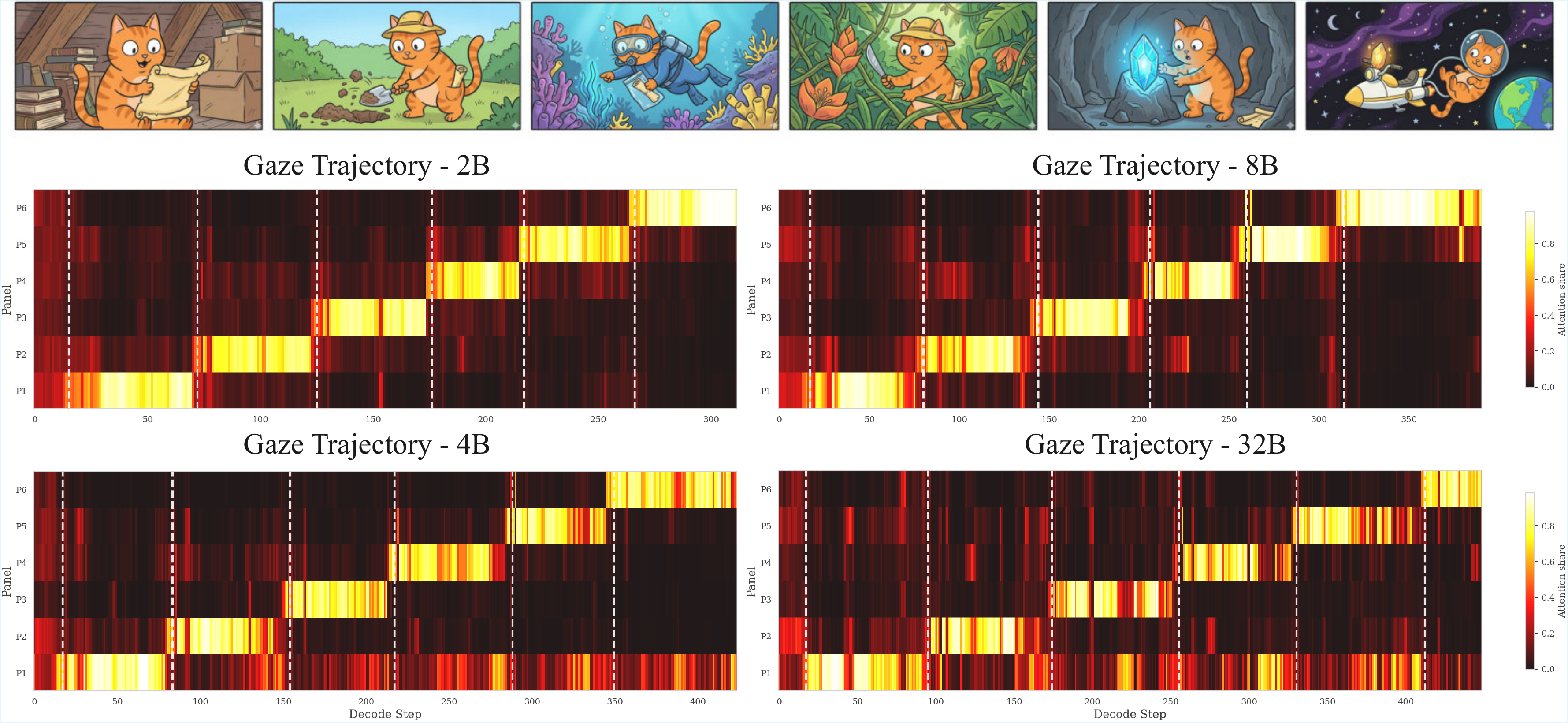}
  \caption{Gaze-head attention trajectories across model sizes.
  All four Qwen3-VL sizes (2B, 4B, 8B, 32B) produce a staircase pattern during free-form narration on the same strip, confirming that gaze heads consistently track the narrated panel across scales.}
  \label{fig:gaze_trajectory_models}
\end{figure*}

\paragraph{Gaze-head trajectories.}
\cref{fig:gaze_trajectory_models} shows gaze-head attention trajectories during free-form narration for all four model sizes on the same strip. All four produce a clear staircase pattern, confirming that gaze heads are a consistent organizational feature across scales.

\begin{figure}[t]
  \centering
  \includegraphics[width=\linewidth]{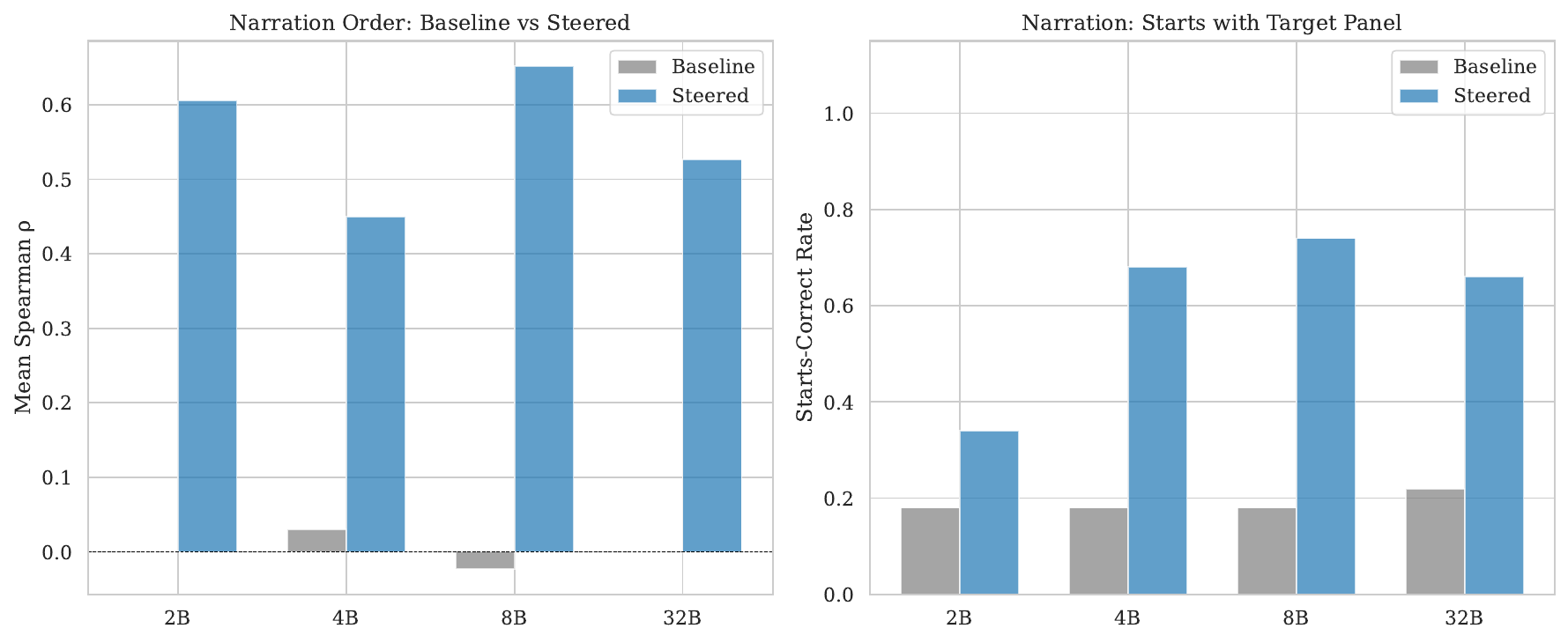}
  \caption{Gaze-steered narration across model sizes.
  All four sizes produce positive steered $\rho$, with 8B achieving the strongest correlation.}
  \label{fig:model_size_narration}
\end{figure}

\paragraph{Gaze-steered narration.}
\cref{fig:model_size_narration} shows baseline versus steered Spearman $\rho$ and starts-correct rate, using an older pilot protocol (per-strip ``first segment matches target'' rate alongside $\rho$ between the steered narration order and the target schedule, on a smaller cross-size batch). Gaze-steered narration produces positive $\rho$ on all four sizes: 2B ($\rho = +0.61$, 34\% starts-correct), 4B ($\rho = +0.45$, 68\%), 8B ($\rho = +0.62$, 68\%), and 32B ($\rho = +0.53$, 66\%). The 8B model achieves the strongest correlation overall.
The 2B narration result is surprisingly strong given its weak layer steering, possibly because gaze-head redirection intervenes more directly on the attention routing mechanism than residual-stream steering. The headline narration result in the main text ($79.4\%$ static narration steering on the 8B model) is not directly comparable to these starts-correct numbers; what is consistent across protocols is that the 8B model shows the strongest gaze-steered narration effect.

\subsection{Cross-Architecture Generalization}
\label{app:cross_arch}

\cref{tab:cross_arch} in the main text reports the peak redirection accuracy for each other architecture we tested: Ovis1.5-8B~\cite{lu2024ovis}, Qwen2-VL-7B~\cite{wang2024qwen2}, InternVL3.5-8B~\cite{wang2025internvl3}, LLaVA-1.5-13B~\cite{liu2024improved}, and LLaVA-NeXT-7B~\cite{liu2024llavanext}. The Qwen3-VL-8B row in that table is the headline result; all rows are evaluated on the full 500-strip validation set. This appendix documents the preprocessing fix that makes the cross-family transfer possible, the full $K$-sweep behind those peak numbers, and a within-family scale comparison.

\paragraph{Preprocessing.} The fixed-resolution families (Ovis, InternVL, and both LLaVA variants) wrap a center-cropping image processor that, on our wide strips, would crop away five of the six panels before they reach the language model. We apply a one-line panel-preservation fix (full details and validation in \cref{app:cross_arch_preproc}) so all six panels remain visible; without it, redirection on these families is indistinguishable from chance. Qwen2-VL, like Qwen3-VL, uses an aspect-ratio-preserving processor and needs no fix.

\subsection{Cross-Architecture Preprocessing Fix}
\label{app:cross_arch_preproc}

\paragraph{The center-crop bug.} The fixed-resolution families wrap a CLIP-style image processor whose default behavior for a non-square input is:
(1) resize so that the \emph{shortest} side equals the model's crop size $S$ ($S{=}336$ for LLaVA-1.5 and LLaVA-NeXT, $S{=}448$ for InternVL, $S{=}384$ for Ovis), then
(2) center-crop to $S \times S$.
For our $1536 \times 256$ comic strips this scales the shortest side ($256$) up to $S$, which blows the longest side up to $\sim 6S$. Center-cropping then leaves only the middle $S$ pixels, about one panel out of the six. The other five panels are simply not in the model's input.

\paragraph{Fix.} Before passing the image to the processor we resize the strip directly to $S \times S$ (a fixed-size square; no center-crop). This is a horizontal squash (panel widths drop from $256$ px to $S / 6$ px), but every panel is preserved as a contiguous column of image tokens in the LM input. The effect is decisive: on LLaVA-1.5 the top gaze score's discovered heads come to correspond to all six panels rather than the center crop alone, and redirection on the validation set rises from $17.8\%$ (near chance) to $24.7\%$ at $K{=}100$, with a peak of $39.0\%$ at $K{=}160$. The other fixed-resolution families behave the same way; their peak accuracies with the fix in place are reported in \cref{tab:cross_arch}.

\paragraph{Implementation.} In \texttt{utils/modeling.py}, \texttt{prepare\_inputs} detects when the loaded processor is a \texttt{CLIPImageProcessor} or an InternVL / SigLIP-style processor and pre-resizes \texttt{image.resize((S, S), BILINEAR)} before building the prompt. For Qwen2-VL and Qwen3-VL the call is a no-op because their processors are aspect-ratio-preserving by construction (they tile patches to match the input's true aspect). Ovis additionally exposes its visual block through an out-of-band placeholder rather than ordinary text tokens, so we record the placeholder's expanded span at merge time to recover the image-token range. The fix and these per-family hooks add $\sim 15$ lines of code each, cost nothing at run time, and change no model weights.

\paragraph{What is squashed.} The squashed strip looks compressed horizontally to humans, but the language model still sees every panel as a separate contiguous block of image-token columns: $24$ wide $\to 4$ cols/panel for LLaVA-1.5 and the LLaVA-NeXT base tile ($24{\times}24{=}576$ tokens), $16$ wide $\to 2$--$3$ cols/panel for InternVL3.5 ($16{\times}16{=}256$ tokens after its $2{\times}2$ pixel-shuffle), and $27$ wide $\to 4$--$5$ cols/panel for Ovis ($27{\times}27{=}729$ tokens plus two visual-indicator tokens that we exclude from panel scoring). LLaVA-NeXT additionally appends an any-resolution tile whose row-end \texttt{image\_newline} tokens we mask out. The panels are visually narrow but their image-token representations are clean.

\subsection{Cross-Architecture: Extended Sweeps}
\label{app:cross_arch_extended}

\cref{tab:cross_arch_sweep} reports the full top-$K$ sweep, gaze (g) vs.\ non-gaze (n) accuracy at each $K$, for the five other-architecture families on the 500-strip validation set, all under the same discovery score and intervention as the main text (\cref{app:cross_arch}). All five show the same hump-and-collapse shape: a mid-$K$ peak, then a collapse into degenerate output as the $-\delta$ over-suppresses (junk\% column). The all-heads condition is omitted because it stays at $\le 3$\% throughout (peak-$K$ all-heads numbers are in \cref{tab:cross_arch}).

\begin{table*}[ht]
\centering
\footnotesize
\caption{Full top-$K$ sweep for the five other-architecture families: gaze (g) / non-gaze (n) redirection accuracy at each $K$ on the 500-strip validation set, with the gaze-condition junk\% in the last column. The discovery score and intervention are identical across families (\cref{app:cross_arch}); the in-grid maximum gaze accuracy is in bold. Exact per-model peaks fall between these grid points and are reported in \cref{tab:cross_arch}: Qwen2-VL $K{=}90$ ($66.2/0.0$), InternVL3.5 $K{=}140$ ($62.7/31.0$), LLaVA-1.5 $K{=}160$ ($39.0/13.8$).}
\label{tab:cross_arch_sweep}
\resizebox{\linewidth}{!}{%
\begin{tabular}{lcccccc}
\toprule
$K$ & Ovis1.5-8B g/n & Qwen2-VL-7B g/n & InternVL3.5-8B g/n & LLaVA-1.5-13B g/n & LLaVA-NeXT-7B g/n & junk\% (Ov/Qw/V3.5/L1.5/LN) \\
\midrule
10  & 20.0 / 13.7 & 21.0 / 8.7 & 9.0 / 10.7  & 15.0 / 2.0  & 15.7 / 4.0  & 13 / 1 / 0 / 2 / 7 \\
25  & 32.0 / 14.7 & 37.3 / 0.0 & 14.0 / 29.7 & 15.0 / 7.0  & 25.0 / 9.7  & 9 / 2 / 3 / 2 / 12 \\
50  & 45.3 / 12.7 & 50.7 / 0.0 & 36.0 / 40.7 & 21.7 / 13.0 & 26.7 / 25.0 & 16 / 5 / 5 / 3 / 11 \\
100 & \textbf{68.7} / 13.0 & \textbf{55.3} / 0.0 & 46.0 / 33.3 & 24.7 / 14.3 & \textbf{35.3} / 26.7 & 8 / 6 / 6 / 15 / 15 \\
150 & 67.3 / 13.7 & 46.7 / 0.3 & \textbf{57.7} / 32.3 & 36.3 / 13.7 & 28.7 / 26.0 & 11 / 11 / 6 / 12 / 23 \\
200 & 21.7 / 13.3 & 45.3 / 0.3 & 54.3 / 32.3 & \textbf{37.0} / 14.3 & 23.7 / 26.7 & 66 / 12 / 10 / 17 / 52 \\
300 & 13.0 / 14.3 & 47.0 / 0.3 & 44.3 / 32.0 & 32.7 / 14.3 & 19.0 / 26.3 & 83 / 16 / 23 / 27 / 62 \\
\bottomrule
\end{tabular}
}
\end{table*}

\begin{figure}[ht]
  \centering
  \includegraphics[width=\linewidth]{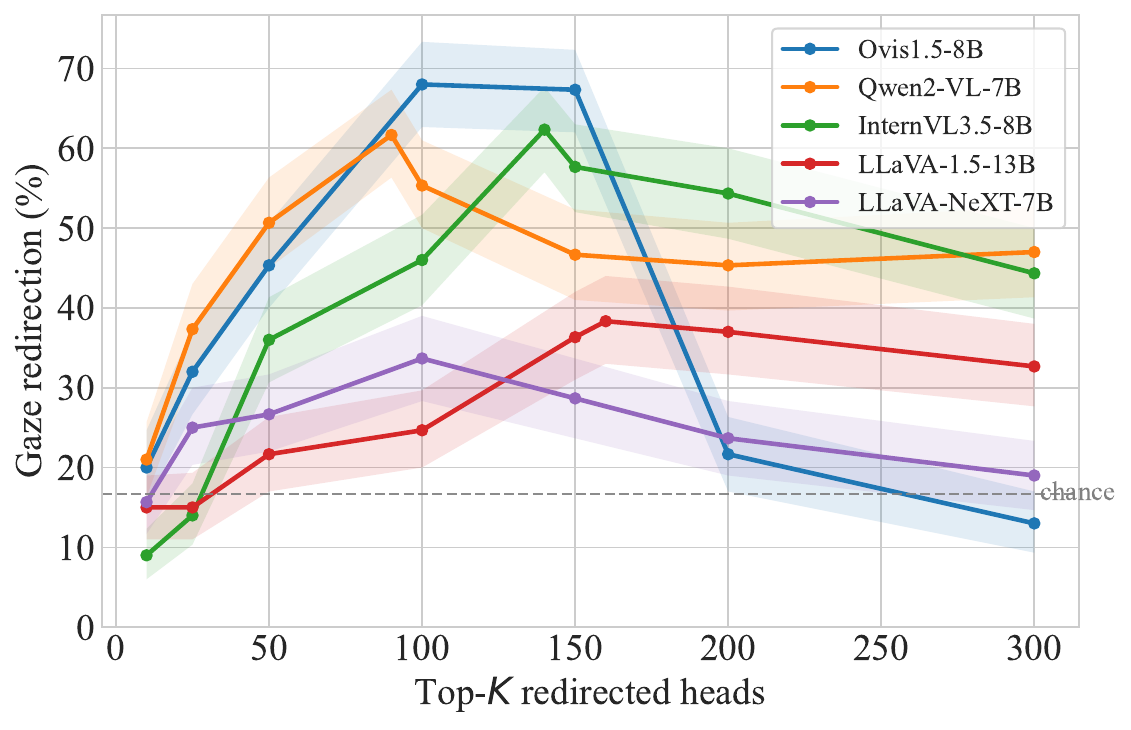}
  \caption{Top-$K$ saturation across the five other-architecture families. Gaze-redirection accuracy (forced 1-of-6 LLM judge, chance $16.7\%$) versus the number of redirected heads $K$, on the 500-strip validation set ($n{=}3{,}000$ per point); shaded bands are bootstrap $95\%$ CIs. Per-model peaks are reported in \cref{tab:cross_arch}. Every family shows the same hump-then-collapse shape but peaks in a different place: Ovis1.5-8B peaks at $K{=}100$ ($68.7\%$) and collapses hardest as the $-\delta$ over-suppression drives outputs to junk, Qwen2-VL-7B and InternVL3.5-8B peak in the mid-$60\%$s (at $K{=}90$ and $K{=}140$) and degrade gently, and the two frozen-encoder LLaVA families plateau near $35$--$39\%$ without a sharp peak.}
  \label{fig:topk_saturation_crossarch}
\end{figure}

\paragraph{Saturation behavior.}
Every other-architecture family shows the same hump-and-collapse shape as Qwen3-VL-8B (\cref{fig:topk_saturation}; \cref{fig:topk_saturation_crossarch} plots the per-family curves), but the peak location and the collapse rate vary. Ovis1.5-8B peaks sharply at $K{=}100$ ($68.7\%$) and then collapses hard: by $K{=}250$ the intervention drives over $80\%$ of outputs to junk and accuracy falls to chance. Qwen2-VL-7B peaks earlier and more gently, at $K{=}90$ ($66.2\%$, between the grid points of \cref{tab:cross_arch_sweep}), and degrades slowly rather than collapsing. InternVL3.5-8B and LLaVA-1.5-13B peak further out: InternVL3.5-8B reaches $62.7\%$ at $K{=}140$ after a steady climb from $K{=}120$, then eases back to $55.0\%$ by $K{=}180$. LLaVA-1.5-13B peaks at $K{=}160$ ($39.0\%$), holding a flat $32$--$39\%$ band from $K{=}125$ to $K{=}175$. LLaVA-NeXT-7B peaks at $K{=}100$ ($35.3\%$) and collapses past $K{=}200$.

\begin{figure}[ht]
  \centering
  \includegraphics[width=\linewidth]{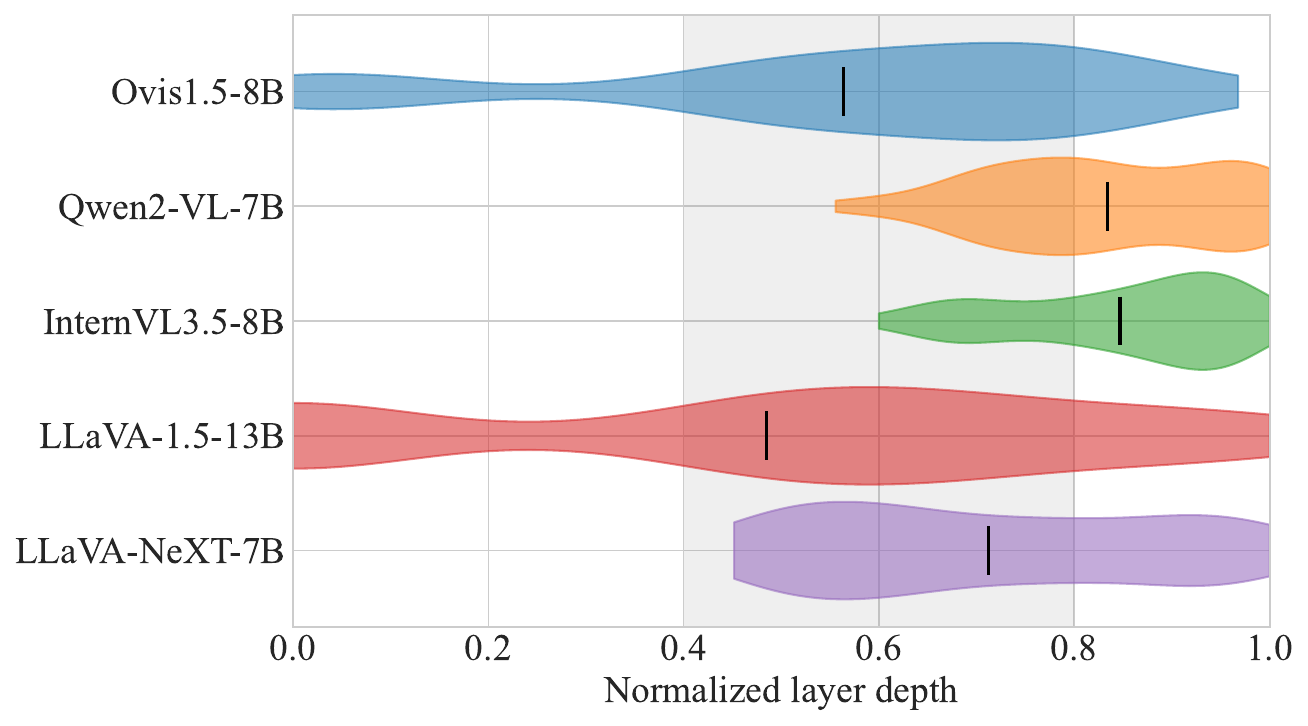}
  \caption{Layer concentration of the top-$100$ gaze heads across the five other-architecture families. Layer indices are normalized to a depth fraction so models with different layer counts ($28$, $32$, $36$, $40$) share one axis; the shaded band marks the mid-to-late region (depth $0.4$--$0.8$) and the tick is each model's mean depth. The top-$100$ gaze heads sit in the second half of every network: Qwen2-VL ($28$ layers) and InternVL3.5 ($36$) concentrate latest (mean depth $\approx 0.84$), Ovis ($32$) and LLaVA-NeXT ($32$) fall in the mid-to-late band, and LLaVA-1.5 ($40$) is the most distributed, with a tail into the early layers. Across architectures the gaze-head construct keeps a consistent geometric meaning: gaze heads are mid-to-late LM heads.}
  \label{fig:layer_distribution_crossarch}
\end{figure}

\paragraph{Non-gaze controls.}
At each model's peak $K$ the non-gaze control, heads sampled from the same layer range as that model's gaze heads (excluding the gaze head set), stays well below the matched gaze condition (\cref{tab:cross_arch}). The level depends on what that band contains for each model: on Qwen2-VL it pins near zero ($0.0\%$) because force-boosting non-gaze heads onto one panel collapses generation to junk on almost every pair; on Ovis ($13.0\%$) and LLaVA-1.5 ($13.8\%$) it sits near the $16.7\%$ chance line; and on InternVL3.5 ($31.0\%$) and LLaVA-NeXT ($26.7\%$) it floats higher, because under the suppress-all intervention some panel signal leaks through even non-gaze heads. In every case the gaze condition clears the non-gaze control by a wide margin at the peak $K$.

\paragraph{Within-family scale.}
\label{app:cross_arch_model_scale}
For the InternVL3.5 and LLaVA-1.5 families we ran both a smaller variant (2B / 7B) and a larger 8B / 13B variant; \cref{tab:cross_arch_scale} reports the peak redirection accuracy for each pair with non-gaze sampled from the same layer range as the gaze heads. In both families the larger variant peaks at a \emph{larger} $K$ ($K{=}50\!\to\!140$ for InternVL3.5, $K{=}150\!\to\!160$ for LLaVA-1.5) at a comparable peak accuracy ($64.0\!\to\!62.7\%$ and $38.7\!\to\!39.0\%$), consistent with the gaze mechanism being spread over more heads at larger scale, so that a fixed-$K$ attention-mask intervention has to reach a larger fraction of them to achieve the same effect. Qwen3-VL remains the family with the cleanest within-family scaling, where the 8B variant is genuinely the strongest size (\cref{app:model_sizes}, 8B peaks at $83.1\%$ at $K{=}100$, above 4B's $72.9\%$ at $K{=}75$ and 32B's $70.2\%$ at $K{=}500$).

\begin{table}[ht]
\centering
\small
\caption{Within-family scale comparison. Peak gaze-redirection accuracy and the $K$ at which it occurs, for each family's smaller and larger variants, on the full 500-strip validation set with non-gaze sampled from the same layer range as the gaze heads.}
\label{tab:cross_arch_scale}
\begin{tabular}{lccc}
\toprule
Family / scale & Peak $K$ & Gaze & Non-gaze \\
\midrule
InternVL3.5-2B & 50  & 64.0 & 24.0 \\
InternVL3.5-8B & 140 & 62.7 & 31.0 \\
\midrule
LLaVA-1.5-7B   & 150 & 38.7 & 16.3 \\
LLaVA-1.5-13B  & 160 & 39.0 & 13.8 \\
\bottomrule
\end{tabular}
\end{table}

\paragraph{Frozen vs.\ trained vision encoders: an exploratory hypothesis.}
We present this section as a hypothesis the data is consistent with, not a conclusion. One pattern we observe is that the redirection magnitude appears to correlate with whether the vision encoder is fine-tuned together with the language model or kept frozen. The three families that exceed $60\%$ all train the encoder on the VLM task. Qwen2-VL learns its native dynamic-resolution ViT end to end, InternVL3.5 trains InternViT at $448$\,px with a learned $2{\times}2$ pixel-shuffle projector, and Ovis fine-tunes its SigLIP-so400m backbone in two of its three training stages while learning the visual vocabulary and embedding table that re-encode it. All three produce a sharp gaze ranking concentrated in the mid-to-late LM layers (\cref{fig:layer_distribution_crossarch}), with the top-$10$ heads in layers $16$--$23$ for Ovis, $19$--$24$ for Qwen2-VL, and $24$--$34$ for InternVL3.5. The two families that plateau near $35\%$ both keep the encoder frozen: LLaVA-1.5 and LLaVA-NeXT bridge a frozen CLIP-ViT-L/14-336 to the LM through a two-layer MLP and never update it on the VLM task. Their strongest heads barely separate from the non-gaze control (LLaVA-NeXT: $35.3\%$ gaze vs.\ $26.7\%$ non-gaze at peak).

A natural reading is that the gaze mechanism may require image tokens that are both spatially addressable and panel-distinct, so that a compact set of heads can learn to select the tokens of the panel being described as a function of the decode query. A frozen encoder optimized for global image-text matching, passed through a thin projector, could give the LM patch features that answer coarse questions but stay too diffuse for such heads to form. This reading is correlational, since the frozen families also use lower input resolution and differ in language backbone. Two further observations are consistent with this hypothesis. Within each family, scale does not move the result, with both LLaVA-1.5 sizes plateauing near $39\%$ and both InternVL3.5 sizes near $62$--$64\%$ (\cref{tab:cross_arch_scale}), so capacity does not appear to be the bottleneck. And the contrastive objective alone is unlikely to be the cause, since Ovis trains a contrastive SigLIP backbone yet supports clean gaze heads. A properly controlled test, freezing versus fine-tuning the same encoder under a fixed language model, would be needed to settle this, and we leave that to future work. The picture above should be read as exploratory analysis pointing toward an open question rather than as a confirmed explanation.

\paragraph{A frozen-vs.-trained comparison on one backbone.}
As a partial step toward such a controlled test, we run a single same-backbone comparison. Bunny-3B bridges a \emph{frozen} SigLIP-so400m encoder, the same family Ovis fine-tunes, to a Phi-2 language model through a two-layer MLP. Its discovered gaze heads do not redirect in our setup: the gaze accuracy peaks at only $8.3\%$ at $K{=}10$ and stays below the $16.7\%$ chance line at every $K$ (\cref{tab:bunny_frozen_sweep}), far below Ovis's $68.7\%$ on the same backbone, while non-gaze redirection is $0\%$ throughout. Under the more aggressive interventions, steering the heads collapses generation into refusals rather than moving the answer to the queried panel. Model size is unlikely to be the full explanation, since a \emph{smaller} trained model, InternVL3.5-2B, redirects at $64\%$ (\cref{tab:cross_arch_scale}). Same backbone family, frozen versus trained, opposite outcome. We treat this single comparison as suggestive evidence consistent with the hypothesis above, not as proof; many other factors differ between Bunny and Ovis, and a fully controlled study remains open.

\begin{table}[ht]
\centering
\small
\caption{Frozen-encoder control. Top-$K$ redirection for Bunny-3B, which bridges a \emph{frozen} SigLIP-so400m encoder to a Phi-2 LM through a two-layer MLP: gaze vs.\ non-gaze accuracy with the gaze-condition junk\%, on the 500-strip validation set under the identical intervention. Unlike every trained-encoder family, Bunny never clears the $16.7\%$ chance line at any $K$.}
\label{tab:bunny_frozen_sweep}
\begin{tabular}{lccc}
\toprule
$K$ & Gaze & Non-gaze & junk\% \\
\midrule
10  & \textbf{8.3} & 0.0 & 25.7 \\
25  & 7.0 & 0.0 & 19.7 \\
50  & 1.7 & 0.0 & 4.7 \\
100 & 1.0 & 0.0 & 5.0 \\
150 & 1.3 & 0.0 & 21.7 \\
200 & 5.0 & 0.0 & 39.3 \\
300 & 2.3 & 0.0 & 82.3 \\
\bottomrule
\end{tabular}
\end{table}

% =============================================================================
\section{Qualitative Samples}
\label{app:qualitative}

\subsection{Visual Question Answering}
\label{app:qual_vqa}

\begin{figure*}[t]
  \centering
  \includegraphics[width=\linewidth]{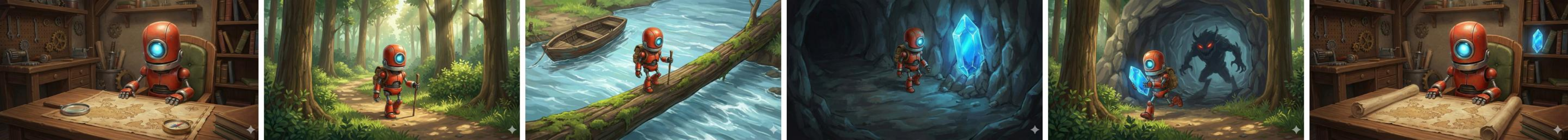}

  \vspace{4pt}
  \begin{tcolorbox}[colback=gray!3, colframe=black!70, fontupper=\small,
    title={\small\textbf{Visual QA task with Gaze Steering}},
    left=4pt, right=4pt, top=3pt, bottom=3pt]
  \scriptsize
  ``Read each of the panels and tell me. What is the main action happening in this particular comic panel? Just output the answer in few words, do not include any other text.''

  \medskip
  \textcolor{red!70!black}{\textbf{Baseline}} Robot plans adventure

  \medskip
  \textcolor{blue!70!black}{Steered Panel 1} Robot studying a map
  \medskip
  
  \textcolor{blue!70!black}{Steered Panel 2} Robot explores forest 
  \medskip
  
  \textcolor{blue!70!black}{Steered Panel 3} Robot explores riverbank.
  \medskip
  
  \textcolor{blue!70!black}{Steered Panel 4} Robot explores cave.
  \medskip
  
  \textcolor{blue!70!black}{Steered Panel 5} Robot explores forest, fights monster, and faces danger.
  \medskip
  
  \textcolor{blue!70!black}{Steered Panel 6} Robot studies blueprint
  \end{tcolorbox}

  \caption{Qualitative example of how the visual QA response of the model changes when steering the gaze heads' attention to any particular target panel. The original baseline response is a summary of all 6 panels. But when we steer and lock the gaze to a fixed panel, the response is panel-specific for the same exact prompt.}
  \label{fig:qual_vqa1}
\end{figure*}

\begin{figure*}[t]
  \centering
  \includegraphics[width=\linewidth]{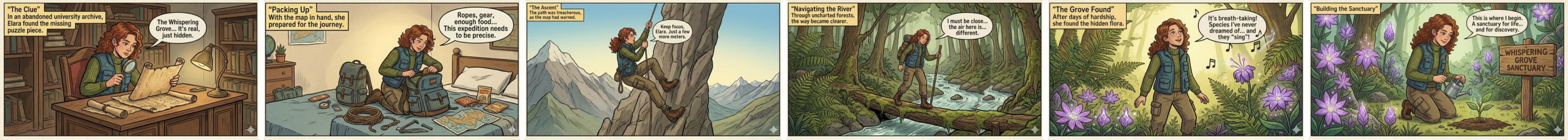}

  \vspace{4pt}
  \begin{tcolorbox}[colback=gray!3, colframe=black!70, fontupper=\small,
    title={\small\textbf{Visual QA task with Gaze Steering}},
    left=4pt, right=4pt, top=3pt, bottom=3pt]
  \scriptsize
  ``Read each of the panels and tell me. What is the main action happening in this particular comic panel? Just output the answer in few words, do not include any other text.''

  \medskip
  \textcolor{red!70!black}{\textbf{Baseline}} Elara goes through an adventure

  \medskip
  \textcolor{blue!70!black}{Steered Panel 1} Elara discovers map clue
  \medskip
  
  \textcolor{blue!70!black}{Steered Panel 2} Elise packs for trip
  \medskip
  
  \textcolor{blue!70!black}{Steered Panel 3} Elaine climbs mountain, climbs, climbs.
  \medskip
  
  \textcolor{blue!70!black}{Steered Panel 4} Elara explores forest.
  \medskip
  
  \textcolor{blue!70!black}{Steered Panel 5} Elara explores magical forest.
  \medskip
  
  \textcolor{blue!70!black}{Steered Panel 6} Elara discovers a new plant species
  \end{tcolorbox}

  \caption{Qualitative example of how the visual QA response of the model changes when steering the gaze heads' attention to any particular target panel. The original baseline response is a summary of all 6 panels. But when we steer and lock the gaze to a fixed panel, the response is panel-specific for the same exact prompt.}
  \label{fig:qual_vqa2}
\end{figure*}

\begin{figure*}[t]
  \centering
  \includegraphics[width=\linewidth]{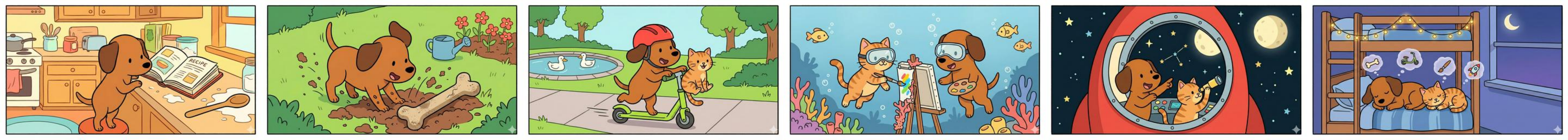}

  \vspace{4pt}
  \begin{tcolorbox}[colback=gray!3, colframe=black!70, fontupper=\small,
    title={\small\textbf{Visual QA task with Gaze Steering}},
    left=4pt, right=4pt, top=3pt, bottom=3pt]
  \scriptsize
  ``Read each of the panels and tell me. What is the main action happening in this particular comic panel? Just output the answer in few words, do not include any other text.''

  \medskip
  \textcolor{red!70!black}{\textbf{Baseline}} Dog befriends a cat

  \medskip
  \textcolor{blue!70!black}{Steered Panel 1} Dog reads recipe book.
  \medskip
  
  \textcolor{blue!70!black}{Steered Panel 2} Dog digs up dirt.
  \medskip
  
  \textcolor{blue!70!black}{Steered Panel 3} Dog and cat enjoy outdoor and park activities.
  \medskip
  
  \textcolor{blue!70!black}{Steered Panel 4}Dog and fish underwater.
  \medskip
  
  \textcolor{blue!70!black}{Steered Panel 5} Dog and cat explore space.
  \medskip
  
  \textcolor{blue!70!black}{Steered Panel 6} Dog and cat sleep together.
  \end{tcolorbox}

  \caption{Qualitative example of how the visual QA response of the model changes when steering the gaze heads' attention to any particular target panel. The original baseline response is a summary of all 6 panels. But when we steer and lock the gaze to a fixed panel, the response is panel-specific for the same exact prompt.}
  \label{fig:qual_vqa3}
\end{figure*}

We present qualitative examples of gaze-head steering on the VQA task in \cref{fig:qual_vqa1}, \cref{fig:qual_vqa2}, and \cref{fig:qual_vqa3}.
Under baseline conditions with no steering, the model produces a summarized answer that draws from multiple panels in the strip.
When gaze heads are redirected to attend to a specific panel, the answer becomes highly specific to that panel's content, accurately reflecting its visual details while ignoring the other panels.

\subsection{Free-Form Narration with Dynamic Gaze Switching}
\label{app:qual_narration}

\begin{figure*}[t]
  \centering
  \includegraphics[width=\linewidth]{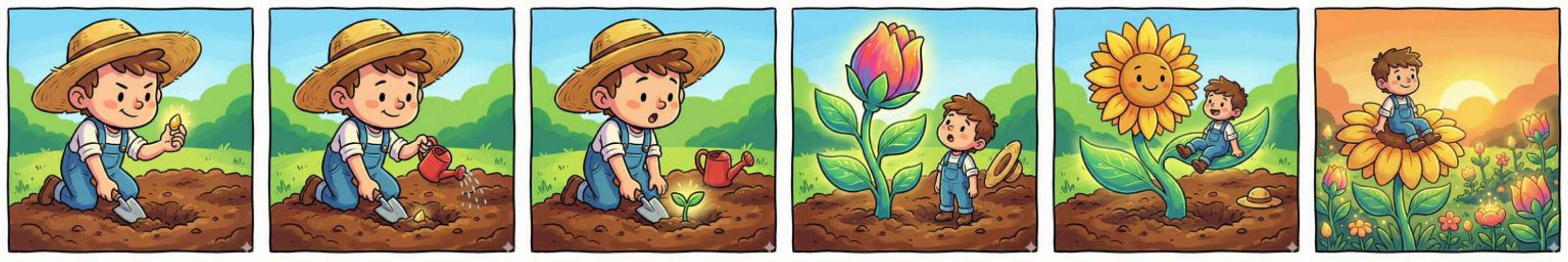}
  % \vspace{2pt}
  \includegraphics[width=\linewidth]{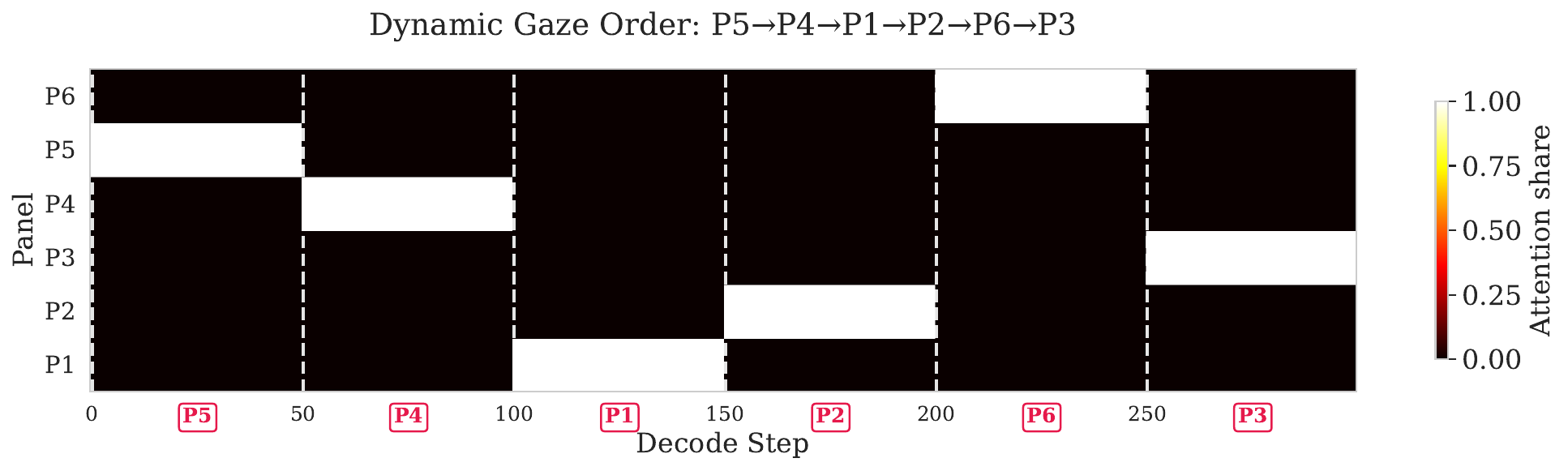}
  % \vspace{2pt}
  \begin{tcolorbox}[colback=white, colframe=black!70, fontupper=\small,
    title={\small\textbf{Generated narration under dynamic gaze switching} \\ {\scriptsize(schedule: \textcolor{red!70!black}{P5}\,$\to$\,\textcolor{blue!70!black}{P4}\,$\to$\,\textcolor{orange!80!black}{P1}\,$\to$\,\textcolor{green!50!black}{P2}\,$\to$\,\textcolor{purple!70!black}{P6}\,$\to$\,\textcolor{brown!70!black}{P3}, switching every 50 tokens)}},
    coltitle=black, colbacktitle=white,
    left=4pt, right=4pt, top=3pt, bottom=3pt]
  \scriptsize
  \textcolor{red!70!black}{Panel 1: A child sits atop a giant sunflower, smiling joyfully as a giant hand gently holds a small, smiling sunflower seedling. The sun shines brightly, and the child's face beams with happiness. Panel 2: A}
  \textcolor{blue!70!black}{child stands beside a large, blooming flower, holding a small plant. A giant hand holds a flower bud, and the child watches a small plant grow beside a tall, budding flower. Panel 3: A child stands beside a large, blo}
  \textcolor{orange!80!black}{oming flower. A small child digs in the ground, and the child watches a small seed before planting it. He is wearing overalls with hat.}
  \textcolor{green!50!black}{The child waters the ground, and the child watches a small plant grow with a large, bright smile. Panel 4: A child stands beside a small, smiling flower next to a small watering can}
  \textcolor{purple!70!black}{watches a small flower grow beside a large, blooming flower field. Panel 5: A child sits atop of bloomed flowers beside a large, blooming flower, holding a small flower. The child looks at sunset, and the child watches a large, blooming}
  \textcolor{brown!70!black}{flower grow. Panel 6: A child kneels beside a red watering can, smiling flower, holding a small shovel. The child looks at a small plant, and the child watches a small plant grow in a awe. The child,}
  \end{tcolorbox}
\caption{\emph{Top:}~The six-panel strip used for evaluation.
\emph{Middle:}~Manually altered Gaze-head attention during generation; every 50 tokens the target switches to a new panel.
\emph{Bottom:}~The model maintains its default numbering (``1, 2, 3\ldots'') but describes the content of whichever panel the gaze heads are steered toward. At the transition point, the model naturally ends and starts a new segment.}
  \label{fig:sequence1}
\end{figure*}

\begin{figure*}[t]
  \centering
  
  \includegraphics[width=\linewidth]{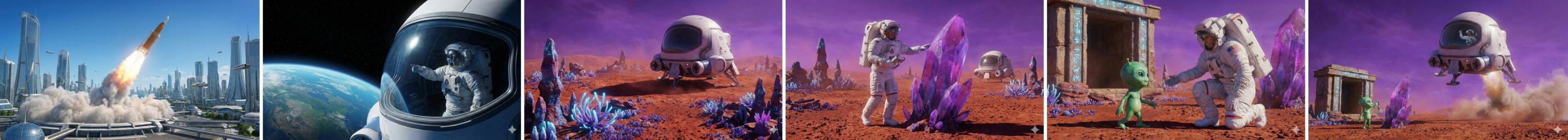}
  % \vspace{2pt}
  \includegraphics[width=\linewidth]{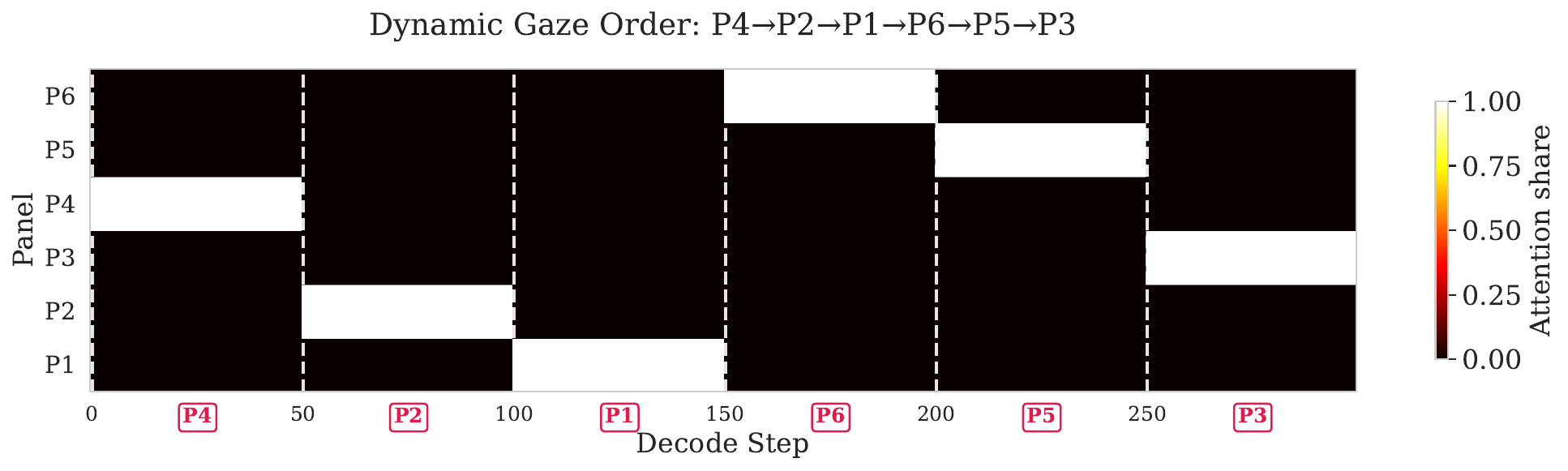}
  % \vspace{2pt}
  \begin{tcolorbox}[colback=white, colframe=black!70, fontupper=\small,
    title={\small\textbf{Generated narration under dynamic gaze switching} \\ {\scriptsize(schedule: \textcolor{red!70!black}{P4}\,$\to$\,\textcolor{blue!70!black}{P2}\,$\to$\,\textcolor{orange!80!black}{P1}\,$\to$\,\textcolor{green!50!black}{P6}\,$\to$\,\textcolor{purple!70!black}{P5}\,$\to$\,\textcolor{brown!70!black}{P3}, switching every 50 tokens)}},
    coltitle=black, colbacktitle=white,
    left=4pt, right=4pt, top=3pt, bottom=3pt]
  \scriptsize
  \textcolor{red!70!black}{1. A futuristic explorer stands amidst alien terrain, surveying a vibrant, otherworldly landscape. A crystalline alien world awaits discovery, with a mysterious, glowing alien artifact nearby. 2. A lone astronaut explores a surreal, alien landscape, standing}
  \textcolor{blue!70!black}{beside a vast, curved window. The astronaut gazes at Earth's curved horizon, a breathtaking view of space and humanity's cosmic journey. 3. A breathtaking view of Earth from space, with a astronaut in a spacesuit gazing out from a}
  \textcolor{orange!80!black}{spacecraft. The futuristic cityscape below contrasts with the serene, futuristic skyline.4. A futuristic cityscape with towering skyscrapers and a bustling urban landscape. A massive, futuristic rocket launches from a futuristic cityscape, with a futuristic cityscape in}
  \textcolor{green!50!black}{the background. 5. A futuristic alien landscape with a mysterious, glowing alien artifact. A lone astronaut explores a mysterious, alien world, with a futuristic spaceship taking off in the background.}
  \textcolor{purple!70!black}{6. A futuristic alien landscape with a mysterious, glowing alien artifact. A large, friendly alien stands beside a human astronaut, both gazing at each other in awe}
  \textcolor{brown!70!black}{. A vibrant, alien landscape stretches out with a spaceship landed surrounded by glowing alien crystals.}
  \end{tcolorbox}
\caption{\emph{Top:}~The six-panel strip used for evaluation.
\emph{Middle:}~Manually altered Gaze-head attention during generation; every 50 tokens the target switches to a new panel.
\emph{Bottom:}~The model maintains its default numbering (``1, 2, 3\ldots'') but describes the content of whichever panel the gaze heads are steered toward. At the transition point, the model naturally ends and starts a new segment.}
  \label{fig:sequence2}
\end{figure*}

\begin{figure*}[t]
  \centering
  
  \includegraphics[width=\linewidth]{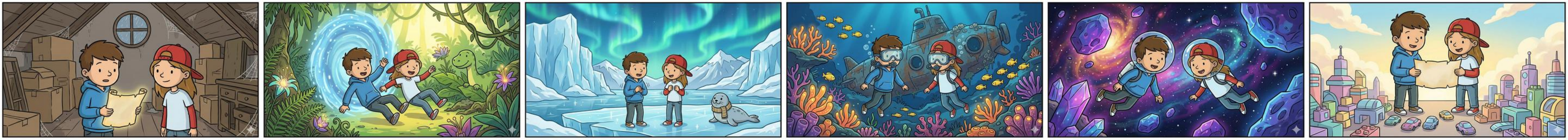}
  % \vspace{2pt}
  \includegraphics[width=\linewidth]{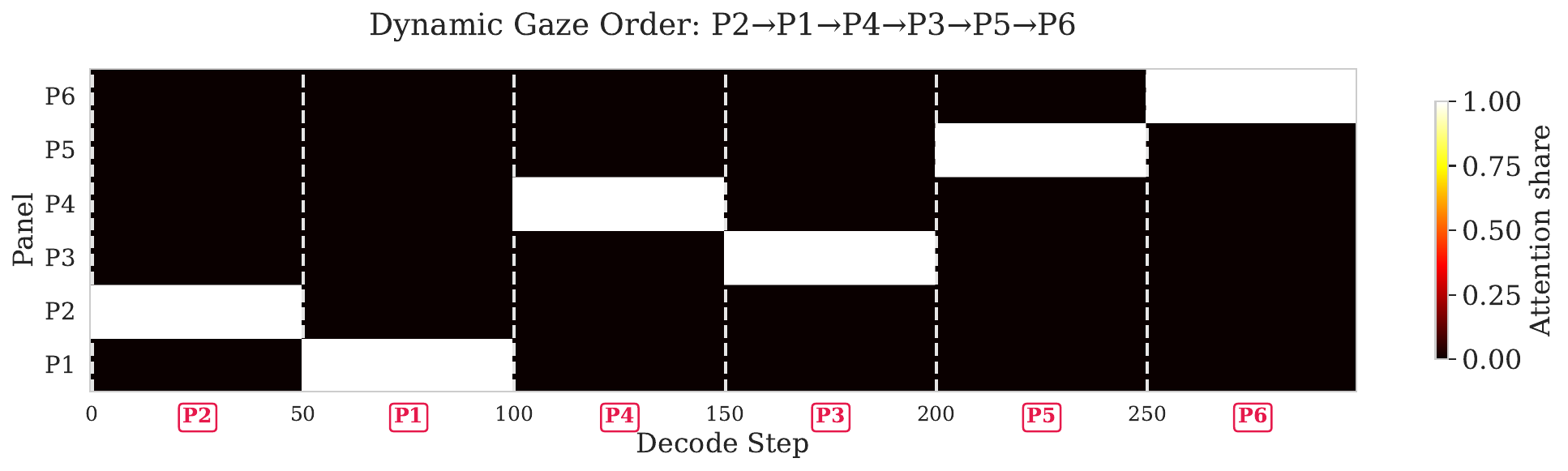}
  % \vspace{2pt}
  \begin{tcolorbox}[colback=white, colframe=black!70, fontupper=\small,
    title={\small\textbf{Generated narration under dynamic gaze switching} \\ {\scriptsize(schedule: \textcolor{red!70!black}{P2}\,$\to$\,\textcolor{blue!70!black}{P1}\,$\to$\,\textcolor{orange!80!black}{P4}\,$\to$\,\textcolor{green!50!black}{P3}\,$\to$\,\textcolor{purple!70!black}{P5}\,$\to$\,\textcolor{brown!70!black}{P6}, switching every 50 tokens)}},
    coltitle=black, colbacktitle=white,
    left=4pt, right=4pt, top=3pt, bottom=3pt]
  \scriptsize
  \textcolor{red!70!black}{1. Children discover a magical forest through a portal, surrounded by vibrant flora and a glowing jungle, as they journey into a fantastical, enchanted world. 2. A vibrant jungle scene with a glowing portal, where a child and a dinosaur coexist}
  \textcolor{blue!70!black}{in a cozy, rustic setting, hinting at a hidden world of adventure and discovery. 3. A cozy attic with a child and a dog, surrounded by books and a mysterious attic, hinting at a hidden world of adventure and discovery. 4}
  \textcolor{orange!80!black}{. A vibrant underwater scene with colorful coral, a school of fish, and a coral reef, where a child and a coral reef, with a school of fish, explore a colorful underwater world. 5. A vibrant underwater scene with colorful coral, a}
  \textcolor{green!50!black}{school of icebergs, and a polar bear, where a child and a polar bear, with a snowy landscape, explore a frozen wonderland. 6. A serene Arctic landscape with a polar bear, where a child and a polar bear, with}
  \textcolor{purple!70!black}{a starry sky, explore a cosmic adventure through space. 7. A cosmic scene with a child and a space explorer, surrounded by stars and planets, as they journey through space, exploring the universe with wonder and excitement. 8. A cosmic scene}
  \textcolor{brown!70!black}{with a child and a cityscape, where a child and a cityscape, with a futuristic city, explore a vibrant, imaginative world with a playful, adventurous spirit.}
  \end{tcolorbox}
\caption{\emph{Top:}~The six-panel strip used for evaluation.
\emph{Middle:}~Manually altered Gaze-head attention during generation; every 50 tokens the target switches to a new panel.
\emph{Bottom:}~The model maintains its default numbering (``1, 2, 3\ldots'') but describes the content of whichever panel the gaze heads are steered toward. At the transition point, the model naturally ends and starts a new segment.}
  \label{fig:sequence3}
\end{figure*}

We present qualitative examples of gaze-head steering during free-form narration in \cref{fig:sequence1}, \cref{fig:sequence2}, and \cref{fig:sequence3}.
Under baseline conditions, the model describes each panel in the default left-to-right order.
When gaze heads are redirected through a sequence of target panels, the model produces a fluid narration that smoothly stitches together descriptions of each targeted panel.
At each gaze switch, the model naturally wraps up its current panel description and transitions into describing the next target, integrating the shift into the flow of the text rather than producing abrupt breaks.

\begin{figure*}[t]
  \centering
  \includegraphics[width=\linewidth]{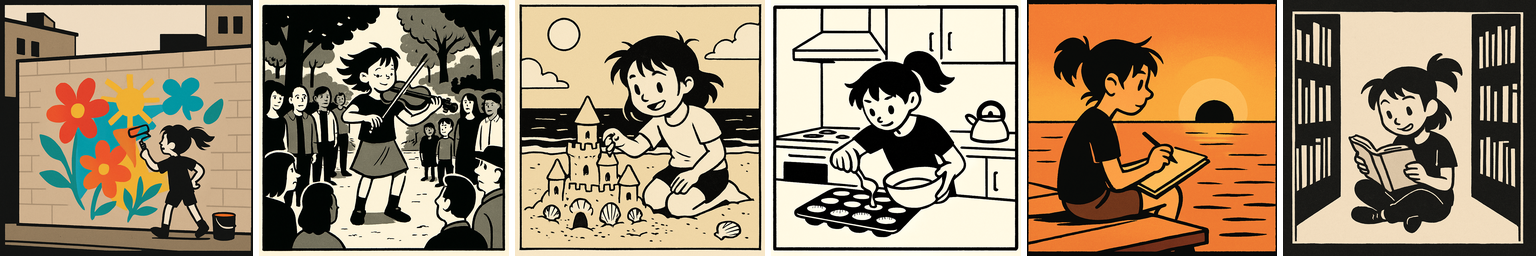}

  \vspace{4pt}
  \begin{tcolorbox}[colback=gray!3, colframe=black!70, fontupper=\small,
    title={\small\textbf{Visual QA task with Gaze Steering (6/6 hits, OpenAI 500 dataset, top-100 gaze heads)}},
    left=4pt, right=4pt, top=3pt, bottom=3pt]
  \scriptsize
  ``What is the main action or event happening in this comic strip? Answer briefly.''

  \medskip
  \textcolor{red!70!black}{\textbf{Baseline (no steering):}} A girl is shown engaging in various activities: painting a mural, playing violin\ldots

  \medskip
  \textcolor{blue!70!black}{Steered Panel 1} \emph{A person is painting a mural on a wall.}
  \medskip

  \textcolor{blue!70!black}{Steered Panel 2} \emph{A group of people are attending a street performance.}
  \medskip

  \textcolor{blue!70!black}{Steered Panel 3} \emph{A child playing in the sand.}
  \medskip

  \textcolor{blue!70!black}{Steered Panel 4} \emph{A woman is cooking in a kitchen.}
  \medskip

  \textcolor{blue!70!black}{Steered Panel 5} \emph{A person is drawing or painting a sunset scene while sitting\ldots}
  \medskip

  \textcolor{blue!70!black}{Steered Panel 6} \emph{A person is reading a book.}
  \end{tcolorbox}
  \caption{Qualitative example of gaze-head VQA steering on a strip from the OpenAI 500 dataset. The baseline answer summarizes the strip generically. Redirecting the top-100 gaze heads to each panel in turn shifts the model's answer to that panel's specific activity (mural, street performance, beach, kitchen, sunset painting, reading). All six steered answers match the target panel's visual content under our forced 1-of-6 LLM judge.}
  \label{fig:qual_vqa_openai_143}
\end{figure*}

\begin{figure*}[t]
  \centering
  \includegraphics[width=\linewidth]{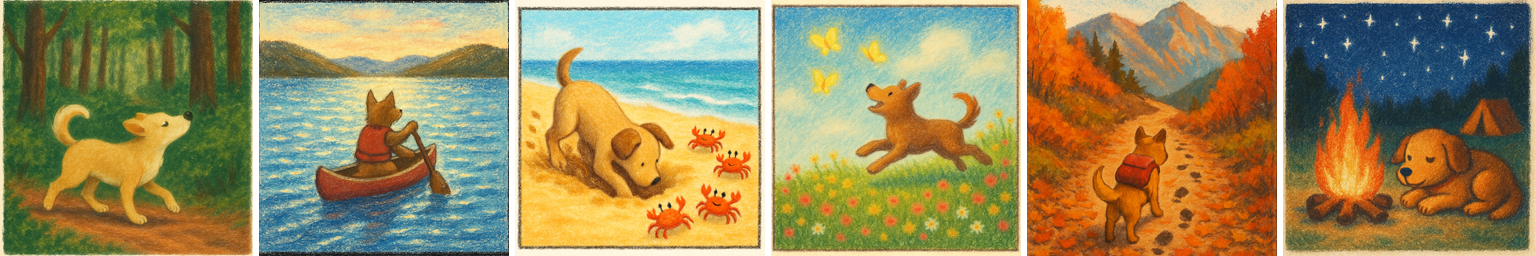}

  \vspace{4pt}
  \begin{tcolorbox}[colback=gray!3, colframe=black!70, fontupper=\small,
    title={\small\textbf{Visual QA task with Gaze Steering (6/6 hits, OpenAI 500 dataset, top-100 gaze heads)}},
    left=4pt, right=4pt, top=3pt, bottom=3pt]
  \scriptsize
  ``What is the main action or event happening in this comic strip? Answer briefly.''

  \medskip
  \textcolor{red!70!black}{\textbf{Baseline (no steering):}} A dog goes on a multi-day camping adventure.

  \medskip
  \textcolor{blue!70!black}{Steered Panel 1} \emph{A dog walks through a forest, then runs through a field, and\ldots}
  \medskip

  \textcolor{blue!70!black}{Steered Panel 2} \emph{A person is paddling a boat in a lake.}
  \medskip

  \textcolor{blue!70!black}{Steered Panel 3} \emph{A crab is playing with a dog.}
  \medskip

  \textcolor{blue!70!black}{Steered Panel 4} \emph{A dog chasing butterflies.}
  \medskip

  \textcolor{blue!70!black}{Steered Panel 5} \emph{A hiker hikes through a scenic mountain trail.}
  \medskip

  \textcolor{blue!70!black}{Steered Panel 6} \emph{A campfire.}
  \end{tcolorbox}
  \caption{Another qualitative example. Each steered answer picks up a distinct activity (forest walk, kayaking, beach with crab, butterflies, mountain hike, campfire) from the corresponding panel, while the baseline collapses to a single sentence ``camping adventure''. The model's answer follows where we point the top-100 gaze heads, even though the prompt asks about the whole strip.}
  \label{fig:qual_vqa_openai_128}
\end{figure*}

\begin{figure*}[t]
  \centering
  \includegraphics[width=\linewidth]{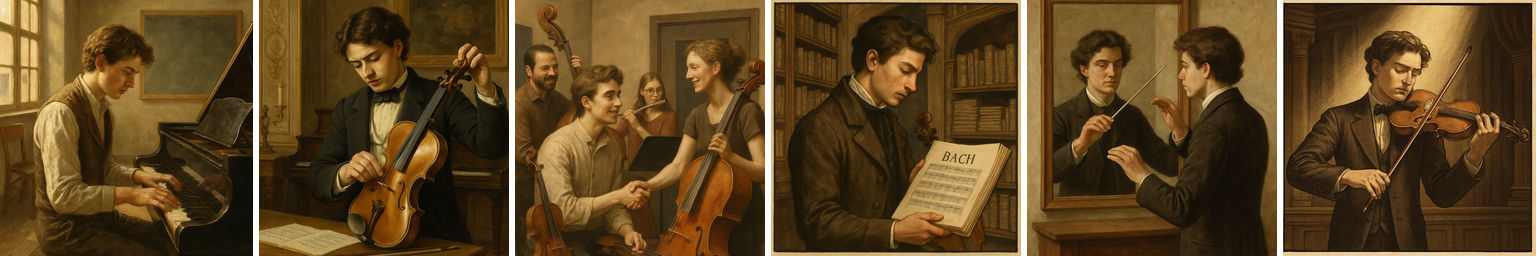}

  \vspace{4pt}
  \begin{tcolorbox}[colback=gray!3, colframe=black!70, fontupper=\small,
    title={\small\textbf{Visual QA task with Gaze Steering (InternVL3.5-2B, top-50 gaze heads, 6/6 distinct)}},
    left=4pt, right=4pt, top=3pt, bottom=3pt]
  \scriptsize
  ``What is the main action or event happening in this comic strip? Answer briefly.''

  \medskip
  \textcolor{red!70!black}{\textbf{Baseline (no steering):}} A group of musicians preparing for a\ldots

  \medskip
  \textcolor{blue!70!black}{Steered Panel 1} \emph{A man playing a guitar while\ldots}
  \medskip

  \textcolor{blue!70!black}{Steered Panel 2} \emph{The musicians playing their instruments and\ldots}
  \medskip

  \textcolor{blue!70!black}{Steered Panel 3} \emph{A group of people discussing and\ldots}
  \medskip

  \textcolor{blue!70!black}{Steered Panel 4} \emph{A man reading a book and\ldots}
  \medskip

  \textcolor{blue!70!black}{Steered Panel 5} \emph{The interaction between the men in\ldots}
  \medskip

  \textcolor{blue!70!black}{Steered Panel 6} \emph{A man playing a violin\ldots}
  \end{tcolorbox}
  \caption{Qualitative example of gaze-head VQA steering on InternVL3.5-2B at the model's saturation peak ($K{=}50$; $64.0\%$ peak accuracy, \cref{tab:cross_arch_scale}). The baseline answer summarizes the strip generically. Redirecting the top-$50$ gaze heads to each panel in turn shifts the model's answer to that panel's specific musical activity (guitar, ensemble, conversation, reading, interaction, violin). All six steered answers are distinct and correspond to the visual content of the targeted panel.}
  \label{fig:qual_vqa_internvl35_187}
\end{figure*}

% Gemma 4 removed from the model set; qualitative example dropped.
% \include{appendix_qualitative/gaze_vqa_gemma4_comic186}


\end{document}